\documentclass{article}

\usepackage[preprint]{neurips_2023}

\usepackage[dvipsnames, svgnames, x11names]{xcolor}
\usepackage[utf8]{inputenc} %
\usepackage[T1]{fontenc}    %
\usepackage{hyperref}       %
\usepackage{url}            %
\usepackage{booktabs}       %
\usepackage{amsfonts}       %
\usepackage{nicefrac}       %
\usepackage{microtype}      %
\usepackage{xcolor}         %
\usepackage{multirow}
\usepackage{CJKutf8}
\usepackage{colortbl}
\usepackage{svg}
\usepackage{amssymb}%
\usepackage{pifont}%

\newcommand{\cmark}{\ding{51}}%
\newcommand{\xmark}{\ding{55}}%
\usepackage{graphicx}

\usepackage{natbib}
\setcitestyle{numbers,square}

\definecolor{'deep1'}{HTML}{C5E6F8} 
\definecolor{'shallow1'}{HTML}{E4F3FC} 
\definecolor{'deep2'}{HTML}{E5F5B7} 
\definecolor{'shallow2'}{HTML}{F3FADF} 

\definecolor{'deep3'}{HTML}{FFE5C6} 
\definecolor{'shallow3'}{HTML}{FFF2E3}

\usepackage{todonotes}
\makeatletter
\newcommand*\iftodonotes{\if@todonotes@disabled\expandafter\@secondoftwo\else\expandafter\@firstoftwo\fi}  %
\makeatother

\title{\includesvg[scale=0.12]{Figure/logo_small.svg} LawBench: Benchmarking Legal Knowledge of Large Language Models}

\author{%
  Zhiwei Fei$^{1\dagger}$, Xiaoyu Shen$^{2\dagger}$\thanks{Work Done Outside Amazon}, Dawei Zhu$^{3\dagger}$, Fengzhe Zhou$^4$ \\
  \textbf{Zhuo Han$^1$, Songyang Zhang$^4$, Kai Chen$^4$, Zongwen Shen$^1$, Jidong Ge$^{1\ddagger}$} \\
  $^1$National Key Laboratory for Novel Software Technology, Nanjing University\\
  $^2$ Amazon Alexa AI \ \ $^3$ Saarland University \ \ $^4$ Shanghai AI Laboratory\\
  \hspace{-0.25cm}$^{\dagger}$ Equal Contribution  \ \ $^{\ddagger}$ Corresponding Author
}

\begin{document}

\maketitle

\begin{abstract}

Large language models (LLMs) have demonstrated strong capabilities in various aspects. However, when applying them to the highly specialized, safe-critical legal domain, it is unclear how much legal knowledge they possess and whether they can reliably perform legal-related tasks. To address this gap, we propose a comprehensive evaluation benchmark \emph{LawBench}. LawBench has been meticulously crafted to have precise assessment of the LLMs' legal capabilities from three cognitive levels: (1) \emph{Legal knowledge memorization}: whether LLMs can memorize needed legal concepts, articles and facts; (2) \emph{Legal knowledge understanding}: whether LLMs can comprehend entities, events and relationships within legal text; (3) \emph{Legal knowledge applying}: whether LLMs can properly utilize their legal knowledge and make necessary reasoning steps to solve realistic legal tasks. LawBench contains 20 diverse tasks covering 5 task types: single-label classification (SLC), multi-label classification (MLC), regression, extraction and generation. We perform extensive evaluations of 51 LLMs on LawBench, including 20 multilingual LLMs, 22 Chinese-oriented LLMs and 9 legal specific LLMs. The results show that GPT-4 remains the best-performing LLM in the legal domain, surpassing the others by a significant margin. While fine-tuning LLMs on legal specific text brings certain improvements, we are still a long way from obtaining usable and reliable LLMs in legal tasks. All data, model predictions and evaluation code are released in \url{https://github.com/open-compass/LawBench/}. We hope this benchmark provides in-depth understanding of the LLMs' domain-specified capabilities and speed up the development of LLMs in the legal domain.

\end{abstract}

\begin{figure}[t]
    \centering
    \includegraphics[width=\columnwidth]{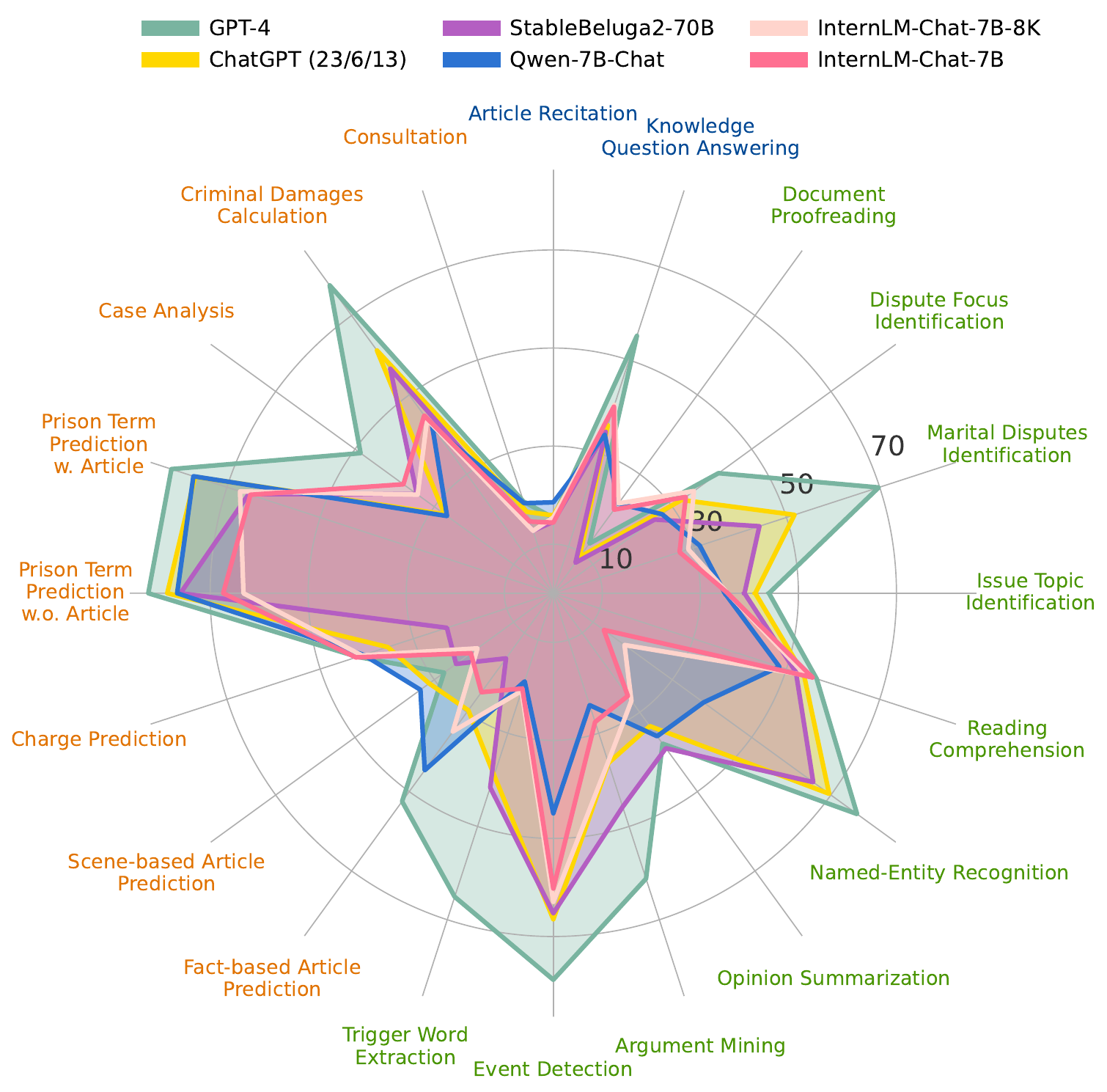}
    \caption{Results (zero-shot) of six best-performing LLMs evaluated on 20 diverse legal tests covering three cognitive dimensions: \textcolor[HTML]{004995}{legal knowledge memorization}, \textcolor[HTML]{499500}{understanding}, and \textcolor[HTML]{e17300}{applying}.}
    \label{fig:law_bench_radar_plot}
\end{figure}

\section{Introduction}
The artificial intelligence community has witnessed notable progress in the large language models (LLMs) recently. The latest large language models, such as GPT-4~\cite{openai2023gpt4} and LLaMA~\cite{touvron2023llama}, have showcased remarkable capabilities and intricate skills that are comparable to, and in some cases, even surpass human capabilities~\cite{huang2023chatgpt}.
To quantitatively assess the capabilities of LLMs, evaluation benchmarks serve a pivotal role in their development. Numerous recent endeavors mainly concentrate on enhancing general and universal capabilities, such as understanding world knowledge and executing complex reasoning~\cite{zhong2023agieval,yu2023kola}.

Previous research on evaluation has predominantly concentrated on exploring the diverse aspects of the general capabilities of LLMs. SuperGLUE~\cite{wang2019superglue} is crafted to assess various facets of language understanding, encompassing reading comprehension, commonsense reasoning, and entailment. Benchmarks such as MMLU~\cite{hendrycks2020measuring} and BIG-bench~\cite{srivastava2023beyond} have been introduced to cover a wide range of NLP tasks for LLM evaluations. Furthermore, there exist benchmarks~\cite{fu2023chain,suzgun2022challenging,fu2023codeapex,chen2021evaluating} explicitly designed to scrutinize the advanced abilities of LLMs that manifest with scale, such as reasoning~\cite{sawada2023arb} and coding~\cite{lu2021codexglue}. Beyond these specific benchmarks, more recent initiatives, like OpenCompass~\cite{2023opencompass}  and HELM~\cite{liang2022holistic}, aspire to offer a comprehensive perspective on the capabilities of LLMs.

However, evaluating Large Language Models (LLMs) necessitates not only a focus on general capabilities but also the incorporation of domain-specific benchmarks for assessing models specialized in particular fields~\cite{zhao2023domain}. Legal tasks encompass a broad spectrum of applications, predominantly text-based, necessitating comprehension and interpretation of highly professional legal texts. Currently, they are primarily conducted by legal experts, who require years of extensive specialized training to process legal cases. Endowing LLMs with legal expertise can not only improve the working efficiency of legal officers, but also address the overwhelming demand of legal assistance from non-professionals, and thereby improve public access to justice~\cite{cui2022survey,trozze2023large}. Consequently, we prioritize establishing benchmarks to measure the legal knowledge of existing LLMs.

In contrast to assessments that primarily focus on testing model's ability to pass legal bar exams~\cite{bommarito2022gpt,zhong2023agieval},  which are not always representative of the actual use-cases for LLMs~\footnote{As shown in \cite{chalkidis2022lexglue}, ChatGPT excels in passing legal bar exams but struggles at performing realistic legal tasks.}, our emphasis is on examining a structured set of legal skills required in real-world scenarios. In the legal domain, there have been works ensembling legal-related tasks including the lexglue~\cite{chalkidis2022lexglue} focusing on EU and American laws, and LBOX OPEN~\cite{hwang2022multi} focusing on South Korean laws, \cite{chalkidis2023chatgpt} further transformed lexglue into zero-shot forms to test the ability of LLMs to complete these tasks. LEGALBENCH~\cite{guha2023legalbench}, in particular, presented the first steps towards constructing an interdisciplinary collaborative legal reasoning benchmark for the English language and evaluated 20 LLMs in 162 legal tasks. 

Nonetheless, legal systems vary significantly among different countries, highlighting the importance of establishing different standards for each legal system. The Chinese legal system is rooted in the civil law family. Unlike the common law system, which is widely accepted in the U.S. and the U.K., judges in the civil law system are obliged to stay neutral, respect the established statutory law articles and ground their decisions on them. Understanding and applying existing statutes and codes, rather than studies of precedents, are of paramount importance~\cite{zheng1986china}. Therefore, it is necessary to design a separate set of evaluation tasks to emphasize the required skill set for the Chinese law system. 

With this in mind, we present \emph{LawBench}:  a meticulously crafted, comprehensive evaluation benchmark to assess the LLMs’ capabilities on performing legal-related tasks under the Chinese civil-law system. LawBench consists of 20 diverse tasks following 5 categories:  single-label classification (SLC), multi-label classification (MLC), regression, extraction and generation.
We divided these tasks into 3 skill levels according to widely accepted Bloom's cognitive models~\cite{krathwohl2002revision}: (1) Legal knowledge memorization: whether LLMs can memorize needed legal concepts, articles and facts; (2) Legal knowledge understanding: whether LLMs can comprehend entities, events and relationships within legal text; (3) Legal knowledge applying: whether LLMs can properly utilize their legal knowledge and make necessary reasoning steps to solve realistic legal tasks. Intuitively LLMs must first obtain lower-level skills before excelling in higher-level skills. This division method provides a structured overview of the skill set required for legal-related tasks. It can also facilitates an exploration of the similarities
and differences between the learning mechanisms of LLMs and humans.

We extensively tested 51 popular LLMs,  including 20 multilingual LLMs, 22 Chinese-oriented LLMs and 9 legal specific LLMs. To effectively extract answers from the predictions of LLM generations, we design suitable rules, regular expressions and metrics for every individual task. The benchmark and evaluation code are integrated into the OpenCompass platform~\cite{2023opencompass} to enable easy reproduction. The evaluation results are shown in Figure \ref{fig:law_bench_radar_plot}. We find that although legal specific fine-tuning usually improves upon their base model, they are still significantly lagging behind general LLMs, which occupy the top six spots in the averaged zero-shot performance. We analyze the impact of various factors on the results, such as supervised fine-tuning (SFT), reinforcement learning from human feedback (RLHF)~\cite{christiano2017deep}, model size, legal specific fine-tuning, etc. Important suggestions are summarized to better guide the future development of legal LLMs for the Chinese community.

\section{Related Work}
\paragraph{Large Language Models}
Large language models (LLMs) trained on massive amounts of data have shown impressive abilities in generating high-quality, coherent text and following zero-shot or few-shot instructions in a diverse set of tasks~\cite{openai_chatgpt,openai2023gpt4,chowdhery2022palm,touvron2023llama1,touvron2023llama}. These models are usually trained following three steps: pre-training, supervised fine-tuning (SFT) and alignment with human or AI feedback ~\cite{christiano2017deep,shen2017estimation,wei2021finetuned,ouyang2022training,lee2023rlaif,casper2023open} etc. As most public available LLMs focus on training on English corpora, many efforts have been devoted to extending LLMs to Chinese. These works either pre-training a new LLM from scratch on Chinese-centric corpora~\cite{su2022welm,zeng2022glm,du2022glm,sun2023moss,2023internlm}, or expanding the vocabulary of an existing English-centric LLM then performing SFT on Chinese instruction data~\cite{chinese-llama-alpaca,BELLE,belle2023exploring}. 
There have also been studies that are dedicated to adapting LLMs to the legal domain by fine-tuning on legal specific corpora~\cite{cui2023chatlaw,huang2023lawyer,yue2023disclawllm}. However, a comprehensive evaluation to compare the existing LLMs regarding their legal knowledge is still lacking. Our focus is primarily on models that can complete corresponding tasks based on instructions, so we exclude pre-trained small language models such as Lawformer~\cite{xiao2021lawformer} and LegalBERT~\cite{chalkidis2020legal}, which require task-specific fine-tuning to perform competiviely.

\paragraph{Existing Benchmarks}

As the rapid development of LLMs, conventional approaches of evaluating a model's performance on a single task through fine-tuning~\cite{xu2020clue,wang2018glue,louis2022statutory,zhang2022mdia,shen2023xpqa} is no longer adequate for evaluating LLMs. A growing body of research works have recently focused on developing more comprehensive and systematic benchmarks to evaluate various capabilities of LLMs. Examples include AGIEval~\cite{zhong2023agieval} which covers human-centric standardized exams, such as college entrance exams, law school admission tests, math competitions, and lawyer qualification tests, HELM~\cite{liang2022holistic} which measures 7 metrics (accuracy, calibration, robustness,
fairness, bias, toxicity, and efficiency) for each of 16 core scenarios to the extent possible, KOLA~\cite{yu2023kola} which focuses on knowledge-oriented assessment under four-level taxonomy of knowledge-related abilities, and MMBench~\cite{liu2023mmbench} which is designed specifically to evaluate vision-language models. There have also been efforts in constructing benchmarks for Chinese such as CMMLU~\cite{li2023cmmlu}, GAOKAO~\cite{zhang2023evaluating} and C-Eval~\cite{huang2023c}. Some of them focus on specific domains such as CMB~\cite{wang2023cmb} in the medical domain and Fin-eval~\cite{zhang2023fineval} in the finance domain. In the legal domain, there have also been works ensembling legal-related tasks including the lexglue~\cite{chalkidis2022lexglue} focusing on EU and American laws, and LBOX OPEN~\cite{hwang2022multi} focusing on South Korean laws, but they did not formulate tasks into the instruction-following formats for LLMs. Recently, legal-bench was released in evaluating LLMs on 162 legal-related tasks based on American laws~\cite{guha2023legalbench}. Our work follows a similar line to extend the evaluation to Chinese laws. Given the uniqueness of the Law system in People's Republic of China and the diversity of the legal tasks covered in this work, we believe
that LawBench will contribute to the multilinguality of global legal research and promote research in specializing language models to a specific domain.

\section{LawBench Construction}
In this section, we provide a detailed description of the principles behind the design of LawBench and the test task selection.

\subsection{The Hierarchical Ability Taxonomy of LawBench}
\begin{figure}[t]
    \centering
    \includegraphics[width=\columnwidth]{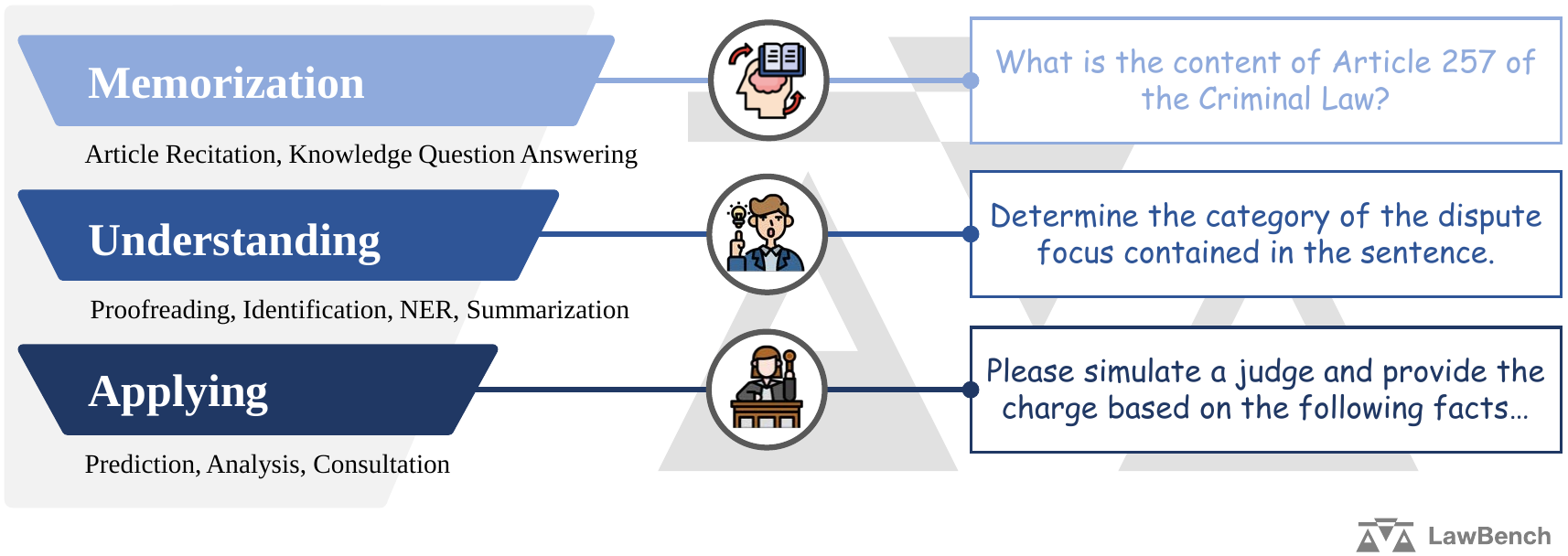}
    \caption{Three cognitive dimensions for evaluating large language models in LawBench. In order to specialize in legal tasks, LLMs must be able to (1) memorize necessary legal concepts, terminologies, articles and facts; (2) understand entities, events and relationships in legal text; and finally (3) simulate law professionals to apply legal knowledge and necessary reasoning in solving realistic tasks.}
    \label{fig:cognitive}
\end{figure}

When evaluating large language models, there is a preference for using a variety of tasks to assess their capabilities. A hierarchical evaluation system allows us to better understand the abilities of large language models in different aspects. Instead of categorizing tasks solely based on their difficulty~\cite{huang2023c}, we refer to the widely used Bloom's cognitive model~\cite{krathwohl2002revision} to classify tasks into different dimensions~\cite{yu2023kola}. Bloom's Taxonomy system was initially proposed by the educational psychologist Benjamin Bloom and his collaborators in 1956 and has been widely applied and developed in the following decades. It has effectively aided teachers in curriculum design and the assessment of student learning outcomes. Bloom's Taxonomy divides learning objectives in the cognitive domain into six levels, from the lowest to the highest: Remember, Understand, Apply, Analyze, Evaluate, and Create. These levels describe the depth and complexity of cognitive learning and provide an organized framework. Teachers can use Bloom's Taxonomy to ensure diversity and completeness in course objectives. By combining learning objectives at different levels, comprehensive student development can be promoted, encouraging them to progress from simple memorization and understanding to higher-level analysis, evaluation, and creation.

Inspired by this classification approach, we simplified Bloom's cognitive hierarchy model and kept the first three categories in Bloom's taxonomy to assess the legal knowledge of LLMs:

\begin{enumerate}
    \item \textbf{Knowledge Memorization}: The memorization level measures the basic requirement of remembering legal-related knowledge. It tests LLMs' ability in memorization and recitation of basic legal-domain knowledge such as regulations, cases, concepts, common sense, legal facts and terminologies.
    \item \textbf{Knowledge Understanding}: The understanding level involves understanding the meanings and connotations of legal documents. This includes the ability to comprehend and interpret legal concepts, text, and issues, e.g., identifying entities and relationships within legal texts, detect types of legal issues and points of dispute, among others.
    \item \textbf{Knowledge Applying}: The applying level requires LLMs to integrate legal knowledge, reason over it and address real-world legal cases. It covers the model's logical reasoning abilities to perform legal consultation, judicial assistance, as well as numerical reasoning abilities to calculate involved amount.
\end{enumerate}

Note that under this taxonomy, some tasks may not strictly belong to just one category; they may involve other abilities as well. We have categorized these tasks based on their primary capabilities.

\begin{table}[htbp]
  \centering
  \small
  \resizebox{\columnwidth}{!}{
    \begin{tabular}{llllcc}
    \toprule
    \textbf{Cognitive Level} & \textbf{ID} & \textbf{Task} & \textbf{Data Source} & \textbf{Metric} & \textbf{Type} \\
    \midrule
    \multirow{2}{*}{\shortstack[l]{\textbf{Legal Knowledge}\\ \textbf{Memorization}}} & 1-1 & Article Recitation & FLK & Rouge-L & Generation \\
          & 1-2 & Knowledge Question Answering & JEC\_QA & Accuracy & SLC \\\midrule
    \multirow{10}{*}{\shortstack[l]{\textbf{Legal Knowledge}\\ \textbf{Understanding}}} & 2-1 &  Document Proofreading & CAIL2022  & F0.5 & Generation\\
    & 2-2 & Dispute Focus Identification & LAIC2021  & F1 & MLC\\
          & 2-3 & Marital Disputes Identification & AIStudio & F1 & MLC\\
          & 2-4  & Issue Topic Identification & CrimeKgAssitant & Accuracy & SLC\\
        & 2-5 & Reading Comprehension & CAIL2019  & rc-F1 & Extraction\\
          & 2-6 & Named-Entity Recognition & CAIL2022  & soft-F1 & Extraction\\
          & 2-7 & Opinion Summarization & CAIL2021  & Rouge-L & Generation\\
          & 2-8 & Argument Mining  & CAIL2022  & Accuracy &SLC\\
          & 2-9 & Event Detection  & LEVEN  & F1 &MLC\\
          & 2-10 & Trigger Word Extraction & LEVEN  & soft-F1 & Extraction\\
          \midrule
    \multirow{8}{*}{\shortstack[l]{\textbf{Legal Knowledge}\\ \textbf{Applying}}} & 3-1 & Fact-based Article Prediction & CAIL2018 & F1 & MLC\\
          & 3-2 & Scene-based Article Prediction & LawGPT & Rouge-L &Generation\\
          & 3-3 & Charge Prediction & CAIL2018 & F1 &MLC\\
          & 3-4 & Prison Term Prediction w.o. Article & CAIL2018 & nLog-distance &Regression\\
          & 3-5 & Prison Term Prediction w. Article & CAIL2018 & nLog-distance &Regression\\
          & 3-6 & Case Analysis & JEC\_QA & Accuracy &SLC\\
          & 3-7 & Criminal Damages Calculation & LAIC2021  & Accuracy &Regression \\
          & 3-8 & Consultation & hualv.com & Rouge-L &Generation\\
    \bottomrule
    \end{tabular}
    }
    \caption{\small Task list in LawBench. Tasks correspond to cognitive dimensions: legal knowledge memorization, understanding and applying, and 5 task types: generation, single-label classification (SLC), multi-label classification (MLC), regression, and extraction.}
  \label{tab:tasks}%
\end{table}%

\subsection{Data Source and Selected Tasks}
We selected 20 tasks falling under the above-mentioned capability levels. Every task is assigned a unique task id for better distinction. The task list is provided in Table~\ref{tab:tasks}. There can exist datasets belonging to the same tasks. When selecting the dataset for every task, we choose the most recent available version. Furthermore, certain tasks like legal case retrieval requires processing very long documents, which can surpass the length limit for most LLMs, so we do not include them to LawBench for now. When constructing LawBench, we have made efforts to format the prompts in a way that best aligns with user habits with clear instructions about the answer format, so that we can assess the ability of LLMs in assisting legal tasks in realistic scenarios.

\textbf{Legal Knowledge Memorization Tasks}

Legal knowledge memorization tasks examine to which extent large language models encode legal knowledge within their parameters. While this knowledge can be enhanced with external retrievers, it is still beneficial to memorize necessary legal knowledge because (1) There is currently no reliable mechanism to guarantee the accurate retrieval of legal provisions. Memorizing useful knowledge within model parameters can help combat the noise from external retrievers~\cite{xie2023adaptive,neeman2023disentqa}; (2) It is very difficult, if not impossible, to retrieve all needed legal knowledge for complicated reasoning tasks. The model must know basic legal concepts to connect the retrieved knowledge smoothly~\cite{shen2022product,ziems2023large,yu2023generate}; (3) Relying on the parametric knowledge instead of external retrievers can significantly reduce the online latency~\cite{shen2022semipqa,shen2022low,tay2022transformer}. 

There are two major types of legal knowledge that requires memorizing: (1) core law articles and regulation content and (2) other fundamental legal concepts, notions and rules. We construct two tasks corresponding to these two types of knowledge: 
\begin{itemize}
    \item \textbf{Article recitation} (1-1): \emph{Given a law article number, recite the article content}. We collected the contents of laws and regulations from the national database~\footnote{\url{https://flk.npc.gov.cn/}} and consulted students with a legal background to select 152 sub-laws under the 5 core laws.
    We further incorporated updated laws and regulations, including constitutional amendments, to evaluate the model's ability to comprehend legal changes.
    \item \textbf{Knowledge question answering} (1-2): \emph{Given a question asking about basic legal knowledge, select the correct answer from 4 candidates}. We collect knowledge-based questions from the JEC-QA tasks~\cite{zhong2020jec}. To simplify the process of locating answers during the test, we exclusively chose single-label questions from them.
\end{itemize}
Examples of these two tasks are in Appendix~\ref{app:instruction_memorization}.

\textbf{Legal Knowledge Understanding Tasks}

Legal knowledge understanding tasks examine to which extent large language models can comprehend entities, events, and relationships within legal texts. Understanding legal text is a pre-condition to utilize the knowledge in concrete downstream applications~\cite{cui2022survey}. In total, we selected 10 tasks corresponding to different levels of legal knowledge understanding:

\begin{itemize}
    \item \textbf{Document Proofreading} (2-1): \emph{Given a sentence extracted from legal documents, correct its spelling, grammar and ordering mistakes, return the corrected sentence}. Legal documents, as carriers of judicial authorities and the exercise of legal rights by citizens, demand utmost precision in their textual content. We sample the original and corrected legal sentences from the CAIL2022 document proofreading task. Possible mistake types are inserted into the instructions to let the model directly output the corrected sentence.
    \item \textbf{Dispute Focus Identification} (2-2): \emph{Given the original claims and responses of the plaintiff and defendant, detect the points of dispute}. In civil cases, the points of dispute represent the core of conflicts, intersection of contradictions, and issues over which the parties involved in the case are in contention. The automated recognition and detection of points of contention have practical significance and necessity for the development of the rule of law in our country. Specifically, we will provide the trial-related content from judgment documents, including the sections on claims and responses. The cases involve various legal matters such as civil loans, divorce, motor vehicle traffic accident liability, financial loan contracts, and more. We have carefully selected common types of points of contention from LAIC2021 to construct this test set.
    \item \textbf{Marital Disputes Identification} (2-3): \emph{Given a sentence describing marital disputes, classify it into one of the 20 pre-defined dispute types}. Marital disputes refer to the total sum of various disputes arising from love, marriage, and divorce. Among civil disputes, marital disputes are a common type of dispute.  We have selected a publicly available marriage text classification dataset on AiStudio~\footnote{\url{https://aistudio.baidu.com/datasetdetail/181754}}. This dataset consists of 20 categories, and a single text entry may have multiple labels.
    \item \textbf{Issue Topic Identification} (2-4): \emph{Given a user inquiry, assign it into one of pre-defined topics}. User inquiries are typically vague. Identifying the relevant topics in legal consultations can help legal professionals better pinpoint key issues. We obtain the data from the CrimeKgAssistant project~\footnote{\url{https://github.com/liuhuanyong/CrimeKgAssitant}}. We keep the most frequent 20 classes and sample 25 questions for each class to form our final test set.
    \item \textbf{Reading Comprehension} (2-5): \emph{Given a judgement document and a corresponding question, extract relevant content from it to answer the question}. Judicial documents contain rich case information, such as time, location, and character relationships. Intelligently reading and comprehending judicial documents through large language models can assist judges, lawyers, and the general public in obtaining the necessary information quickly and conveniently. We use the CAIL2019 reading comprehension dataset to build this task, removing question types related to binary and unanswerable questions. We retain single and multiple-segment data as our test set.

    \item \textbf{Named-Entity Recognition} (2-6): \emph{Given a sentence from a judgement document, extract entity information corresponding to a set of pre-defined entity types such as suspect, victim or evidence}. We sampled 500 examples from the CAIL2022 Information Extraction dataset as our test set. These 500 samples contain 10 entity types related to theft crimes.

    \item \textbf{Opinion Summarization} (2-7): \emph{Given a legal-related public news report, generate a concise summary}. Legal summaries typically include key facts of the case, points of contention, legal issues, legal principles applied, and the judgment's outcome. It can provide a quick overview of the case content to improve the efficiency of legal professionals. 
We randomly select 500 samples from the CAIL2021 Legal Public Opinion Summary dataset for this task. We only select samples with less than 400 words to fit the length constraint of LLMs.

    \item \textbf{Argument Mining} (2-8): \emph{Given a plaintiff's perspective and five candidate defendant's viewpoints, select one viewpoint that can form a point of dispute with the plaintiff's perspective}. In court's trial process, judgment documents play a crucial role in recording the arguments and evidence presented by both the plaintiff and the defendant. Due to differences in their positions and perspectives, as well as inconsistencies in their factual statements, disputes arise between the plaintiff and the defendant during the trial process. These points of contention are the key to the entire trial and the essence of judgment documents. This task aims to extract valuable arguments and supporting materials from a large volume of legal texts, providing strong support for legal debates and case analysis. We use CAIL2022's Argument Mining dataset to construct our dataset, transforming the identification of focal points of disputes into a multiple-choice question format.
    \item \textbf{Event Detection} (2-9):
\emph{Given a sentence from a legal judgement document, detect which events are mentioned in this sentence}. Events are the essence of facts in legal cases. Therefore, Legal Event Detection is fundamentally important and naturally beneficial to case understanding and other Legal AI tasks. We construct the test set from the LEVEN dataset\cite{yao2022leven} by sampling sentences corresponding to the top 20 most frequent event types. Multiple events can be mentioned in every sentence.
    \item \textbf{Trigger Word Extraction} (2-10):
\emph{Given a sentence from a legal judgment document and its corresponding events, predict which words in the sentence triggered these events}. Trigger words directly cause events and are an important feature that determines the event category, providing post-hoc explanation for the event types we identify. Directly identifying trigger words is very difficult, so we simplified this task by providing the events contained in the text along with the text information,  examining the ability of LLMs to recognize trigger words related to events. When constructing the trigger word test set, we removed trigger words that were the same as the event type, as well as events with multiple or duplicate trigger words from the LEVEN dataset\cite{yao2022leven}, to include as different trigger words as possible.
\end{itemize}
Examples of the 10 understanding tasks are in Appendix~\ref{app:instruction_understanding}.

\textbf{Legal Knowledge Applying Tasks}

Legal knowledge applying tasks primarily examine the ability of LLMs to not only understand legal knowledge but also simulate law professionals to apply the knowledge in solving realistic legal tasks. In the task design, we extensively examine the model's different reasoning abilities, including 3 legal content reasoning tasks: legal judgement prediction, case analysis, consultation, and 1 numerical reasoning task: criminal damages calculation. 

When predicting case judgments, judges follow a certain order when hearing a case~\cite{zhong2018legal,huang2021dependency}. Therefore, in constructing the case judgment prediction task, we simulated this process by decomposing the CAIL2018 dataset into three tasks: fact-based article prediction (3-1), charge prediction (3-3) and prison term prediction. We further separate the task of prison term prediction into two scenarios: without article content (3-4) and with article content (3-5) to examine LLMs' capability in utilizing the article content to make accurate judgement predictions. Besides, we also add the task scene-based fact prediction  to simulate judges’
recognition of legal provisions (3-2).

\begin{itemize}
    \item \textbf{Fact-based Article Prediction} (3-1): \emph{Given a fact statement from the legal judgement document, predict which article items should be applied}.
When judges make decisions, they usually associate relevant articles with the facts of the case~\cite{ge2021learning,louis2023finding}. Article prediction can assist judges in quickly locating legal articles related to legal texts. Legal articles are written expressions of legal norms, which are rules and regulations with clear meanings and legal effects. 
The model needs to deduce potentially applicable legal provisions based on the given case description and related background information. We sample 500 cases from the CAIL2018 dataset for this task.

    \item \textbf{Scene-based Article Prediction} (3-2): \emph{Given a described scenario and a related question, predict the corresponding article item}.
The CAIL2018 dataset only covers criminal law-related legal provisions. In order to comprehensively evaluate the ability of LLMs to analyze case facts and infer relevant legal provisions, we collected high-quality legal scenario-based question-and-answer data from public sources on GitHub\cite{LAWGPT-zh}. This dataset was generated by inputting legal provisions into chatGPT to construct corresponding scenario-based questions and answers. We manually selected 5,000 question-and-answer pairs with accurate answers from the generated dataset. Based on this, we selected 252 core legal provisions' scenario-based question-and-answer content as the test dataset.

    \item \textbf{Charge Prediction} (3-3): \emph{Given fact statement from the legal judgement document and the applied article number, predict the cause of action (charge)}.
Cause of action is a summary of the nature of the legal relationship involved in a litigation case, as formulated by the people's court. Accurately predicting the cause of action can help improve judicial efficiency and fairness. In the process of filing and hearing cases, accurate prediction of the cause of action can help the court to allocate cases, allocate resources, and arrange trials, thereby improving judicial efficiency and fairness. 
We sampled 500 pieces of data from the CAIL2018 cause of action prediction dataset for this task.
    \item \textbf{Prison Term Prediction w.o. Article} (3-4): \emph{Given fact statement from the legal judgement document, the applied article number and charge, predict the prison term.}
Prison term prediction refers to the process of predicting and estimating the possible sentence that a defendant may face during the criminal justice process based on the facts of the case, legal provisions, and relevant guiding precedents. It aims to make reasonable inferences about the length and form of the sentence by comprehensively considering various factors such as the nature of the crime, the circumstances of the offense, the social impact, and the defendant's personal situation. 
We used the prison term prediction dataset from CAIL2018, removed some cases with the death penalty and life imprisonment, and randomly sampled 500 cases as the test dataset. During the process of judges' sentencing, more information is usually taken into account to determine the prison term outcome. We simulated the judge's analysis process by providing the relevant legal provisions and the charge of the case.
    \item \textbf{Prison Term Prediction w. Article} (3-5):
\emph{Given fact statement from the legal judgement document, the applied article content and charge, predict the prison term.}
Large language models typically use retrieval-argument methods to introduce new information. Some publicly available models also include retrieval modules that provide detailed reference information for the model by retrieving legal provisions. We simulated this process, and unlike the previous task where only the legal provision number was provided, we provided the specific content of the legal provision in this task. When constructing the sentence prediction dataset, we appended the content of the legal provisions to the end of the question, allowing the model to complete the sentence prediction task in this scenario.
    \item \textbf{Case Analysis} (3-6): \emph{Given a case and a corresponding question, select the correct answer from 4 candidates}. We use the case analysis part from JEC\_QA dataset~\cite{zhong2020jec} for this task. The case analysis part tests the ability of models to analyze real cases. Models must possess five types of reasoning in order to perform this analysis including word matching, concept understanding, numerical analysis, multi-paragraph reading, and multi-hop reasoning. In order to reduce the difficulty of the test and facilitate the acquisition of answers, we sampled 500 multiple-choice questions from the JEC\_QA Case-Analysis part as the testing dataset.
    \item \textbf{Criminal Damages Calculation} (3-7): \emph{Given a fact description about a criminal process, predict the amount of money involved in this case}.
There are some numerical computing tasks in the process of judicial trials, such as the calculation of the total amount of legal crimes. The total amount of the crime is an important sentencing factor. In some charges such as theft, financial fraud, and bribery, China's laws determine the severity of the sentence based on the amount involved in the case. This task mainly tests the computing ability of LLMs. First, we examine whether the model understands the rules of case amount calculation, and second, we examine whether the model can accurately complete numerical calculations. We selected the LAIC2021 numerical computing task to construct our dataset.
    \item \textbf{Consultation} (3-8):
\emph{Given a user consultation, generate a suitable answer.}
Legal consultation is a way for the public to access legal services. Legal consultation can help people understand legal disputes and seek targeted advice and solutions from professional lawyers, as well as receive support and guidance. Some law firms and legal consulting companies also provide online legal consultation services, making it more convenient for people to obtain legal help. 
We collected legal consultation contents from the Hualv website~\footnote{\url{www.66law.com}}, and our dataset contains both the answers to legal consultations and the corresponding legal basis, i.e., legal articles.
\end{itemize}

Examples of the 8 applying tasks are in Appendix~\ref{app:instruction_application}.

\subsection{Evaluation}
For every task,  the evaluation is done following two steps: (1) answer extraction, which extracts the answer from the model prediction, and (2) metric computation, which computes the metric score based on the question, extracted answer and the gold answer. Answer extraction is a necessary step since many LLMs often do not generate output directly comparable with gold labels~\cite{adlakha2023evaluating}. We explain these two steps in detail in the following section.
\paragraph{Answer Extraction}
Most of the tasks require the prediction to be in the standard format in order to compare with the ground truth, we define a set of task-specific rules to extract the answer from the model prediction.
\begin{itemize}
    \item Article Number Extraction (3-1): this type of tasks requires us to extract the article numbers predicted by the model. To do this, we use the delimiter \begin{CJK*}{UTF8}{gbsn} ``、'' \end{CJK*} to separate the prediction text into chunks of text, and then the cn2an\footnote{\url{https://github.com/Ailln/cn2an}} library is used to convert the Chinese numerals to Arabic numerals within each of those chunks. Using a regular expression, we extract the converted Arabic numerals as the expected article numbers; if more than one number appears in the same chunk, only the first number is extracted. All extracted numbers are combined to form the final set of predictions.
    \item Prison Term Extraction (3-4, 3-5): for this type of tasks, we need to extract the predicted prison terms from the prediction text. To begin, we use cn2an to convert all the Chinese numerals in the prediction to Arabic numerals; we then extract digits that are followed by time intervals in Chinese, such as \begin{CJK*}{UTF8}{gbsn} ``个月'' (month) and ``年'' (year)\end{CJK*}. The extracted prison terms are normalized to months, meaning that the numbers appearing before \begin{CJK*}{UTF8}{gbsn} ``年'' \end{CJK*} will be multiplied by 12. Note that the time unit in the ground truth answer is also month.
    \item Criminal Damages Extraction (3-7): We extract all numbers appearing in the prediction text using regular expression. The set of of the extracted numbers is considered as the predicted criminal damages.
    \item Named-Entity Recognition (2-6): We find all occurrences of entity types from the model prediction, then obtain the substring from its occurrence to the delimiter ``\textbackslash n'', then apply a regular expression to extract the entity value.
    \item Trigger Word Extraction (2-10): We split the model prediction by the delimiter \begin{CJK*}{UTF8}{gbsn} ``；'' \end{CJK*}, then treat the split array as a list of extracted key words.
    \item Option Extraction (1-2, 2-2, 2-3, 2-4, 2-8, 2-9, 3-3): this type of task is similar to selecting the correct options from a list of options in a multiple-choice task. We run through all possible options and check if they appear in the prediction text. The set of options that occur in the prediction text is collected and used for evaluation.
    \item Others (1-1, 2-1, 2-5, 2-7, 3-2, 3-8): we take the model prediction as the answer without performing any extraction step. 
    
\end{itemize}

\paragraph{Metrics}
After the answer extraction phase, we compute the final metric based on the extracted answer. We defined 7 different metrics in total to measure different types of tasks:
\begin{itemize}
    \item \textbf{Accuracy}: Accuracy is a binary score that performs exact match between the model prediction and the gold answer. This applies to all single-label classification tasks including task 1-2, 2-4, 2-8, 3-6, and the regression task 3-7. For SLC tasks, if multiple valid answers are extracted from the model prediction, then we always treat it as wrong~\footnote{For the criminal damages calculation task, we treat the model prediction correct as long as one of the extracted answers match the ground truth as we find LLMs often output the whole calculation process.}.
    \item \textbf{F1}: When there are multiple output labels, F1 score measures the harmonic mean of the precision and recall. This applies to all multi-label classification tasks including task 2-2, 2-3, 2-9, 3-1 and 3-3.
    \item \textbf{rc-F1}: rc-F1 is the F1 score tailored for the reading comprehension task 2-5. It treats every token as a label, removes punctuation, stories, extra whitespace, performs other necessary normalizations then compute the F1 score. We adopt the official script from CAIL2019 to compute the instance-level rc-F1 score~\footnote{\url{https://github.com/china-ai-law-challenge/CAIL2019/tree/master}}.
    \item \textbf{soft-F1}: For extraction tasks 2-6 and 2-10, the output is a set of phrases. Instead of using the standard F1 score, we use a soft version by replacing the phrase-level exact match with the rc-F1 score, then computing the F1 on top of it. We find using the soft version helpful since LLMs often use wording choices different from the ground truth.
    \item \textbf{nLog-distance}: For the prison term prediction tasks 3-4 and 3-5, we evaluate them with the normalized log distance (nLog-distance) to capture the continuity of prison terms. We compute the logarithm of the difference between the extracted and gold answer, then normalize it to the space between 0 and 1 for better compatibility with other metrics.
    \item \textbf{F0.5}: For the document proofreading task 2-1, we use the F0.5 metric to evaluate it. The F0. 5 score gives more weight to precision than to recall we want to prevent introducing more false positives than identify every other error in proofreading~\cite{zhang2022mucgec}. We use the ChERRANT toolkit to align the extracted and gold answer before computing the F0.5 score~\footnote{\url{https://github.com/HillZhang1999/MuCGEC/tree/main/scorers/ChERRANT}}. As the alignment can take too long to respond for very bad generations, we add a time-out of 10 seconds. If a time-out happened, then the prediction is assigned a score of 0.
    \item \textbf{Rouge-L}: For other generation tasks 1-1, 2-7, 3-3 and 3-8, we use the Rouge-L score to evaluate them. Rouge-L is a commonly used metric in generation tasks. It takes into account sentence-level structure similarity naturally and identifies longest co-occurring in sequence n-grams automatically to compare the extracted and gold answers~\cite{lin2004rouge}.
\end{itemize}
Several large language models may decline to respond to legal-related inquiries due to security policies or simply fail to follow the instructions. To capture this issue, we also report the \textbf{abstention rate} of LLMs in each task (how often an LLM abstains to answer). An abstention happens if an answer cannot be extracted from the model prediction. The abstention rate does not apply to task 2-5 and all generation tasks since they do not need the answer extraction step.

\section{Experiment}
\begin{table*}[t]
    \centering
    \resizebox{13.5cm}{!}{
    \begin{tabular}{l|cccccc}
        \toprule
        \textbf{Model} & \textbf{Parameters} & \textbf{SFT} & \textbf{RLHF} &\textbf{Access} & \textbf{BaseModel} &\\ 
        \midrule
        \textbf{Multilingual LLMs} \\
        MPT & 7B & \xmark & \xmark & Weights & - \\
        MPT-Instruct & 7B & \cmark & \xmark & Weights & MPT-7B \\
        LLaMA & 7/13/30/65B & \xmark & \xmark & Weights & - \\
        LLaMA-2 & 7/13/70B & \cmark & \xmark & Weights & - \\
        LLaMA-2-Chat & 7/13/70B & \cmark & \cmark & Weights & LLaMA-2-7/13/70B \\
        Alpaca-v1.0 & 7B & \cmark & \xmark & Weights & LLaMA-7B \\
        Vicuna-v1.3 & 7/13/33B & \cmark & \xmark & Weights & LLaMA-7/13/33B \\
        WizardLM & 7B & \cmark & \xmark & Weights & LLaMA-7B \\
        StableBeluga2 & 70B & \cmark & \xmark & Weights & LLaMA-2-70B \\
        ChatGPT & N/A & \cmark & \cmark & API & - \\
        GPT-4 & N/A & \cmark & \cmark & API & - \\
        \midrule
        \textbf{Chinese-oriented LLMs} \\
        MOSS-Moon & 16B & \xmark & \xmark & Weights & - \\
        MOSS-Moon-SFT & 16B & \cmark & \xmark & Weights & MOSS-Moon \\
        TigerBot-Base & 7B & \xmark & \xmark & Weights & - \\
        TigerBot-SFT & 7B & \cmark & \xmark & Weights & TigerBot-Base \\
        GoGPT & 7B & \cmark & \xmark & Weights & LLaMA-7B \\
        ChatGLM2 & 6B & \cmark & \xmark & Weights & ChatGLM \\
        Ziya-LLaMA & 13B & \cmark & \cmark & Weights & LLaMA-13B \\
        Baichuan & 7/13B & \xmark & \xmark & Weights & - \\
        Baichuan-13B-Chat & 13B & \cmark & \xmark & Weights & Baichuan-13B \\
        XVERSE & 13B & \xmark & \xmark & Weights & - \\
        InternLM & 7B & \xmark & \xmark & Weights & - \\
        InternLM-Chat & 7B & \cmark & \xmark & Weights & InternLM-7B \\
        InternLM-Chat-8K & 7B & \cmark & \xmark & Weights & InternLM-7B \\
        Qwen & 7B & \xmark & \xmark & Weights & - \\
        Qwen-Chat & 7B & \cmark & \xmark & Weights & Qwen-7B \\
        Yulan-Chat-2 & 13B & \cmark & \xmark & Weights & LLaMA-2-13B \\
        BELLE-LLaMA-2 & 13B & \cmark & \xmark & Weights & LLaMA-2-13B \\
        Chinese-LLaMA-2 & 7B & \cmark & \xmark & Weights & LLaMA-2-7B \\
        Chinese-Alpaca-2 & 7B & \cmark & \xmark & Weights & LLaMA-2-7B \\
        LLaMA-2-Chinese & 7/13B & \cmark & \xmark & Weights & LLaMA-2-7/13B \\
        \midrule
        \textbf{Legal Specific LLMs} \\
        HanFei & 7B & \cmark & \xmark & Weights & HanFei \\
        LaWGPT-7B-beta1.0 & 7B & \cmark & \xmark & Weights & Chinese-LLaMA \\
        LaWGPT-7B-beta1.1 & 7B & \cmark & \xmark & Weights & Chinese-alpaca-plus-7B \\
        LexiLaw & 6B & \cmark & \xmark & Weights & ChatGLM-6B \\
        Wisdom-Interrogatory & 7B & \cmark & \xmark & Weights & Baichuan-7B \\
        Fuzi-Mingcha & 6B & \cmark & \xmark & Weights & ChatGLM-6B \\
        Lawyer-LLaMA & 13B & \cmark & \xmark & Weights & LLaMA \\
        ChatLaw & 13/33B & \cmark & \xmark & Weights & Ziya-LLaMA-13B/Anima-33B\\
        \bottomrule
       
    \end{tabular}}
    \caption{\small LLMs tested on LawBench. We classify these models by their main training corpora.}
    \label{tab:models}
\end{table*}

\subsection{Models}
We evaluate a wide spectrum of large language models of various sizes, grouping them into three major categories based on their pre-training and fine-tuning domains: multilingual LLMs, Chinese-oriented LLMs and legal specific LLMs. We provide a short review over them in the following section. The detailed model list is shown in Table~\ref{tab:models}.

\paragraph{Multilingual LLMs} We consider 18 open-source multilingual models: MPT-7B, MPT-Instruct-7B, LLaMA-7B / 13B / 30B / 65B, LLaMA-2-7B / 13B / 70B, LLaMA-2-Chat-7B / 13B / 70B, Alpaca-v1.0-7B, Vicuna-v1.3-7B / 13B / 33B, WizardLM-7B, StableBeluga2. In addition, two commercial models, ChatGPT and GPT-4, are included.

\paragraph{Chinese Oriented LLMs} A number of Chinese-oriented LLMs are proposed to enhance Chinese comprehension. Their typically perform better than multilingual models on Chinese NLP tasks. We include 21 open-sourced, Chinese-oriented LLMs in our evaluation: CMOSS-Moon / Moon-SFT, TigerBot-Base / SFT, GoGPT, ChatGLM2-6B, Ziya-LLaMA-13B, Baichuan-7B / 13B / 13B-Chat, XVERSE-13B, InternLM-7B / Chat-7B / Chat-7B-8K, Qwen-7B / 7B-Chat, BELLE-LLaMA-2, Chinese-LLaMA-2-7B, Chinese-Alpaca-2-7B, LLaMA-2-Chinese-7B / 13B.

\paragraph{Legal Specific LLMs} Certain Chinese-oriented LLMs are further fine-tuned on Chinese corpus in legal domain to improve LLMs' understanding of Chinese laws. They are of particular interest to us; through our benchmark, we can rigorously gauge their true advance compared to general-purpose LLMs and identify their limitations. Here, we provide detailed descriptions of these models: 
\begin{itemize}
    \item ChatLaw\cite{cui2023chatlaw}: ChatLaw-13B is fine-tuned based on Ziya-LLaMA-13B-v1, ChatLaw-33B is fine-tuned based on Anima-33B.
    \item LaywerLLaMA\cite{huang2023lawyer}: based on Chinese-LLaMA-13B, fine-tuned with general and legal instructions.
    \item FuziMingcha\cite{fuzi.mingcha}: based on ChatGLM, fine-tuned with unsupervised Chinese judicial texts along with supervised legal fine-tuning datasets.
    \item HanFei\cite{HanFei2023}: a fully pre-trained and fine-tuned legal model with 7 billion parameters.
    \item LaWGPT\cite{LaWGPT2023}: LaWGPT-7B-beta1.0 is further pre-trained on 500k Chinese judgment documents upon Chinese-LLaMA-7B and fine-tuned based on the Legal-Base-7B with instructions. LaWGPT-7B-beta1.1 is fine-tuned based on the Chinese-alpaca-plus-7B with 350k legal Q\&A datasets.
    \item LexiLaw\cite{LexiLaw2023}: a fine-tuned Chinese legal model based on the ChatGLM-6B with legal datasets. 
    \item WisdomInterrogatory\cite{WisdomInterrogatory2023}: a further pre-trained and fine-tuned model built upon Baichuan-7B.
\end{itemize}

\subsection{Experiment Setting}
We employ OpenCompass~\cite{2023opencompass} to perform model inference. For both ChatGPT and GPT-4, we set the temperature at 0.7 and top $p$ to 1. For other chat models, we tailor the prompt using prefixes and suffixes specific to each model. Greedy decoding is performed during generation for all open-sourced models. We set the input token length limit to 2048 and an output token length to 1024. Right truncation is performed for input prompts exceeding the length limitation. We evaluate all models in both zero-shot and one-shot settings. The model input in zero-shot inference is merely the task instruction and the query (see Appendix \ref{appendix:details_of_task_intruction} for individual instructions and queries). To build the model input for one-shot inference, a single example including the query and corresponding answer is attached after the instruction, followed by the actual query to the model.

\subsection{Main Results}
\begin{figure}[t]
    \centering
    \includegraphics[width=\columnwidth]{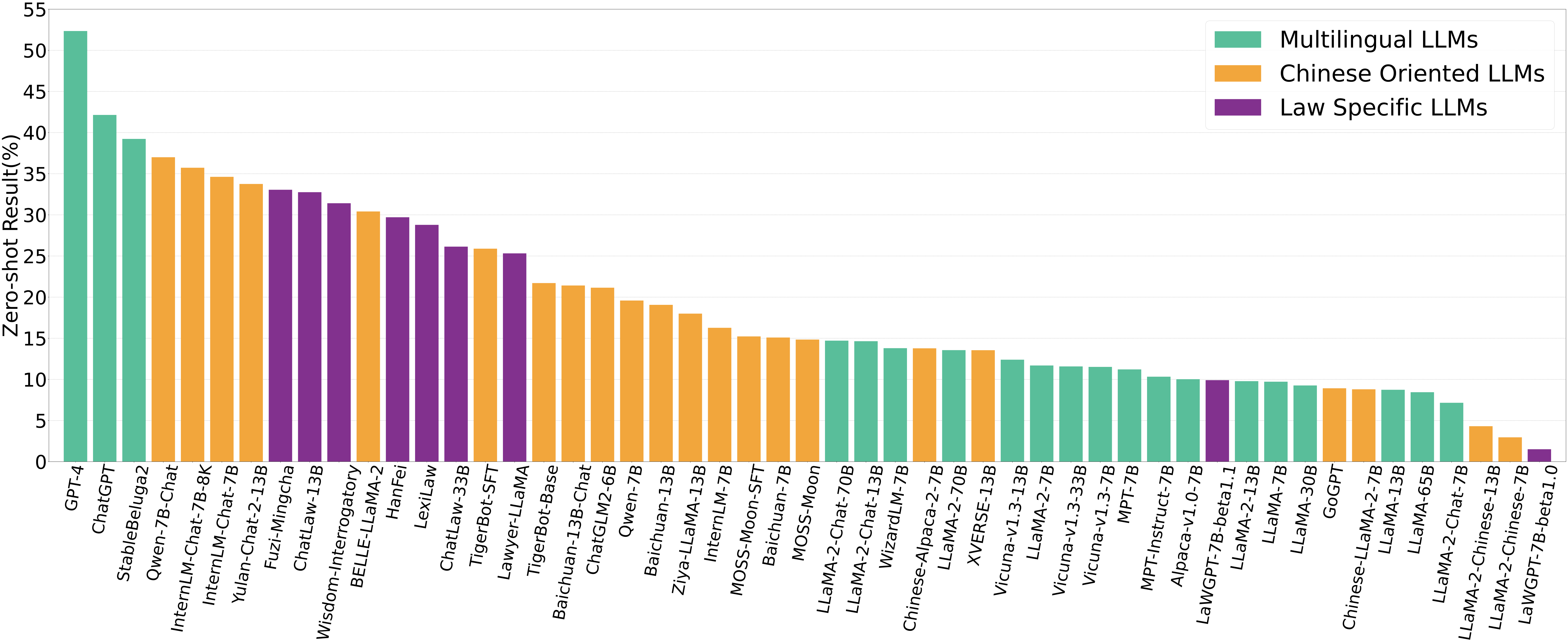}
    \caption{\small Average performance (zero-shot) of 51 LLMs evaluated on LawBench.}
    \label{fig:overall_performance_one_shot_main_content}
\end{figure}

\begin{table*}
\centering
\small
 \resizebox{\columnwidth}{!}{
\begin{tabular}{l|cccc|cccc|cccccccc}
\toprule
\multicolumn{1}{l|}{\multirow{3}*{Task}} & \multicolumn{4}{c}{Multilingual LLMs} & \multicolumn{4}{c}{Chinese Oriented LLMs} & \multicolumn{4}{c}{Legal Specific LLMs} \\
\multicolumn{1}{l|}{} & \multicolumn{2}{c}{GPT-4} & \multicolumn{2}{c}{ChatGPT} & \multicolumn{2}{c}{Qwen-Chat}  & \multicolumn{2}{c}{InternLM-Chat-8K} & \multicolumn{2}{c}{Fuzi-Mingcha} & \multicolumn{2}{c}{ChatLaw-13B} \\
\multicolumn{1}{l|}{} & \multicolumn{1}{c}{0-shot} & \multicolumn{1}{c}{1-shot} & 0-shot & 1-shot & \multicolumn{1}{c}{0-shot} & \multicolumn{1}{c}{1-shot} & \multicolumn{1}{c}{0-shot} & \multicolumn{1}{c|}{1-shot} & \multicolumn{1}{c}{0-shot} & \multicolumn{1}{c}{1-shot} & \multicolumn{1}{c}{0-shot} & \multicolumn{1}{c}{1-shot} \\ 
\midrule
1-1 & \cellcolor{Green!0.0}15.38 & \cellcolor{Green!0.0}17.21 & \cellcolor{Green!0.0}15.86 & \cellcolor{Green!0.0}16.15 & \cellcolor{Green!0.0}18.54 & \cellcolor{Green!0.0}17.73 & \cellcolor{Green!0.0}15.45 & \cellcolor{Green!0.0}15.16 & \cellcolor{Green!0.0}25.22 & \cellcolor{Green!0.0}20.21 & \cellcolor{Green!0.0}14.85 & \cellcolor{Green!0.0}15.98 \\
1-2 & \cellcolor{Green!0.2}55.2 & \cellcolor{Green!0.2}54.8 & \cellcolor{Green!0.0}36.0 & \cellcolor{Green!0.0}37.2 & \cellcolor{Green!0.2}34.0 & \cellcolor{Green!0.0}28.6 & \cellcolor{Green!0.0}40.4 & \cellcolor{Green!0.2}40.6 & \cellcolor{Green!11.4}7.8 & \cellcolor{Green!5.8}12.8 & \cellcolor{Green!0.4}28.4 & \cellcolor{Green!0.0}29.4 \\
\hline
2-1 & \cellcolor{Green!0.0}12.53 & \cellcolor{Green!0.0}18.31 & \cellcolor{Green!0.0}9.1 & \cellcolor{Green!0.0}13.5 & \cellcolor{Green!0.0}22.56 & \cellcolor{Green!0.0}25.16 & \cellcolor{Green!0.0}22.64 & \cellcolor{Green!0.0}21.64 & \cellcolor{Green!0.0}4.93 & \cellcolor{Green!0.0}2.86 & \cellcolor{Green!0.0}12.22 & \cellcolor{Green!0.0}13.01 \\
2-2 & \cellcolor{Green!0.0}41.65 & \cellcolor{Green!0.0}46.0 & \cellcolor{Green!3.0}32.37 & \cellcolor{Green!2.2}40.6 & \cellcolor{Green!28.4}27.42 & \cellcolor{Green!38.4}27.4 & \cellcolor{Green!6.8}35.46 & \cellcolor{Green!10.0}36.6 & \cellcolor{Green!23.4}19.59 & \cellcolor{Green!21.6}2.4 & \cellcolor{Green!11.6}2.68 & \cellcolor{Green!17.0}9.0 \\
2-3 & \cellcolor{Green!0.0}69.79 & \cellcolor{Green!0.0}69.99 & \cellcolor{Green!1.0}51.73 & \cellcolor{Green!0.4}54.01 & \cellcolor{Green!19.4}31.42 & \cellcolor{Green!21.8}32.96 & \cellcolor{Green!14.2}28.96 & \cellcolor{Green!16.6}30.91 & \cellcolor{Green!27.2}28.46 & \cellcolor{Green!42.4}17.44 & \cellcolor{Green!27.0}42.24 & \cellcolor{Green!47.6}30.91 \\
2-4 & \cellcolor{Green!1.6}44.0 & \cellcolor{Green!3.0}44.4 & \cellcolor{Green!3.4}41.2 & \cellcolor{Green!2.8}41.4 & \cellcolor{Green!13.0}35.0 & \cellcolor{Green!4.0}31.2 & \cellcolor{Green!7.8}35.6 & \cellcolor{Green!6.0}33.2 & \cellcolor{Green!29.4}18.6 & \cellcolor{Green!38.6}8.8 & \cellcolor{Green!15.8}27.6 & \cellcolor{Green!12.6}26.6 \\
2-5 & \cellcolor{Green!0.0}56.5 & \cellcolor{Green!0.0}64.8 & \cellcolor{Green!0.0}53.75 & \cellcolor{Green!0.0}61.98 & \cellcolor{Green!0.0}48.48 & \cellcolor{Green!0.0}46.71 & \cellcolor{Green!0.0}54.13 & \cellcolor{Green!0.0}54.35 & \cellcolor{Green!0.0}97.59 & \cellcolor{Green!0.0}93.35 & \cellcolor{Green!0.0}39.11 & \cellcolor{Green!0.0}41.41 \\
2-6 & \cellcolor{Green!0.0}76.6 & \cellcolor{Green!0.0}79.96 & \cellcolor{Green!0.0}69.55 & \cellcolor{Green!0.0}74.04 & \cellcolor{Green!0.0}37.88 & \cellcolor{Green!0.0}57.34 & \cellcolor{Green!0.0}17.95 & \cellcolor{Green!0.0}26.86 & \cellcolor{Green!0.0}44.07 & \cellcolor{Green!0.0}42.28 & \cellcolor{Green!0.0}54.89 & \cellcolor{Green!0.0}60.68 \\
2-7 & \cellcolor{Green!0.0}37.92 & \cellcolor{Green!0.0}40.52 & \cellcolor{Green!0.0}33.49 & \cellcolor{Green!0.0}40.68 & \cellcolor{Green!0.0}36.04 & \cellcolor{Green!0.0}42.58 & \cellcolor{Green!0.0}27.11 & \cellcolor{Green!0.0}30.56 & \cellcolor{Green!0.0}54.32 & \cellcolor{Green!0.0}31.43 & \cellcolor{Green!0.0}38.45 & \cellcolor{Green!0.0}42.71 \\
2-8 & \cellcolor{Green!0.0}61.2 & \cellcolor{Green!0.0}59.0 & \cellcolor{Green!0.0}36.4 & \cellcolor{Green!0.0}37.4 & \cellcolor{Green!0.0}24.0 & \cellcolor{Green!0.0}26.8 & \cellcolor{Green!0.0}36.2 & \cellcolor{Green!0.0}30.6 & \cellcolor{Green!20.6}8.8 & \cellcolor{Green!1.4}11.4 & \cellcolor{Green!1.0}18.6 & \cellcolor{Green!1.8}20.2 \\
2-9 & \cellcolor{Green!2.2}78.82 & \cellcolor{Green!2.6}76.55 & \cellcolor{Green!4.8}66.48 & \cellcolor{Green!5.4}67.59 & \cellcolor{Green!20.2}44.88 & \cellcolor{Green!6.4}50.63 & \cellcolor{Green!10.6}62.93 & \cellcolor{Green!4.4}63.42 & \cellcolor{Green!2.4}16.9 & \cellcolor{Green!23.2}21.26 & \cellcolor{Green!1.0}31.74 & \cellcolor{Green!3.2}40.27 \\
2-10 & \cellcolor{Green!0.0}65.09 & \cellcolor{Green!0.0}65.26 & \cellcolor{Green!0.0}39.05 & \cellcolor{Green!0.0}40.04 & \cellcolor{Green!0.0}18.9 & \cellcolor{Green!0.0}21.27 & \cellcolor{Green!0.0}20.94 & \cellcolor{Green!0.0}20.69 & \cellcolor{Green!0.0}7.78 & \cellcolor{Green!0.0}7.04 & \cellcolor{Green!0.0}14.56 & \cellcolor{Green!0.0}17.37 \\
\hline
3-1 & \cellcolor{Green!0.4}52.47 & \cellcolor{Green!0.4}53.2 & \cellcolor{Green!1.2}29.5 & \cellcolor{Green!1.2}30.81 & \cellcolor{Green!8.4}44.62 & \cellcolor{Green!3.4}52.86 & \cellcolor{Green!7.0}34.86 & \cellcolor{Green!7.4}38.88 & \cellcolor{Green!0.0}25.19 & \cellcolor{Green!0.2}3.86 & \cellcolor{Green!5.8}33.28 & \cellcolor{Green!0.4}25.99 \\
3-2 & \cellcolor{Green!0.0}27.54 & \cellcolor{Green!0.0}33.15 & \cellcolor{Green!0.0}31.3 & \cellcolor{Green!0.0}34.49 & \cellcolor{Green!0.0}33.5 & \cellcolor{Green!0.0}34.49 & \cellcolor{Green!0.0}19.11 & \cellcolor{Green!0.0}28.7 & \cellcolor{Green!0.0}22.18 & \cellcolor{Green!0.0}32.96 & \cellcolor{Green!0.0}31.55 & \cellcolor{Green!0.0}33.96 \\
3-3 & \cellcolor{Green!25.0}41.99 & \cellcolor{Green!23.8}41.3 & \cellcolor{Green!29.0}35.52 & \cellcolor{Green!32.4}34.55 & \cellcolor{Green!24.6}40.67 & \cellcolor{Green!26.2}39.91 & \cellcolor{Green!23.0}41.05 & \cellcolor{Green!22.6}42.25 & \cellcolor{Green!5.4}55.93 & \cellcolor{Green!11.6}43.6 & \cellcolor{Green!35.4}27.9 & \cellcolor{Green!26.0}12.24 \\
3-4 & \cellcolor{Green!0.4}82.62 & \cellcolor{Green!0.4}83.21 & \cellcolor{Green!3.8}78.75 & \cellcolor{Green!5.6}77.12 & \cellcolor{Green!7.6}76.74 & \cellcolor{Green!6.0}78.47 & \cellcolor{Green!21.8}63.21 & \cellcolor{Green!18.8}67.74 & \cellcolor{Green!2.4}77.23 & \cellcolor{Green!1.4}78.95 & \cellcolor{Green!3.8}76.18 & \cellcolor{Green!5.6}74.31 \\
3-5 & \cellcolor{Green!0.4}81.91 & \cellcolor{Green!0.4}82.74 & \cellcolor{Green!4.0}76.84 & \cellcolor{Green!5.4}73.72 & \cellcolor{Green!7.4}77.19 & \cellcolor{Green!11.6}73.92 & \cellcolor{Green!16.6}67.2 & \cellcolor{Green!13.0}71.1 & \cellcolor{Green!2.0}75.52 & \cellcolor{Green!1.2}79.0 & \cellcolor{Green!5.0}73.57 & \cellcolor{Green!5.4}73.01 \\
3-6 & \cellcolor{Green!0.0}48.6 & \cellcolor{Green!0.0}49.6 & \cellcolor{Green!0.6}27.4 & \cellcolor{Green!0.0}31.6 & \cellcolor{Green!0.4}26.8 & \cellcolor{Green!0.0}26.8 & \cellcolor{Green!0.0}34.2 & \cellcolor{Green!0.0}36.2 & \cellcolor{Green!29.2}7.0 & \cellcolor{Green!20.0}13.8 & \cellcolor{Green!0.6}28.8 & \cellcolor{Green!0.0}26.8 \\
3-7 & \cellcolor{Green!0.4}77.6 & \cellcolor{Green!0.2}77.0 & \cellcolor{Green!0.4}61.2 & \cellcolor{Green!0.0}66.4 & \cellcolor{Green!0.4}42.0 & \cellcolor{Green!0.0}44.6 & \cellcolor{Green!0.6}43.8 & \cellcolor{Green!0.8}44.0 & \cellcolor{Green!0.8}47.2 & \cellcolor{Green!0.0}38.2 & \cellcolor{Green!6.0}41.4 & \cellcolor{Green!3.6}42.0 \\
3-8 & \cellcolor{Green!0.0}19.65 & \cellcolor{Green!0.0}19.9 & \cellcolor{Green!0.0}17.45 & \cellcolor{Green!0.0}17.17 & \cellcolor{Green!0.0}19.32 & \cellcolor{Green!0.0}20.39 & \cellcolor{Green!0.0}13.37 & \cellcolor{Green!0.0}12.11 & \cellcolor{Green!0.0}16.64 & \cellcolor{Green!0.0}13.95 & \cellcolor{Green!0.0}17.17 & \cellcolor{Green!0.0}16.72 \\
\hline
AVG &  \cellcolor{Green!1.5}52.35 & \cellcolor{Green!1.6}53.85 & \cellcolor{Green!2.6}42.15 & \cellcolor{Green!2.8}44.52 & \cellcolor{Green!6.5}37.00 & \cellcolor{Green!5.9}38.99 & \cellcolor{Green!5.4}35.73 & \cellcolor{Green!5.0}37.28 & \cellcolor{Green!7.7}33.05 & \cellcolor{Green!8.4}28.78 & \cellcolor{Green!5.7}32.76 & \cellcolor{Green!6.2}32.63 \\
\bottomrule
\end{tabular}
}
\begin{tabular}{ccccccc}
	\multicolumn{2}{c}{\%abstention\qquad} & \cellcolor{Green!0.0}\quad0\% \quad & \cellcolor{Green!25}\quad25\% \quad & \cellcolor{Green!50}\quad50\% \quad &
	\cellcolor{Green!75}\quad75\% \quad & \cellcolor{Green!100}\quad100\% \quad \\
\end{tabular}
\caption{\small Model performance of top two performing systems from each category. qwen-chat and InternLM-chat both have 7B parameters. Cells are colored according to model abstention rate. Further results are in Appendix \ref{appendix:details_of_results}. It can be observed that the performance of one-shot surpasses that of zero-shot.}
\label{tab:top_2_zero_shot_performance}
\end{table*}

Figure \ref{fig:overall_performance_one_shot_main_content} shows the overall zero-shot performance of each model. As can be seen, GPT-4 and ChatGPT clearly lead the benchmark, substantially outperform all other models. Under the same model size (7B-13B), Chinese oriented LLMs outperform multilingual models such as MPT and Llama by a significant margin, confirming the effectiveness of pre-training and fine-tuning on Chinese data. Interestingly, legal specific LLMs do not necessarily outperform general-purpose Chinese oriented LLMs. Close inspection reveals that existing legal specific LLMs are based on rather weak foundation models, implying that improved models may be obtained by fine-tuning a stronger foundation model (see Section \ref{sec:analysis} for more detail).

In Table \ref{tab:top_2_zero_shot_performance}, we demonstrate the top 2 performing LLMs from each category on all tasks, together with the model abstention rate.\footnote{We present the complete results with the performance achieved by every model on all tasks in Table \ref{tab:Zero-shot Results on Legal Knowledge Memorization Tasks}-\ref{tab:zero-shotResults on Overall} (complete zero-shot results), Table \ref{tab:one-shotResults on legal knowledge memorisation tasks}-\ref{tab:overall_results} (complete one-shot results), in Appendix \ref{appendix:details_of_results}.} We made the following observations. First, there is substantial variation in the distribution of scores across tasks. The best-performing models, for example, can achieve a score of more than 60 on tasks 3-4 and 3-5, yet no models manage to exceed 30 on tasks 1-1 and 2-1. This shows that our benchmark adequately assess model capabilities in various aspects. Second, we notice that GPT-4/ChaGPT not only performs well on the majority of tasks, but also has a low abstention rate, suggesting that they are excellent at following instructions and providing responses that are more relevant to the query. Third, it can be observed that GPT models and the Chinese oriented LLMs can successfully leverage the one-shot example and make more accurate predictions compared to zero-shot cases. The top-performing legal specific LLMs, however, suffers a drop in performance after seeing the one-shot example. We hypothesize that due to the fact that they are primarily trained on legal specific instruction data, their instruction-following skills are negatively impacted. Fourth, Fuzi-Mingcha scores 97.59 on task 2-4, whereas all other models score less than 65. Considering its performance on other legal tasks, we suspect that there is a data contamination. This also highlights potential caveats when evaluating LLMs~\cite{schaeffer2023pretraining}. Overall, it is promising that most LLMs show some capability in handling legal tasks, but there's still a substantial room for improvement. Even the top-performing model, GPT-4, achieves only an average score of 52.35 (zero-shot) / 53.85 (one-shot), highlighting the need for additional efforts in the future.

\begin{figure}[t]
    \centering
    \includegraphics[width=\columnwidth]{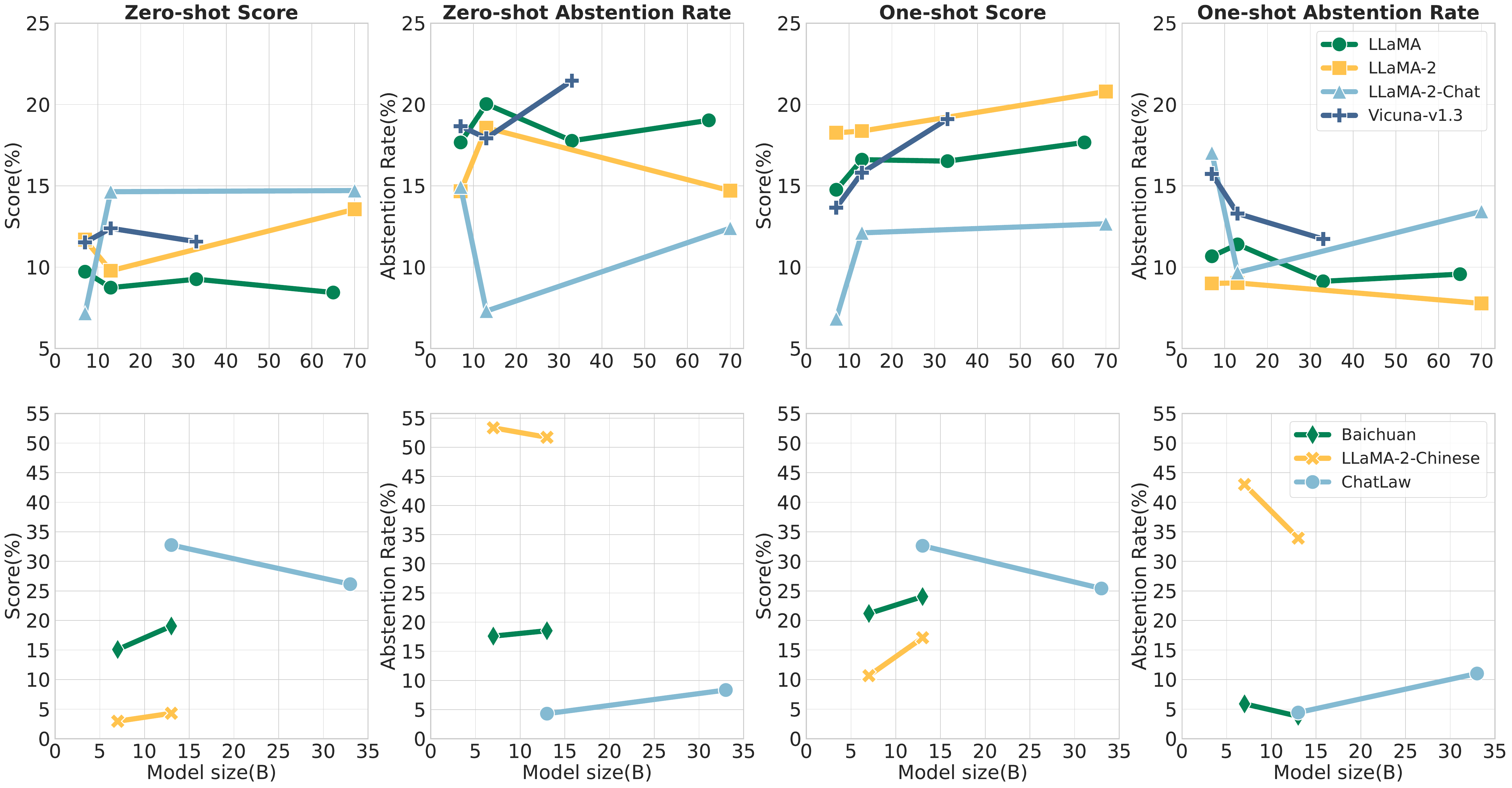}
    \caption{\small Performances of models with various sizes. It can be observed that scaling up the model size usually improves the performance, but the improvement is more consistent in the one-shot than in the zero-shot scenario. }
    \label{fig:effects_scale}
\end{figure}

\begin{figure}[t]
    \centering
    \includegraphics[width=\columnwidth]{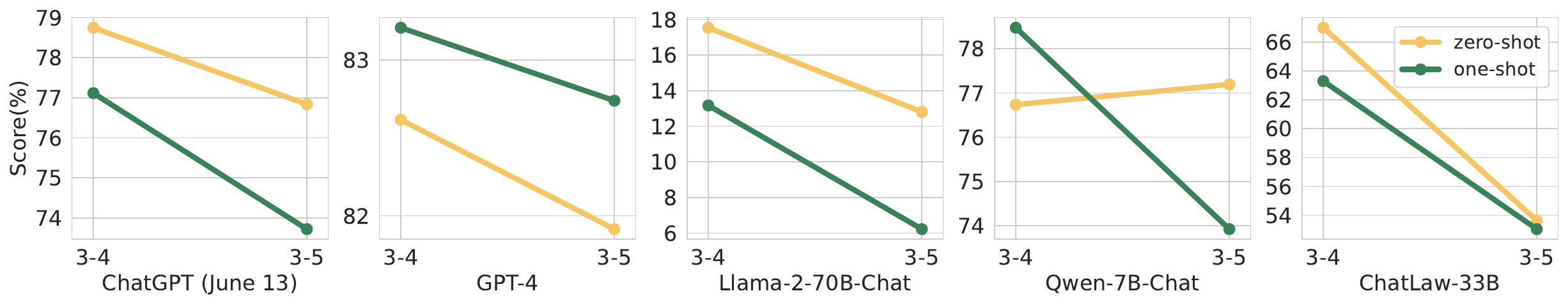}
    \caption{\small Comparison between different models for 3-4 and 3-5 tasks. Most LLMs are \emph{unable} of properly utilizing legal article content information to aid in judgment prediction. Including the article content often degrades the performance.}
    \label{fig:effects_article}
\end{figure}

\subsection{Analysis}
\label{sec:analysis}

We find that the current legal specific LLMs do not necessarily outperform general large language models. We analyze the effect of model size and training approach on large language models to better understand which aspects most influence model performance.

\paragraph{Scaling up the model size results in better performance in one-shot case.}

\begin{figure}[t]
    \centering
    \includegraphics[width=\columnwidth]{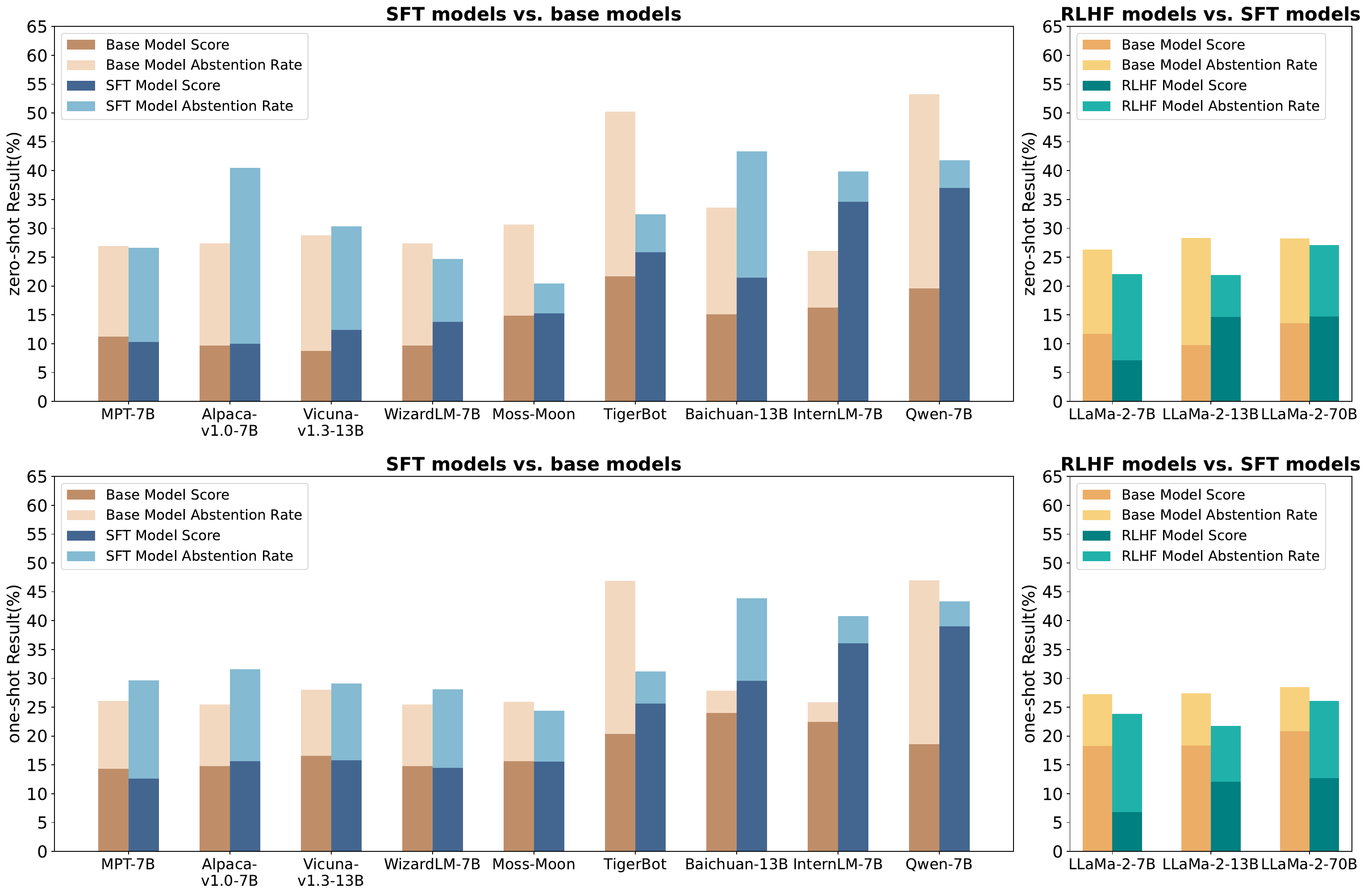}
    \caption{\small Comparison of base LLMs and their SFT and RLHF variants. We put the abstention rate bar on top of the model score bar to visualize the improvement space for (1) correctly following instructions and (2) task-specific precision. The abstention rate score is averaged only on tasks requiring answer extraction.}
    \label{fig:effects_sft}
\end{figure}

Prior work shows that larger models perform better in general NLP tasks~\cite{kaplan2020scaling}. We analyze whether this finding still holds in legal domain. Specifically, we select representative models from different categories with varying model sizes and calculate the overall performance and abstention rate of different tasks. The results are shown in Figure \ref{fig:effects_scale}. We observe that increasing the model size typically helps improve model performance in one-shot settings. Also, the abstention rate becomes lower, indicating that larger models are better at following instructions. Nevertheless, we find that ChatLaw is an outlier, a larger size results in lower performance. In zero-shot scenarios, mixed results are observed, simply increasing the model size may not automatically lead to better performance.

\paragraph{Most LLMs can not efficiently leverage article content.} 
Retrieval augmentation is a common way to improve the accuracy of generative models~\cite{lewis2020retrieval,jiang2023active,shen2023neural}. By including the content of the related legal articles in task 3-4 and feeding them into the LLM input, we replicate the retrieval augmentation scenario to form task 3-5. The goal is to see if the model can successfully use this additional knowledge to predict the proper jail sentence when supplied with the relevant article reference that articulates the range of prison terms. We compare 5 models of various types and visualize the comparison between task 3-4 and 3-5 in Figure \ref{fig:effects_article}. The results show that the vast majority of models \emph{fail} to make any progress on the jail sentence prediction challenge by using the provided article content. There is a decrease in performance across the board for most models, including GPT-4. This suggests that simple retrieval enhancement methods may not bring further improvement, and how to obtain LLMs that are able to effectively utilize the retrieval information is still an open problem. 

\begin{figure}[t]
    \centering
    \includegraphics[width=\columnwidth]{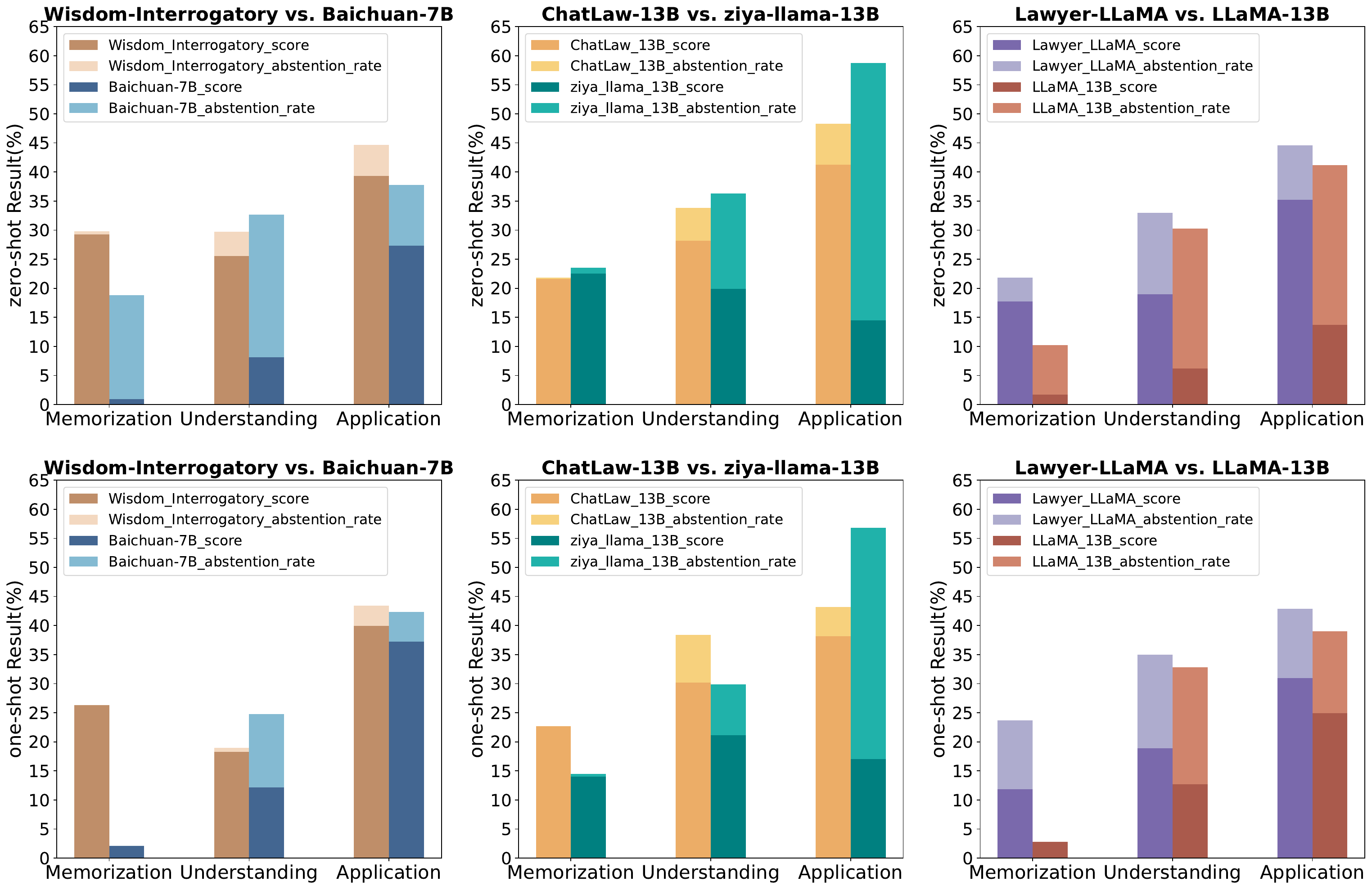}
    \caption{\small Comparison between different legal specific LLMs and their base models. Legal specific fine-tuning significantly improves the performance and reduces the abstention rate.}
    \label{fig:effects_law_adapt}
\end{figure}

\paragraph{SFT may improve the performance but RLHF may not.}
We illustrate the performance difference of 9 models before and after SFT in Figure \ref{fig:effects_sft}. In most cases, models perform better through SFT. Notably, the SFT data are collected from the general domain, yet they still substantially improve model performance on legal tasks. The LLaMA-2 series further applies RLHF on top of SFT. However, we do find that RLHF-trained models can refuse to answer some questions~\cite{ouyang2022training} (resulting in a higher abstention rate and lower scores), which can lead to a drop in performance on legal tasks. This suggests that when creating models for legal tasks, stacking RLHF on top of SFT can be counterproductive.

\paragraph{Legal specific fine-tuning is helpful.} 
To assess the impact of legal specific fine-tuning, we contrast three LLMs fine-tuned for legal tasks with their corresponding base models, as depicted in Figure~\ref{fig:effects_law_adapt}. It is evident that after legal specific fine-tuning, there is a consistent enhancement of model scores and reduction of abstention rates. This underscores the efficacy of the training strategies employed. Closer inspection of the three cognitive levels reveals that Baichuan-7B and LLaMA-13B perform very poorly on memorizing tasks, which suggests they have not been pre-trained on large, high-quality legal corpora. Nonetheless, fine-tuning them on legal corpora both lead to significant improvement. Even for LLMs like Ziya-LLama-13b who have excellent memorization of legal principles, fine-tuning that is unique to the field of law can significantly boost performance on tests requiring comprehension and applying skills. Comparing to the LawGPT series, we found that version 1.1, which is fine-tuned only on 350k instruction data, outperformed version 1.0 which is fine-tuned on 500k judgement documents followed by 300k instructions. This suggests that beginning with a lower-performing model and undertaking continuous pre-training, which is both time-consuming and resource-intensive, is inferior to starting with a superior foundation model and fine-tuning on high-quality instruction data. Future research may want to try fine-tuning a stronger base model (like qwen-chat or InternLM-chat) for better performance on legal tasks.

\section{Conclusion}
Using LLMs to benefit the legal domain is a promising topic. However, existing benchmarks to measure the legal knowledge of LLMs either focus on a limited subset of tasks, or are based on American laws in English language. This paper presents \emph{LawBench}: a meticulously crafted, comprehensive evaluation benchmark to assess LLMs in performing legal-related tasks under the Chinese civil law system. We provide  structured taxonomy of the skill set required for legal-related tasks, including 20 diverse tasks corresponding to 3 cognitive dimensions: legal knowledge memorization, understanding and applying. We undertake a thorough examination of 51 LLMs and assess their performance. The results demonstrate that current LLMs are still unable to give meaningful judicial aid, and their scores on most tasks are often poor. While fine-tuning open-source LLMs on legal specific language results in some advances, they still lag far below GPT-4. As the legal field is highly professional, much of the data used in practical applications is confidential. Developing a high-quality large language model for legal tasks necessitates collaboration among multiple institutions. We hope the release of LawBench can serve as a foundation for future research and we seek to encourage cooperation in order to further this effort.

\section*{Limitations}
The majority of our datasets are acquired through sampling publicly available data on the internet. Even though we have made efforts to select newest versions of datasets, there can still be risks of test data leakage given that existing LLMs have been exhaustively trained on massive amount of Internet data. It is possible that LLMs explicitly trained on these task formats, or even the exact test data, can exhibit exceptionally high scores~\cite{schaeffer2023pretraining}. We will seek more principled ways to prevent data contamination in the future.

Another notable limitation is the answer extraction methods and evaluation metrics for generative tasks. Even though we have hand-engineered task-specific rules to extract the answer, there still can be cases that the rule fails to match. For generative tasks, we only use Rouge-L to evaluate the model predictions for convenience, which cannot fully reflect the human judgement about the answer quality. Currently, there is a lack of automated methods to effectively evaluate model predictions from legal aspects. We plan to consider training an evaluation model tailored for legal tasks in the future, or experiment with LLM-based evaluations~\cite{liu2023mmbench,yu2023kola}.

\section*{Acknowledgments}
This work is an extension which was supported by National Key R\&D Program of China (2016YFC0800803).  This work is also partly
supported by the National Key R\&D Program of China
No.2022ZD016100 and Shanghai Postdoctoral Excellence
Program (No.2022235). We appreciate Zhixin Yin for helping arrange tables and alter chart layouts. We also thank Qi Li for collecting and organizing some public data.

\bibliography{neurips_2023}
\bibliographystyle{plain}
\clearpage
\appendix
\section{Details of Task Instruction}
\label{appendix:details_of_task_intruction}

\subsection{Legal Knowledge Memorization Tasks}
\label{app:instruction_memorization}

\definecolor{'shallow1'}{HTML}{E4F3FC} 
\begingroup
\begin{table*}[ht]

    \centering
    \small
    \begin{tabular}{p{\linewidth}}
        \toprule

            \vspace{-2mm}
         \cellcolor{'shallow1'} \textbf{\textsc{Instruction:}} Answer the following questions, just give the content of the law directly: \\
        \midrule
        \vspace{-1mm}
        \cellcolor{'shallow1'} \textbf{\textsc{Question:}} What is the content of Article 257 of the Criminal Law? \\
        \midrule
        \vspace{-1mm}
        \cellcolor{'shallow1'} \textbf{\textsc{Answer:}} Anyone who uses violence to interfere with the marriage freedom of others shall be sentenced to fixed-term imprisonment of no more than two years or detention. Anyone who commits the crime mentioned in the preceding paragraph and causes the death of the victim shall be sentenced to fixed-term imprisonment of no less than two years and no more than seven years.\\
        \bottomrule
    \end{tabular}
        \caption{
  The instruction and an example of Task 1-1 Article Recitation.
    }
    \label{tab:instruction_1_1}
\end{table*}
\endgroup
\definecolor{'shallow1'}{HTML}{E4F3FC} 
\begingroup
\begin{table*}[ht]

    \centering
    \small
    \begin{tabular}{p{\linewidth}}
        \toprule

            \vspace{-2mm}
        \cellcolor{'shallow1'} \textbf{\textsc{Instruction:}} Please apply your legal knowledge to select the correct answer from A, B, C, or D and write it between [Correct Answer] and <eoa>. For example, [Correct Answer] A <eoa>. Please strictly follow this format when answering. \\
        \midrule
        \vspace{-1mm}
        \cellcolor{'shallow1'} \textbf{\textsc{Question:}} According to the laws of our country, in the process of trial of foreign-related divorce cases accepted by our courts, which of the following should be used as the basis for determining the validity of the marriage? A: The law of the place where the marriage was concluded; B: The law of the parties' home country; C: The law of the parties' place of residence; D: The law of the court.
         \\
        \midrule
        \vspace{-1mm}
        \cellcolor{'shallow1'} \textbf{\textsc{Answer:}} [Correct Answer] A<eoa>\\
        \bottomrule
    \end{tabular}
        \caption{
  The instruction and an example of Task 1-2 Knowledge Question Answering.
    }
    \label{tab:instruction_1_2}
\end{table*}
\endgroup

\clearpage
\subsection{Legal Knowledge Understanding Tasks}
\label{app:instruction_understanding}

\definecolor{'shallow2'}{HTML}{F3FADF} 
\begingroup
\begin{table*}[ht]

    \centering
    \small
    \begin{tabular}{p{\linewidth}}
        \toprule

            \vspace{-2mm}
        \cellcolor{'shallow2'} \textbf{\textsc{Instruction:}} Correct the spelling, redundancy, omission, and disorder errors of the words in the following legal document sentences, and minimize modifications to the original sentences while preserving their semantics. Only the modified sentences need to be provided in the answer, strictly following this format. Sentence: \\
        \midrule
        \vspace{-1mm}
        \cellcolor{'shallow2'} \textbf{\textsc{Query:}} The above-mentioned procedure for collecting the evidence for the appeal is legal, and the content is objective and true, which is sufficient to establish the accusation.\\
        \midrule
        \vspace{-1mm}
        \cellcolor{'shallow2'} \textbf{\textsc{Answer:}} The above-mentioned procedure for collecting evidence is legal, and the content is objective and true, which is sufficient to establish the facts of the accusation.\\
        \bottomrule
    \end{tabular}
        \caption{
  The instruction and an example of Task 2-1 Document Proofread.
    }
    \label{tab:instruction_2_1}
\end{table*}
\endgroup
\begingroup
\begin{table*}[ht]

    \centering
    \small
    \begin{tabular}{p{\linewidth}}
        \toprule

            \vspace{-2mm}
        \cellcolor{'shallow2'} \textbf{\textsc{Instruction:}} Determine the category of the dispute focus contained in the sentence. Each sentence contains only one dispute focus category. Categories include: litigation parties, rent situation, interest, principal dispute, liability determination, liability division, loss determination and handling, whether the original judgment is appropriate, contract effectiveness, property division, liability assumption, admissibility of appraisal conclusions, statute of limitations, breach of contract, contract termination, hit-and-run. Write the answer between [Dispute Focus] and <eoa>, such as [Dispute Focus] Principal Dispute <eoa>. \\
        \midrule
        \vspace{-1mm}
        \cellcolor{'shallow2'} \textbf{\textsc{Query:}} Sentence: The plaintiff alleges: I have a father-son relationship with the defendant. The house located at Unit 321, Building 7, 3 North Xiaojie, Sanlitun North, Chaoyang District, Beijing (hereinafter referred to as the 'involved property') was allocated to me by the Ministry of Foreign Affairs Diplomatic Service Bureau during my time at TIME. My unit underwent housing reform, and I purchased the property at TIME. I have been living in the involved property from TIME to the present. In TIME, the defendant started giving various reasons for not allowing me to reside in the involved property, which raised my suspicions. In TIME, I went to ORG to inquire about the status of the involved property and discovered that in TIME, the defendant deceived me into signing an 'Existing Property Sale Contract' and processed the property transfer procedures at the Housing Management Bureau, transferring the involved property into the defendant's name. I believe I have never sold the involved property, and the defendant deceived me into signing the 'Existing Property Sale Contract.' The defendant also did not actually pay the purchase price specified in the property sale contract. Therefore, I have filed a lawsuit to the court requesting the cancellation of the 'Property Sale Contract' signed between me and the defendant and the transfer of the property to my name. The defendant argues: The plaintiff and I signed the contract voluntarily. The involved property is state-owned property, and the sale of state-owned property requires strict approval procedures. All the documents were signed by the plaintiff himself, and I also fully paid the purchase price. Now the plaintiff is backtracking because several other children are manipulating the elderly, and the elderly are confused. I do not agree with the plaintiff's lawsuit request.\\
        \midrule
        \vspace{-1mm}
        \cellcolor{'shallow2'} \textbf{\textsc{Answer:}} Dispute focus category: contract effectiveness <eoa>\\
        \bottomrule
    \end{tabular}
        \caption{
  The instruction and an example of Task 2-2 Dispute Focus Identification.
    }
    \label{tab:instruction_2_2}
\end{table*}
\endgroup
\begingroup
\begin{table*}[ht]

    \centering
    \small
    \begin{tabular}{p{\linewidth}}
        \toprule

            \vspace{-2mm}
        \cellcolor{'shallow2'} \textbf{\textsc{Instruction:}} Please categorize these sentences based on tags, which include: having children after marriage, custody of children with limited capacity, having joint property, paying child support, dividing real estate, post-marital separation, filing for divorce for the second time, monthly payment of child support, granting divorce, having joint debt, personal property before marriage, legal divorce, failure to fulfill family obligations, existence of non-marital children, appropriate assistance, failure to comply with divorce agreements, compensation for damages, separation due to emotional discord for over two years, children living with non-custodial parent, personal property after marriage. Please write the answer between [category] and <eoa>, for example, [category] division of real estate, joint property <eoa>. Please strictly follow this format. \\
        \midrule
        \vspace{-1mm}
        \cellcolor{'shallow2'} \textbf{\textsc{Query:}} In 2014, Wang X sued and demanded a share of the down payment of 150,000 yuan on the grounds that he and Xiao X had not divided the down payment of the building when they agreed to divorce.\\
        \midrule
        \vspace{-1mm}
        \cellcolor{'shallow2'} \textbf{\textsc{Answer:}} [category] dividing real estate, having joint property <eoa>\\
        \bottomrule
    \end{tabular}
        \caption{
  The instruction and an example of Task 2-3 Marital Disputes Identification.
    }
    \label{tab:instruction_2_3}
\end{table*}
\endgroup
\begingroup
\begin{table*}[ht]

    \centering
    \small
    \begin{tabular}{p{\linewidth}}
        \toprule

            \vspace{-2mm}
        \cellcolor{'shallow2'} \textbf{\textsc{Instruction:}} Please determine the category for the following consultation. Each consultation belongs to only one category, which includes: Marriage and Family, Labor Disputes, Traffic Accidents, Debt Collection and Debt Disputes, Criminal Defense, Contract Disputes, Real Estate Disputes, Infringement, Corporate Law, Medical Disputes, Demolition and Resettlement, Administrative Litigation, Construction Engineering, Intellectual Property Rights, Comprehensive Consultation, Personal Injury, Foreign-related Law, Maritime Law, Consumer Rights, Mortgage and Guarantee. Please write the answer between [Category] and <eoa>, for example [Category] Marriage and Family <eoa>. Please strictly follow this format when answering.

        Consultation: \\
        \midrule
        \vspace{-1mm}
        \cellcolor{'shallow2'} \textbf{\textsc{Query:}}It has been two years since we separated and have not been together. Will the court automatically grant a divorce? The other party has sued for the dowry money, but the case has not been settled. Will this affect the other party's ability to remarry and obtain a marriage certificate? \\
        \midrule
        \vspace{-1mm}
        \cellcolor{'shallow2'} \textbf{\textsc{Answer:}} Marriage and Family \\
        \bottomrule
    \end{tabular}
        \caption{
  The instruction and an example of Task 2-4 Issue Topic Identification.
    }
    \label{tab:instruction_2_4}
\end{table*}
\endgroup
\begingroup
\begin{table*}[ht]

    \centering
    \small
    \begin{tabular}{p{\linewidth}}
        \toprule

            \vspace{-2mm}
        \cellcolor{'shallow2'} \textbf{\textsc{Instruction:}} Please answer the questions based on the provided paragraph as concisely as possible. \\
        \midrule
        \vspace{-1mm}
        \cellcolor{'shallow2'} \textbf{\textsc{Query:}} Paragraph: Based on the valid evidence submitted by the plaintiff to the court and the plaintiff's oral statements in court, this court confirms the following facts in the case: The defendant Li2 and the defendant Li x3 are mother and daughter. On September 26, 2014, the defendant Li2 borrowed 50,000 yuan from Sang x0, and issued a promissory note stating: 'I have borrowed 50,000 yuan in cash from Sang x0 today. The loan term is 3 months, and I will repay it on time every month with an interest of 2,500 yuan per month. If I cannot repay it at the due date, I am willing to use all my assets to settle the debt and take full responsibility. Borrower: Si x1. Spouse: Li2. Willing to take full responsibility.' On November 13, 2014, with the guarantee of the defendant Li x3, the defendant Li2 borrowed 150,000 yuan from Sang x0 and issued a promissory note stating: 'I have borrowed 150,000 yuan in cash from Sang x0 today. The loan term is three months, and I promise to pay the principal and interest monthly and return it on time. I hereby promise. Borrower: Li2, Guarantor: Li x3, Spouse: Si x1.' On the same day, Li2 issued a repayment guarantee commitment, and Si x1 signed it as a spouse. At the same time, Li2 and Si x1 issued a mortgage commitment, promising to use the existing house, car, and any other property in Zone 13 of Dongjiang New Village, Dongjiang Town, as collateral for this loan. However, both parties did not complete the mortgage registration procedure. After the loan maturity date, the defendant Si x1, besides paying 22,000 yuan in interest to the plaintiff, did not repay the remaining principal and interest. The plaintiff then sued in this court, requesting: 1. The defendant Li2 and Si x1 to repay the principal of the loan, which is 200,000 yuan; 2. The defendant Li2 and Si x1 to repay the loan interest at a rate of 2\% per month; 3. The defendant Li x3 to assume the guarantee liability within the range of 150,000 yuan; 4. The defendant is to bear the litigation costs. Furthermore, it was found that the defendant Li2 and the defendant Si x1 did not complete the marriage registration procedure. Question: Who borrowed money from the lender? \\
        \midrule
        \vspace{-1mm}
        \cellcolor{'shallow2'} \textbf{\textsc{Answer:}} Answer: Si x1, Li2 \\
        \bottomrule
    \end{tabular}
        \caption{
  The instruction and an example of Task 2-5 Reading Comprehension.
}
    \label{tab:instruction_2_5}
\end{table*}
\endgroup
\begingroup
\begin{table*}[ht]

    \centering
    \small
    \begin{tabular}{p{\linewidth}}
        \toprule

            \vspace{-2mm}
        \cellcolor{'shallow2'} \textbf{\textsc{Instruction:}} Based on the provided entity types, extract entity information from the sentence. The entity types include: suspected criminals, victims, stolen currency, item value, theft profit, stolen items, crime tools, time, location, and organizations. List the entity information one by one. \\
        \midrule
        \vspace{-1mm}
        \cellcolor{'shallow2'} \textbf{\textsc{Query:}} Sentence: After the case was solved, the public security organ lawfully returned the seized mobile phones to the victims, Yan and Xiao.\\
        \midrule
        \vspace{-1mm}
        \cellcolor{'shallow2'} \textbf{\textsc{Answer:}} Victims: Yan, Xiao; stolen items: the seized mobile phones; organizations: the public security organ \\
        \bottomrule
    \end{tabular}
        \caption{
  The instruction and an example of Task 2-6 Name Entity Recognition.
    }
    \label{tab:instruction_2_6}
\end{table*}
\endgroup
\begingroup
\begin{table*}[ht]

    \centering
    \small
    \begin{tabular}{p{\linewidth}}
        \toprule

            \vspace{-2mm}
        \cellcolor{'shallow2'} \textbf{\textsc{Instruction:}} Here is a news report, please provide a summary of this report in one sentence. \\ 
        \midrule
        \vspace{-1mm}
        \cellcolor{'shallow2'} \textbf{\textsc{Query:}} According to a report by Xinmin Evening News (Reporter Xia Yun), at about 9 am today, some citizens reported to the journalist that a unit on Cao'an Road in Jiading District, Shanghai had a sudden fire, with thick smoke at the scene; the fire department arrived promptly and the fire was still being put out by the time of the report, with no casualties reported; the source is the official Weibo account of Shanghai Fire Rescue Bureau.According to witnesses, when the fire broke out, there was a large amount of smoke above the unit, and some drivers on the Jiamin elevated road also saw the smoke; after receiving the alarm, the police, fire and other departments rushed to the scene for disposal; as of 9:30 am, the disposal work was still in progress; according to the official Weibo of Shanghai Fire Rescue Bureau, at about 8:33 am on March 24, a company in Cao'an Road, Jiading District had a fire, and the fire department arrived at the scene for disposal, with the fire now under control and no casualties reported.\\
        \midrule
        \vspace{-1mm}
        \cellcolor{'shallow2'} \textbf{\textsc{Answer:}} At around 9 a.m. this morning, a unit on Cao'an Road in Jiading District, Shanghai had a sudden fire, with thick smoke at the scene.\\
        \bottomrule
    \end{tabular}
        \caption{
  The instruction and an example of Task 2-7 Opinion Summarization.
    }
    \label{tab:instruction_2_7}
\end{table*}
\endgroup
\begingroup
\begin{table*}[ht]

    \centering
    \small
    \begin{tabular}{p{\linewidth}}
        \toprule

            \vspace{-2mm}
        \cellcolor{'shallow2'} \textbf{\textsc{Instruction:}} Based on a given prosecution point of view, please select a point of view that can form a controversial point of view from the five defense candidate points of view A, B, C, D, and E, and write the answer in [Correct Answer] and <eoa>. For example [correct answer]A<eoa>. Please answer strictly according to this format. \\
        \midrule
        \vspace{-1mm}
        \cellcolor{'shallow2'} \textbf{\textsc{Query:}} Sentence: The private prosecutor Yu XX alleged that at around 17:00, the defendants Bao XX, Han XX, and Han XX came to my house and beat me on the head with a wooden stick on the grounds that I had detained their cattle. After I was injured, I was hospitalized at the Xing'an League Mongolian Medical Hospital for 73 days. A: The defender provided the court with a copy of Lv’s testimony, requesting the court to declare the defendant Lv innocent and reject the victim’s incidental civil claim. B: I think I am not guilty. C: The defendant Bao Moumou claimed that I injured the private prosecutor with a whip and agreed to compensate for the medical expenses. D: 2. The victim’s incidental civil lawsuit should be dismissed. E: In addition, disability compensation should not be included in the scope of judgment compensation. \\
        \midrule
        \vspace{-1mm}
        \cellcolor{'shallow2'} \textbf{\textsc{Answer:}} [Correct answer]C<eoa> \\
        \bottomrule
    \end{tabular}
        \caption{
  The instruction and an example of Task 2-8 Argument Mining.
}
    \label{tab:instruction_2_8}
\end{table*}
\endgroup

\clearpage
\subsection{Legal Knowledge Applying Tasks}
\label{app:instruction_application}
\begingroup
\begin{table*}[ht]

    \centering
    \small
    \begin{tabular}{p{\linewidth}}
        \toprule

            \vspace{-2mm}
        \cellcolor{'shallow3'} \textbf{\textsc{Instruction:}} Provide the relevant articles in the criminal code based on the following facts and charges. Please only include the criminal code article numbers, and place your answers between [Article] and <eoa>. For example, [Article] Article 128 of the Criminal Code, Article 341 of the Criminal Code <eoa>. \\
        \midrule
        \vspace{-1mm}
        \cellcolor{'shallow3'} \textbf{\textsc{Query:}}   Fact: The People's Procuratorate of Gongzhuling City, Jilin Province, alleges that on June 18, 2014, the defendant, Mr. Zhang, applied for a 'Da Jilin' credit card at the China Bank Gongzhuling City Branch. As of August 10, 2014, the outstanding balance reached 14,974.52 yuan. Despite two collection attempts by the issuing bank's staff, the defendant still did not repay the amount for over three months.

Charge: Fraud.\\
        \midrule
        \vspace{-1mm}
        \cellcolor{'shallow3'} \textbf{\textsc{Answer:}} Article: Article 196 of the Criminal Code\\
        \bottomrule
    \end{tabular}
        \caption{
  The instruction and an example of Task 3-1 Fact-based Article Prediction.
    }
    \label{tab:instruction_3_1}
\end{table*}
\endgroup
\begingroup
\begin{table*}[ht]

    \centering
    \small
    \begin{tabular}{p{\linewidth}}
        \toprule

            \vspace{-2mm}
        \cellcolor{'shallow3'} \textbf{\textsc{Instruction:}} Please provide legal justification based on specific scenarios and questions, presenting only the relevant legal articles. Each scenario should be associated with a single legal article.\\
        \midrule
        \vspace{-1mm}
        \cellcolor{'shallow3'} \textbf{\textsc{Query:}} Scenario: A company's board of directors has decided to appoint one of its directors as a concurrent manager to enhance the company's management. According to which legal provision, can the board of directors make the decision to have a board member serve concurrently as a manager? \\
        \midrule
        \vspace{-1mm}
        \cellcolor{'shallow3'} \textbf{\textsc{Answer:}} According to Article 114 of the Company Law, the board of directors of a company can decide to appoint one of its members as a concurrent manager.\\
        \bottomrule
    \end{tabular}
        \caption{
  The instruction and an example of Task 3-2 Scene-based Article Prediction.
    }
    \label{tab:instruction_3_2}
\end{table*}
\endgroup
\begingroup
\begin{table*}[ht]

    \centering
    \small
    \begin{tabular}{p{\linewidth}}
        \toprule

            \vspace{-2mm}
        \cellcolor{'shallow3'} \textbf{\textsc{Instruction:}} Please simulate a judge and provide the charge based on the following facts. Only the name of the charge is required. Place your answer between [Charge] and <eoa>. For example, [Charge] Theft; Fraud<eoa>. Please strictly adhere to this format in your response. \\
        \midrule
        \vspace{-1mm}
        \cellcolor{'shallow3'} \textbf{\textsc{Query:}} Facts: The People's Procuratorate of Haidian District, Beijing, in the indictment, alleges the following:

On June 27, 2013, the defendant, Ren (owner of an East Wind Nissan Yida sedan with license plate number Jing PNB057 and insured under this vehicle), in Haidian District of Beijing, fabricated a fictitious traffic accident involving an East Wind Nissan Yida sedan (license plate number Jing PNB057) and a BMW sedan (license plate number Jing QSU596) to deceive China People's Property Insurance Co., Ltd., Beijing Branch, into paying RMB 14,160.8.

On August 15, 2013, the defendant, Ren, in Haidian District of Beijing, fabricated a fictitious traffic accident involving a Jinbei sedan (license plate number Jing KP2320) and a BMW sedan (license plate number Jing QSU596) to deceive China People's Property Insurance Co., Ltd., Beijing Branch, into paying RMB 10,231.2.

On September 26, 2013, the defendant, Ren, was summoned by the public security authorities. All the funds involved in the case were refunded to the victim company after the incident. \\
        \midrule
        \vspace{-1mm}
        \cellcolor{'shallow3'} \textbf{\textsc{Answer:}} Charge: Insurance fraud\\
        \bottomrule
    \end{tabular}
        \caption{
  The instruction and an example of Task 3-3 Charge Prediction.
    }
    \label{tab:instruction_3_3}
\end{table*}
\endgroup
\begingroup
\begin{table*}[ht]

    \centering
    \small
    \begin{tabular}{p{\linewidth}}
        \toprule

            \vspace{-2mm}
        \cellcolor{'shallow3'} \textbf{\textsc{Instruction:}}  Based on the following facts, charges, and articles of the criminal code, predict the length of the sentence. Only provide the sentence length in months, and place your answer between [Sentence] and <eoa>. For example, [Sentence] 12 months <eoa>.\\
        \midrule
        \vspace{-1mm}
        \cellcolor{'shallow3'} \textbf{\textsc{Query:}} Facts: The prosecuting authority alleges that on December 23, 2015, the defendant, Yang, took advantage of his position as the former secretary of Qiansuo Street in Jiaojiang District, Taizhou City, and misappropriated RMB 140,000 of pre-collected construction fees from Shangxu Village residents. He used this money to cover expenses related to his eyeglass factory.

On March 20, 2016, the defendant, Yang, once again took advantage of his position and misappropriated RMB 51,765 of Shangxu Village's funds. He used this money to repay a bank loan for his eyeglass factory.

On March 31 of the same year, the defendant, Yang, returned all the misappropriated funds to the village collective account. On February 23, 2017, at approximately 10 o'clock, Yang was summoned to the case by the Economic Investigation Brigade of Jiaojiang Branch, Taizhou Public Security Bureau, at his residence at No. 2 Shangxu Village, Qiansuo Street, Jiaojiang District, Taizhou City.

Charge: Embezzlement \\
        \midrule
        \vspace{-1mm}
        \cellcolor{'shallow3'} \textbf{\textsc{Answer:}} Prison term: 10 months\\
        \bottomrule
    \end{tabular}
        \caption{
  The instruction and an example of Task 3-4 Prison Term Prediction w.o Article.
    }
    \label{tab:instruction_3_4}
\end{table*}
\endgroup
\begingroup
\begin{table*}[ht]

    \centering
    \small
    \begin{tabular}{p{\linewidth}}
        \toprule

            \vspace{-2mm}
        \cellcolor{'shallow3'} \textbf{\textsc{Instruction:}}  Based on the following facts, charges, and articles of the criminal code, predict the length of the sentence. Only provide the sentence length in months, and place your answer between [Sentence] and <eoa>. For example, [Sentence] 12 months <eoa>.\\
        \midrule
        \vspace{-1mm}
        \cellcolor{'shallow3'} \textbf{\textsc{Query:}} Facts: The public prosecution accuses that on July 3, 2014, at around 12 o'clock, the defendant, Sun, became very angry when he found that a fellow villager, Zhang, had not returned home after going out due to drinking. He then climbed over the wall into Zhang's courtyard. Inside the house, he smashed and tore apart Zhang's TV, stereo, clothes, and other belongings, burning some of the clothes. When Zhang returned home and tried to stop the defendant, Sun, he was punched in the face by the defendant. According to the assessment, the damaged property was worth 1,580 yuan. On October 10, 2014, at around 8 pm, the defendant, Sun, sent a message to Zhang, threatening to jump into his house. Zhang hastily called his brother-in-law, and upon learning of the situation, his brother-in-law, Sun, reported it to the police. After the incident, both parties reached a compensation agreement. The defendant, Sun, compensated the victim, Zhang, for his losses and obtained the victim's forgiveness.

Charge: Illegal intrusion into a residence.

Legal Article: Article 245 of the Criminal Law. 

Legal article content: Article 245 - Whoever illegally searches another person's body or residence, or illegally intrudes into another person's residence, shall be sentenced to fixed-term imprisonment of not more than three years or criminal detention. If judicial personnel abuse their power and commit the crime mentioned in the preceding paragraph, they shall be punished more severely. \\
        \midrule
        \vspace{-1mm}
        \cellcolor{'shallow3'} \textbf{\textsc{Answer:}} Prison term: 4 months\\
        \bottomrule
    \end{tabular}
        \caption{
  The instruction and an example of Task 3-5 Prison Term Prediction w. Article.
    }
    \label{tab:instruction_3_5}
\end{table*}
\endgroup
\begingroup
\begin{table*}[ht]

    \centering
    \small
    \begin{tabular}{p{\linewidth}}
        \toprule

            \vspace{-2mm}
        \cellcolor{'shallow3'} \textbf{\textsc{Instruction:}} Please analyze the following case using legal knowledge and select the correct answer from A, B, C, D, and write it between [Correct Answer] and <eoa>. For example, [Correct Answer] A <eoa>. Please adhere strictly to this format in your response. \\
        \midrule
        \vspace{-1mm}
        \cellcolor{'shallow3'} \textbf{\textsc{Query:}} During the May Day holiday, tourist A had a conflict with tourist B over the purchase of tickets to the Potala Palace in Tibet and injured tourist B. Upon investigation, it was found that tourist A, who is of Han ethnicity, is from Guangdong province but has been living in Fujian for several years and has some knowledge of Tibetan. In this case, what language should the Lhasa police use to interrogate tourist A? A: Mandarin; B: Fujianese; C: Cantonese; D: Tibetan.\\ 
        \midrule
        \vspace{-1mm}
        \cellcolor{'shallow3'} \textbf{\textsc{Answer:}} Correct answer: D.\\
        \bottomrule
    \end{tabular}
        \caption{
  The instruction and an example of Task 3-6 Case Analysis.
    }
    \label{tab:instruction_3_6}
\end{table*}
\endgroup
\begingroup
\begin{table*}[ht]

    \centering
    \small
    \begin{tabular}{p{\linewidth}}
        \toprule

            \vspace{-2mm}
        \cellcolor{'shallow3'} \textbf{\textsc{Instruction:}} Please calculate the total amount of the crimes mentioned in the document carefully. You don't need to provide the calculation process; just provide the final amount between [Amount] and <eoa>. For example, [Amount] 2000 yuan <eoa>. \\
        \midrule
        \vspace{-1mm}
        \cellcolor{'shallow3'} \textbf{\textsc{Query:}} Document: Upon trial and investigation, it was ascertained that in the early hours of September 23, 2016, the defendant, Liang, drove a motorcycle to the entrance of Hutou Village in Pingma Village Committee, Xiaojiang Street, Pubei County, Guangxi. He stole 11 castrated chickens and 12 hens from the victim, Huang 1, as well as 2 castrated chickens from Huang 2 and 1 hen from Ning. The defendant was apprehended by Huang 1 and others while fleeing the scene and subsequently reported to the public security authorities. According to the appraisal conducted by the Price Certification Center of Pubei County Price Bureau, the total market value of the stolen chickens amounted to RMB 2,094.\\ 
        \midrule
        \vspace{-1mm}
        \cellcolor{'shallow3'} \textbf{\textsc{Answer:}} The amount involved in the above-mentioned crime: 2094.0 yuan.\\
        \bottomrule
    \end{tabular}
        \caption{
  The instruction and an example of Task 3-7 Crime Amount Calculation.
    }
    \label{tab:instruction_3_7}
\end{table*}
\endgroup
\begingroup
\begin{table*}[ht]

    \centering
    \small
    \begin{tabular}{p{\linewidth}}
        \toprule

            \vspace{-2mm}
        \cellcolor{'shallow3'} \textbf{\textsc{Instruction:}}  Please answer the following questions, first provide the answer, and then provide the corresponding legal basis:\\
        \midrule
        \vspace{-1mm}
        \cellcolor{'shallow3'} \textbf{\textsc{Query:}} Is the accident determination related to the driver crossing the centerline and hitting a person, resulting in their death despite rescue efforts? How long should it take to issue the accident determination report?\\ 
        \midrule
        \vspace{-1mm}
        \cellcolor{'shallow3'} \textbf{\textsc{Answer:}} Answer: In the case of a traffic accident resulting in a fatality, the traffic management department of the public security organ should issue the Road Traffic Accident Determination Report within ten days from the date of on-site investigation. For hit-and-run traffic accident cases, the Road Traffic Accident Determination Report should be issued within ten days after the apprehension of the hit-and-run vehicle and driver.

Legal Basis: Article 62 of the 'Provisions on the Procedures for Handling Road Traffic Accidents' states that the traffic management department of the public security organ should issue the Road Traffic Accident Determination Report within ten days from the date of on-site investigation in cases of traffic accidents resulting in fatalities. In hit-and-run cases, the Road Traffic Accident Determination Report should be issued within ten days after the apprehension of the hit-and-run vehicle and driver. In cases requiring examination or appraisal, the Road Traffic Accident Determination Report should be issued within five days from the date of determination based on the examination report or appraisal opinion.\\
        \bottomrule
    \end{tabular}
        \caption{
  The instruction and an example of Task 3-8 Consultation.
    }
    \label{tab:instruction_3_8}
\end{table*}
\endgroup

\clearpage
\subsection{Statics Information of Our Datasets}
\begin{table}[htbp]
  \centering
    \begin{tabular}{crrrrrrrrr}
    \toprule
    \multirow{2}{*}{\textbf{ID}} & \multicolumn{2}{c}{\textbf{Instruction}} & \multicolumn{3}{c}{\textbf{Input}}&\multicolumn{3}{c}{\textbf{Output}}\\
    &\textbf{Zeroshot} & \textbf{Oneshot} & \multicolumn{1}{c}{\textbf{Max}} & \multicolumn{1}{c}{\textbf{Min}} & \multicolumn{1}{c}{\textbf{Avg}} & \multicolumn{1}{c}{\textbf{Max}} & \multicolumn{1}{c}{\textbf{Min}} & \multicolumn{1}{c}{\textbf{Avg}}  \\
    \midrule
    \textbf{1-1} & 18 & 71 & 36 & 13 & 21.742 & 631 & 16 & 103.742 \\
    \textbf{1-2} & 73 & 141 & 429 & 32 & 145.41 & 7 & 7 & 7.0 \\
    \midrule
    \textbf{2-1} & 74 & 133 & 452 & 20 & 79.614 & 452 & 19 & 79.604 \\
    \textbf{2-2} & 159 & 342 & 4364 & 156 & 663.052 & 16 & 10 & 12.534 \\
    \textbf{2-3} & 214 & 256 & 283 & 10 & 61.438 & 48 & 8 & 16.006 \\
    \textbf{2-4} & 174 & 262 & 182 & 6 & 38.21 & 4 & 2 & 3.85  \\
    \textbf{2-5} & 21 & 632 & 1419 & 433 & 734.01 & 149 & 6 & 25.336 \\
    \textbf{2-6} & 81 & 122 & 293 & 13 & 69.758 & 248 & 0 & 57.818 \\
    \textbf{2-7} & 26 & 556 & 369 & 42 & 323.628 & 393 & 27 & 144.236 \\
    \textbf{2-8} & 101 & 389 & 943 & 180 & 382.334 & 12 & 12 & 12.0 \\
    \textbf{2-9} & 159 & 208 & 314 & 7 & 55.308 & 23 & 2 & 5.328 \\
    \textbf{2-10} & 95 & 146 & 281 & 6 & 62.014 & 27 & 1 & 4.912 \\
    \midrule
    \textbf{3-1} & 75 & 182 & 22142 & 51 & 603.968 & 30 & 10 & 11.848 \\
    \textbf{3-2} & 41 & 138 & 154 & 31 & 71.206 & 389 & 34 & 127.55 \\
    \textbf{3-3} & 76 & 182 & 9784 & 26 & 508.15 & 52 & 5 & 13.082 \\
    \textbf{3-4} & 68 & 214 & 22172 & 77 & 615.816 & 8 & 5 & 6.77 \\
    \textbf{3-5} & 68 & 214 & 23421 & 174 & 938.584 & 8 & 5 & 6.77 \\
    \textbf{3-6} & 80 & 170 & 858 & 54 & 212.434 & 7 & 7 & 7.0 \\
    \textbf{3-7} & 72 & 186 & 1980 & 63 & 462.584 & 22 & 17 & 19.386 \\
    \textbf{3-8} & 28 & 185 & 100 & 50 & 57.62 & 731 & 187 & 395.02 \\
    \bottomrule
    \end{tabular}%
    \caption{Detailed statistical information for each task, where Instruction represents the length of instructions, Input represents the length of input questions, and Output represents the length of answers.}
  \label{tab:len_statistics}%
\end{table}%
\section{Details of Results}
\label{appendix:details_of_results}

\begin{table*}
\centering
\small
\resizebox{\columnwidth}{!}{
\begin{tabular}{l|ccccccccc}
\toprule
\multicolumn{1}{l|}{\multirow{2}*{Task}} & 6b & \multicolumn{5}{c}{7b} & \multicolumn{2}{c}{13b} & 33b\\
\multicolumn{1}{l|}{} & lexilaw & lawgpt-beta1.0 & lawgpt-beta1.1 & hanfei-1.0 & fuzi-mingcha & wisdom-INT & chatlaw & lawyer-llama & chatlaw \\
\midrule
1-1 & \cellcolor{Green!0.0}15.47\% & \cellcolor{Green!0.0}1.24\% & \cellcolor{Green!0.0}3.03\% & \cellcolor{Green!0.0}16.23\% & \cellcolor{Green!0.0}20.21\% & \cellcolor{Green!0.0}37.41\% & \cellcolor{Green!0.0}15.98\% & \cellcolor{Green!0.0}13.04\% & \cellcolor{Green!0.0}14.36\% \\
1-2 & \cellcolor{Green!4.2}14.4\% & \cellcolor{Green!99.4}0.0\% & \cellcolor{Green!4.2}6.6\% & \cellcolor{Green!0.8}8.6\% & \cellcolor{Green!5.8}12.8\% & \cellcolor{Green!0.0}15.2\% & \cellcolor{Green!0.0}29.4\% & \cellcolor{Green!23.8}10.6\% & \cellcolor{Green!0.0}27.8\% \\
\hline
2-1 & \cellcolor{Green!0.0}4.18\% & \cellcolor{Green!0.0}2.08\% & \cellcolor{Green!0.0}4.7\% & \cellcolor{Green!0.0}14.98\% & \cellcolor{Green!0.0}2.86\% & \cellcolor{Green!0.0}22.16\% & \cellcolor{Green!0.0}13.01\% & \cellcolor{Green!0.0}4.9\% & \cellcolor{Green!0.0}4.3\% \\
2-2 & \cellcolor{Green!25.0}15.4\% & \cellcolor{Green!83.6}4.0\% & \cellcolor{Green!57.6}3.4\% & \cellcolor{Green!4.2}9.2\% & \cellcolor{Green!21.6}2.4\% & \cellcolor{Green!6.8}10.0\% & \cellcolor{Green!17.0}9.0\% & \cellcolor{Green!16.0}19.2\% & \cellcolor{Green!59.4}12.2\% \\
2-3 & \cellcolor{Green!31.2}21.49\% & \cellcolor{Green!100.0}0.0\% & \cellcolor{Green!53.6}4.61\% & \cellcolor{Green!11.2}32.71\% & \cellcolor{Green!42.4}17.44\% & \cellcolor{Green!0.0}24.29\% & \cellcolor{Green!47.6}30.91\% & \cellcolor{Green!54.4}9.03\% & \cellcolor{Green!23.0}33.49\% \\
2-4 & \cellcolor{Green!20.2}27.2\% & \cellcolor{Green!97.0}0.4\% & \cellcolor{Green!14.8}10.2\% & \cellcolor{Green!11.2}12.6\% & \cellcolor{Green!38.6}8.8\% & \cellcolor{Green!0.6}13.4\% & \cellcolor{Green!12.6}26.6\% & \cellcolor{Green!75.8}3.0\% & \cellcolor{Green!85.2}4.2\% \\
2-5 & \cellcolor{Green!0.0}41.64\% & \cellcolor{Green!0.0}0.79\% & \cellcolor{Green!0.0}2.26\% & \cellcolor{Green!0.0}36.64\% & \cellcolor{Green!0.0}93.35\% & \cellcolor{Green!0.0}13.23\% & \cellcolor{Green!0.0}41.41\% & \cellcolor{Green!0.0}39.65\% & \cellcolor{Green!0.0}38.87\% \\
2-6 & \cellcolor{Green!0.0}31.54\% & \cellcolor{Green!0.0}3.19\% & \cellcolor{Green!0.0}8.11\% & \cellcolor{Green!0.0}52.33\% & \cellcolor{Green!0.0}42.28\% & \cellcolor{Green!0.0}40.1\% & \cellcolor{Green!0.0}60.68\% & \cellcolor{Green!0.0}36.33\% & \cellcolor{Green!0.0}28.83\% \\
2-7 & \cellcolor{Green!0.0}34.57\% & \cellcolor{Green!0.0}0.0\% & \cellcolor{Green!0.0}4.71\% & \cellcolor{Green!0.0}31.15\% & \cellcolor{Green!0.0}31.43\% & \cellcolor{Green!0.0}39.71\% & \cellcolor{Green!0.0}42.71\% & \cellcolor{Green!0.0}37.1\% & \cellcolor{Green!0.0}34.2\% \\
2-8 & \cellcolor{Green!1.4}6.0\% & \cellcolor{Green!99.8}0.0\% & \cellcolor{Green!51.8}0.6\% & \cellcolor{Green!0.2}17.0\% & \cellcolor{Green!1.4}11.4\% & \cellcolor{Green!0.0}0.2\% & \cellcolor{Green!1.8}20.2\% & \cellcolor{Green!6.6}0.4\% & \cellcolor{Green!2.8}15.4\% \\
2-9 & \cellcolor{Green!1.4}19.84 & \cellcolor{Green!99.0}0.7 & \cellcolor{Green!40.0}16.66 & \cellcolor{Green!6.2}25.38 & \cellcolor{Green!23.2}21.26 & \cellcolor{Green!0.0}15.81 & \cellcolor{Green!3.2}40.27 & \cellcolor{Green!7.8}33.19 & \cellcolor{Green!1.4}26.18 \\
2-10 & \cellcolor{Green!0.0}8.36 & \cellcolor{Green!0.0}0.43 & \cellcolor{Green!0.0}0.25 & \cellcolor{Green!0.0}4.09 & \cellcolor{Green!0.0}7.04 & \cellcolor{Green!0.0}4.02 & \cellcolor{Green!0.0}17.37 & \cellcolor{Green!0.0}6.12 & \cellcolor{Green!0.0}15.95 \\
\hline
3-1 & \cellcolor{Green!18.8}15.41\% & \cellcolor{Green!93.4}0.0\% & \cellcolor{Green!17.4}0.49\% & \cellcolor{Green!18.4}7.98\% & \cellcolor{Green!0.2}3.86\% & \cellcolor{Green!0.0}20.02\% & \cellcolor{Green!0.4}25.99\% & \cellcolor{Green!2.4}0.33\% & \cellcolor{Green!6.0}4.89\% \\
3-2 & \cellcolor{Green!0.0}33.94\% & \cellcolor{Green!0.0}23.75\% & \cellcolor{Green!0.0}12.54\% & \cellcolor{Green!0.0}34.27\% & \cellcolor{Green!0.0}32.96\% & \cellcolor{Green!0.0}23.33\% & \cellcolor{Green!0.0}33.96\% & \cellcolor{Green!0.0}27.23\% & \cellcolor{Green!0.0}27.97\% \\
3-3 & \cellcolor{Green!23.0}34.03\% & \cellcolor{Green!71.4}0.8\% & \cellcolor{Green!31.2}13.8\% & \cellcolor{Green!4.8}9.8\% & \cellcolor{Green!11.6}43.6\% & \cellcolor{Green!24.0}39.22\% & \cellcolor{Green!26.0}12.24\% & \cellcolor{Green!22.0}19.36\% & \cellcolor{Green!58.0}17.54\% \\
3-4 & \cellcolor{Green!7.0}73.66\% & \cellcolor{Green!99.0}0.01\% & \cellcolor{Green!44.2}40.18\% & \cellcolor{Green!8.6}74.34\% & \cellcolor{Green!1.4}78.95\% & \cellcolor{Green!1.0}81.16\% & \cellcolor{Green!5.6}74.31\% & \cellcolor{Green!6.0}70.99\% & \cellcolor{Green!20.0}63.3\% \\
3-5 & \cellcolor{Green!10.4}70.93\% & \cellcolor{Green!97.8}0.8\% & \cellcolor{Green!37.0}47.3\% & \cellcolor{Green!10.6}72.19\% & \cellcolor{Green!1.2}79.0\% & \cellcolor{Green!0.8}81.57\% & \cellcolor{Green!5.4}73.01\% & \cellcolor{Green!3.2}73.56\% & \cellcolor{Green!33.0}53.03\% \\
3-6 & \cellcolor{Green!28.4}12.2\% & \cellcolor{Green!97.8}0.0\% & \cellcolor{Green!1.6}17.6\% & \cellcolor{Green!3.6}10.4\% & \cellcolor{Green!20.0}13.8\% & \cellcolor{Green!0.0}13.0\% & \cellcolor{Green!0.0}26.8\% & \cellcolor{Green!59.2}6.6\% & \cellcolor{Green!0.2}26.2\% \\
3-7 & \cellcolor{Green!4.0}33.2\% & \cellcolor{Green!36.2}8.0\% & \cellcolor{Green!7.6}26.0\% & \cellcolor{Green!0.2}13.4\% & \cellcolor{Green!0.0}38.2\% & \cellcolor{Green!1.8}40.4\% & \cellcolor{Green!3.6}42.0\% & \cellcolor{Green!2.6}33.8\% & \cellcolor{Green!10.2}43.2\% \\
3-8 & \cellcolor{Green!0.0}14.68\% & \cellcolor{Green!0.0}0.46\% & \cellcolor{Green!0.0}8.77\% & \cellcolor{Green!0.0}14.99\% & \cellcolor{Green!0.0}13.95\% & \cellcolor{Green!0.0}20.67\% & \cellcolor{Green!0.0}16.72\% & \cellcolor{Green!0.0}16.02\% & \cellcolor{Green!0.0}16.29\% \\
\bottomrule
\end{tabular}}
\begin{tabular}{ccccccc}
	\multicolumn{2}{c}{\%abstention\qquad} & \cellcolor{Green!0.0}\quad0\% \quad & \cellcolor{Green!25}\quad25\% \quad & \cellcolor{Green!50}\quad50\% \quad &
	\cellcolor{Green!75}\quad75\% \quad & \cellcolor{Green!100}\quad100\% \quad \\
\end{tabular}
\caption{One-shot Results of legal specific LLMs}
\end{table*}

\begingroup
\begin{table*}
\centering
\small
\begin{tabular}{llll}
\hline
 model\_name           & 1-1                        & 1-2                         & Memorization                \\
\hline
\textbf{Multilingual LLMs} \\
 MPT-7B               & \cellcolor{Green!0.0}3.57  & \cellcolor{Green!0.4}0.00   & \cellcolor{Green!0.2}1.78   \\
 MPT-Instruct-7B      & \cellcolor{Green!0.0}2.30  & \cellcolor{Green!4.8}1.00   & \cellcolor{Green!2.4}1.65   \\
 LLaMA-7B             & \cellcolor{Green!0.0}2.42  & \cellcolor{Green!10.0}0.60  & \cellcolor{Green!5.0}1.51   \\
 LLaMA-13B            & \cellcolor{Green!0.0}2.64  & \cellcolor{Green!17.0}0.80  & \cellcolor{Green!8.5}1.72   \\
 LLaMA-30B            & \cellcolor{Green!0.0}2.99  & \cellcolor{Green!7.8}0.60   & \cellcolor{Green!3.9}1.80   \\
 LLaMA-65B            & \cellcolor{Green!0.0}1.99  & \cellcolor{Green!8.2}0.40   & \cellcolor{Green!4.1}1.20   \\
 LLaMA-2-7B           & \cellcolor{Green!0.0}2.65  & \cellcolor{Green!3.4}0.40   & \cellcolor{Green!1.7}1.52   \\
 LLaMA-2-13B          & \cellcolor{Green!0.0}1.90  & \cellcolor{Green!1.8}1.80   & \cellcolor{Green!0.9}1.85   \\
 LLaMA-2-70B          & \cellcolor{Green!0.0}3.15  & \cellcolor{Green!0.2}0.40   & \cellcolor{Green!0.1}1.77   \\
 LLaMA-2-Chat-7B      & \cellcolor{Green!0.0}0.80  & \cellcolor{Green!1.8}6.40   & \cellcolor{Green!0.9}3.60   \\
 LLaMA-2-Chat-13B     & \cellcolor{Green!0.0}1.05  & \cellcolor{Green!1.4}3.20   & \cellcolor{Green!0.7}2.12   \\
 LLaMA-2-Chat-70B     & \cellcolor{Green!0.0}0.87  & \cellcolor{Green!0.0}10.00  & \cellcolor{Green!0.0}5.43   \\
 Alpaca-v1.0-7B       & \cellcolor{Green!0.0}1.41  & \cellcolor{Green!49.0}0.20  & \cellcolor{Green!24.5}0.80  \\
 Vicuna-v1.3-7B       & \cellcolor{Green!0.0}6.40  & \cellcolor{Green!1.6}8.80   & \cellcolor{Green!0.8}7.60   \\
 Vicuna-v1.3-13B      & \cellcolor{Green!0.0}7.21  & \cellcolor{Green!0.2}18.60  & \cellcolor{Green!0.1}12.91  \\
 Vicuna-v1.3-33B      & \cellcolor{Green!0.0}10.45 & \cellcolor{Green!3.0}24.40  & \cellcolor{Green!1.5}17.43  \\
 WizardLM-7B          & \cellcolor{Green!0.0}0.48  & \cellcolor{Green!0.8}13.80  & \cellcolor{Green!0.4}7.14   \\
 StableBeluga2        & \cellcolor{Green!0.0}14.58 & \cellcolor{Green!0.0}34.60  & \cellcolor{Green!0.0}24.59  \\
 ChatGPT              & \cellcolor{Green!0.0}15.86 & \cellcolor{Green!0.0}36.00  & \cellcolor{Green!0.0}25.93  \\
 GPT-4                & \cellcolor{Green!0.0}15.38 & \cellcolor{Green!0.2}55.20  & \cellcolor{Green!0.1}35.29  \\
\midrule
\textbf{Chinese Oriented LLMs} \\
 MOSS-Moon            & \cellcolor{Green!0.0}5.04  & \cellcolor{Green!7.4}11.60  & \cellcolor{Green!3.7}8.32   \\
 MOSS-Moon-SFT        & \cellcolor{Green!0.0}2.89  & \cellcolor{Green!8.8}7.60   & \cellcolor{Green!4.4}5.25   \\
 TigerBot-Base        & \cellcolor{Green!0.0}15.62 & \cellcolor{Green!81.6}3.80  & \cellcolor{Green!40.8}9.71  \\
 TigerBot-SFT         & \cellcolor{Green!0.0}11.18 & \cellcolor{Green!5.4}13.60  & \cellcolor{Green!2.7}12.39  \\
 GoGPT                & \cellcolor{Green!0.0}6.71  & \cellcolor{Green!31.0}17.40 & \cellcolor{Green!15.5}12.05 \\
 ChatGLM2-6B          & \cellcolor{Green!0.0}15.37 & \cellcolor{Green!1.8}10.60  & \cellcolor{Green!0.9}12.98  \\
 Ziya-LLaMA-13B       & \cellcolor{Green!0.0}15.04 & \cellcolor{Green!2.0}30.00  & \cellcolor{Green!1.0}22.52  \\
 Baichuan-7B          & \cellcolor{Green!0.0}0.03  & \cellcolor{Green!35.8}1.80  & \cellcolor{Green!17.9}0.92  \\
 Baichuan-13B         & \cellcolor{Green!0.0}1.29  & \cellcolor{Green!44.6}11.40 & \cellcolor{Green!22.3}6.35  \\
 Baichuan-13B-Chat    & \cellcolor{Green!0.0}12.32 & \cellcolor{Green!15.2}34.80 & \cellcolor{Green!7.6}23.56  \\
 XVERSE-13B           & \cellcolor{Green!0.0}1.71  & \cellcolor{Green!1.2}0.00   & \cellcolor{Green!0.6}0.85   \\
 InternLM-7B          & \cellcolor{Green!0.0}1.56  & \cellcolor{Green!0.2}4.60   & \cellcolor{Green!0.1}3.08   \\
 InternLM-Chat-7B     & \cellcolor{Green!0.0}14.39 & \cellcolor{Green!0.0}40.00  & \cellcolor{Green!0.0}27.20  \\
 InternLM-Chat-7B-8K  & \cellcolor{Green!0.0}15.45 & \cellcolor{Green!0.0}40.40  & \cellcolor{Green!0.0}27.93  \\
 Qwen-7B              & \cellcolor{Green!0.0}0.00  & \cellcolor{Green!93.4}0.00  & \cellcolor{Green!46.7}0.00  \\
 Qwen-7B-Chat         & \cellcolor{Green!0.0}18.54 & \cellcolor{Green!0.2}34.00  & \cellcolor{Green!0.1}26.27  \\
 Yulan-Chat-2-13B     & \cellcolor{Green!0.0}15.03 & \cellcolor{Green!0.0}22.40  & \cellcolor{Green!0.0}18.72  \\
 BELLE-LLaMA-2        & \cellcolor{Green!0.0}14.39 & \cellcolor{Green!0.2}30.60  & \cellcolor{Green!0.1}22.50  \\
 Chinese-LLaMA-2-7B   & \cellcolor{Green!0.0}3.24  & \cellcolor{Green!74.0}1.20  & \cellcolor{Green!37.0}2.22  \\
 Chinese-Alpaca-2-7B  & \cellcolor{Green!0.0}14.95 & \cellcolor{Green!6.8}4.40   & \cellcolor{Green!3.4}9.67   \\
 LLaMA-2-Chinese-7B   & \cellcolor{Green!0.0}0.00  & \cellcolor{Green!99.2}0.00  & \cellcolor{Green!49.6}0.00  \\
 LLaMA-2-Chinese-13B  & \cellcolor{Green!0.0}0.00  & \cellcolor{Green!100.0}0.00 & \cellcolor{Green!50.0}0.00  \\
\midrule
\textbf{Law Specific LLMs} \\
 LexiLaw              & \cellcolor{Green!0.0}16.96 & \cellcolor{Green!3.4}21.00  & \cellcolor{Green!1.7}18.98  \\
 Fuzi-Mingcha         & \cellcolor{Green!0.0}25.22 & \cellcolor{Green!11.4}7.80  & \cellcolor{Green!5.7}16.51  \\
 HanFei               & \cellcolor{Green!0.0}17.03 & \cellcolor{Green!1.2}19.80  & \cellcolor{Green!0.6}18.42  \\
 LaWGPT-7B-beta1.0    & \cellcolor{Green!0.0}9.34  & \cellcolor{Green!100.0}0.00 & \cellcolor{Green!50.0}4.67  \\
 LaWGPT-7B-beta1.1    & \cellcolor{Green!0.0}4.53  & \cellcolor{Green!13.6}4.20  & \cellcolor{Green!6.8}4.37   \\
 Wisdom-Interrogatory & \cellcolor{Green!0.0}43.05 & \cellcolor{Green!1.2}15.40  & \cellcolor{Green!0.6}29.23  \\
 ChatLaw-13B          & \cellcolor{Green!0.0}14.85 & \cellcolor{Green!0.4}28.40  & \cellcolor{Green!0.2}21.63  \\
 Lawyer-LLaMA         & \cellcolor{Green!0.0}12.33 & \cellcolor{Green!8.2}23.20  & \cellcolor{Green!4.1}17.77  \\
 ChatLaw-33B          & \cellcolor{Green!0.0}11.74 & \cellcolor{Green!0.0}28.60  & \cellcolor{Green!0.0}20.17  \\
\hline
\end{tabular}

\begin{tabular}{ccccccc}
	\multicolumn{2}{c}{\%abstention\qquad} & \cellcolor{Green!0.0}\quad0\% \quad & \cellcolor{Green!25}\quad25\% \quad & \cellcolor{Green!50}\quad50\% \quad &
	\cellcolor{Green!75}\quad75\% \quad & \cellcolor{Green!100}\quad100\% \quad \
\end{tabular}
\caption{Zero-shot Results on Legal Knowledge Memorization Tasks}
\label{tab:Zero-shot Results on Legal Knowledge Memorization Tasks}

\end{table*}
\endgroup

\begingroup
\begin{table*}
\centering
\small
\resizebox{\columnwidth}{!}{
\begin{tabular}{llllllllllll}
\hline
 model\_name           & 2-1                        & 2-2                         & 2-3                         & 2-4                         & 2-5                        & 2-6                        & 2-7                        & 2-8                         & 2-9                         & 2-10                       & Understanding               \\
\hline
\textbf{Multilingual LLMs} \\
 MPT-7B               & \cellcolor{Green!0.0}0.71  & \cellcolor{Green!67.4}7.42  & \cellcolor{Green!38.0}12.50 & \cellcolor{Green!78.6}2.00  & \cellcolor{Green!0.0}3.39  & \cellcolor{Green!0.0}6.21  & \cellcolor{Green!0.0}14.82 & \cellcolor{Green!3.4}2.40   & \cellcolor{Green!58.2}11.81 & \cellcolor{Green!0.0}0.32  & \cellcolor{Green!24.6}6.16  \\
 MPT-Instruct-7B      & \cellcolor{Green!0.0}0.93  & \cellcolor{Green!72.0}4.54  & \cellcolor{Green!35.0}11.86 & \cellcolor{Green!38.0}10.80 & \cellcolor{Green!0.0}4.21  & \cellcolor{Green!0.0}17.85 & \cellcolor{Green!0.0}4.38  & \cellcolor{Green!38.6}3.80  & \cellcolor{Green!68.0}9.66  & \cellcolor{Green!0.0}0.66  & \cellcolor{Green!25.2}6.87  \\
 LLaMA-7B             & \cellcolor{Green!0.0}0.43  & \cellcolor{Green!76.4}5.77  & \cellcolor{Green!44.8}12.14 & \cellcolor{Green!68.0}3.20  & \cellcolor{Green!0.0}4.57  & \cellcolor{Green!0.0}4.66  & \cellcolor{Green!0.0}15.31 & \cellcolor{Green!9.4}1.00   & \cellcolor{Green!67.8}10.61 & \cellcolor{Green!0.0}0.35  & \cellcolor{Green!26.6}5.80  \\
 LLaMA-13B            & \cellcolor{Green!0.0}0.38  & \cellcolor{Green!70.0}6.39  & \cellcolor{Green!36.4}13.00 & \cellcolor{Green!76.4}2.20  & \cellcolor{Green!0.0}4.89  & \cellcolor{Green!0.0}5.59  & \cellcolor{Green!0.0}17.12 & \cellcolor{Green!9.6}0.20   & \cellcolor{Green!48.6}11.90 & \cellcolor{Green!0.0}0.35  & \cellcolor{Green!24.1}6.20  \\
 LLaMA-30B            & \cellcolor{Green!0.0}0.44  & \cellcolor{Green!45.0}4.74  & \cellcolor{Green!61.2}7.52  & \cellcolor{Green!81.4}2.20  & \cellcolor{Green!0.0}3.95  & \cellcolor{Green!0.0}2.33  & \cellcolor{Green!0.0}2.66  & \cellcolor{Green!11.4}0.60  & \cellcolor{Green!49.2}9.73  & \cellcolor{Green!0.0}0.39  & \cellcolor{Green!24.8}3.46  \\
 LLaMA-65B            & \cellcolor{Green!0.0}0.48  & \cellcolor{Green!55.8}5.15  & \cellcolor{Green!74.6}5.35  & \cellcolor{Green!95.0}1.00  & \cellcolor{Green!0.0}5.31  & \cellcolor{Green!0.0}2.23  & \cellcolor{Green!0.0}3.12  & \cellcolor{Green!24.6}0.20  & \cellcolor{Green!38.0}12.70 & \cellcolor{Green!0.0}0.42  & \cellcolor{Green!28.8}3.60  \\
 LLaMA-2-7B           & \cellcolor{Green!0.0}0.89  & \cellcolor{Green!64.8}3.92  & \cellcolor{Green!32.4}20.55 & \cellcolor{Green!83.6}2.20  & \cellcolor{Green!0.0}5.68  & \cellcolor{Green!0.0}29.27 & \cellcolor{Green!0.0}16.58 & \cellcolor{Green!6.4}0.20   & \cellcolor{Green!49.0}11.84 & \cellcolor{Green!0.0}0.38  & \cellcolor{Green!23.6}9.15  \\
 LLaMA-2-13B          & \cellcolor{Green!0.0}0.73  & \cellcolor{Green!65.4}6.19  & \cellcolor{Green!38.6}20.90 & \cellcolor{Green!95.6}1.80  & \cellcolor{Green!0.0}5.30  & \cellcolor{Green!0.0}16.36 & \cellcolor{Green!0.0}12.05 & \cellcolor{Green!7.0}0.60   & \cellcolor{Green!53.2}10.13 & \cellcolor{Green!0.0}0.72  & \cellcolor{Green!26.0}7.48  \\
 LLaMA-2-70B          & \cellcolor{Green!0.0}2.88  & \cellcolor{Green!38.4}8.87  & \cellcolor{Green!28.6}26.85 & \cellcolor{Green!22.0}10.40 & \cellcolor{Green!0.0}5.59  & \cellcolor{Green!0.0}42.57 & \cellcolor{Green!0.0}16.54 & \cellcolor{Green!0.8}0.00   & \cellcolor{Green!48.2}14.55 & \cellcolor{Green!0.0}0.87  & \cellcolor{Green!13.8}12.91 \\
 LLaMA-2-Chat-7B      & \cellcolor{Green!0.0}0.25  & \cellcolor{Green!32.0}5.36  & \cellcolor{Green!41.8}16.92 & \cellcolor{Green!46.6}5.20  & \cellcolor{Green!0.0}4.90  & \cellcolor{Green!0.0}5.62  & \cellcolor{Green!0.0}1.13  & \cellcolor{Green!1.2}6.20   & \cellcolor{Green!0.8}14.77  & \cellcolor{Green!0.0}2.06  & \cellcolor{Green!12.2}6.24  \\
 LLaMA-2-Chat-13B     & \cellcolor{Green!0.0}0.84  & \cellcolor{Green!24.2}9.48  & \cellcolor{Green!2.6}40.80  & \cellcolor{Green!10.2}24.80 & \cellcolor{Green!0.0}6.08  & \cellcolor{Green!0.0}2.20  & \cellcolor{Green!0.0}1.16  & \cellcolor{Green!0.0}9.20   & \cellcolor{Green!19.0}35.74 & \cellcolor{Green!0.0}4.25  & \cellcolor{Green!5.6}13.46  \\
 LLaMA-2-Chat-70B     & \cellcolor{Green!0.0}0.74  & \cellcolor{Green!35.4}7.22  & \cellcolor{Green!8.4}33.38  & \cellcolor{Green!1.4}30.40  & \cellcolor{Green!0.0}5.22  & \cellcolor{Green!0.0}34.72 & \cellcolor{Green!0.0}1.25  & \cellcolor{Green!0.0}8.40   & \cellcolor{Green!21.4}39.75 & \cellcolor{Green!0.0}12.61 & \cellcolor{Green!6.7}17.37  \\
 Alpaca-v1.0-7B       & \cellcolor{Green!0.0}0.30  & \cellcolor{Green!78.8}4.33  & \cellcolor{Green!58.6}6.91  & \cellcolor{Green!41.2}3.80  & \cellcolor{Green!0.0}15.96 & \cellcolor{Green!0.0}4.14  & \cellcolor{Green!0.0}3.85  & \cellcolor{Green!97.0}0.00  & \cellcolor{Green!83.6}5.94  & \cellcolor{Green!0.0}1.07  & \cellcolor{Green!35.9}4.63  \\
 Vicuna-v1.3-7B       & \cellcolor{Green!0.0}1.00  & \cellcolor{Green!45.8}6.60  & \cellcolor{Green!16.8}21.56 & \cellcolor{Green!21.0}7.40  & \cellcolor{Green!0.0}4.54  & \cellcolor{Green!0.0}40.92 & \cellcolor{Green!0.0}4.00  & \cellcolor{Green!49.2}2.60  & \cellcolor{Green!41.0}9.62  & \cellcolor{Green!0.0}3.38  & \cellcolor{Green!17.4}10.16 \\
 Vicuna-v1.3-13B      & \cellcolor{Green!0.0}1.68  & \cellcolor{Green!44.8}3.09  & \cellcolor{Green!17.2}24.54 & \cellcolor{Green!13.0}11.20 & \cellcolor{Green!0.0}21.82 & \cellcolor{Green!0.0}5.25  & \cellcolor{Green!0.0}3.79  & \cellcolor{Green!11.2}11.00 & \cellcolor{Green!19.4}17.30 & \cellcolor{Green!0.0}4.75  & \cellcolor{Green!10.6}10.44 \\
 Vicuna-v1.3-33B      & \cellcolor{Green!0.0}1.76  & \cellcolor{Green!45.0}6.39  & \cellcolor{Green!17.2}19.30 & \cellcolor{Green!55.4}8.20  & \cellcolor{Green!0.0}12.51 & \cellcolor{Green!0.0}18.38 & \cellcolor{Green!0.0}3.35  & \cellcolor{Green!39.0}16.40 & \cellcolor{Green!52.8}9.75  & \cellcolor{Green!0.0}5.88  & \cellcolor{Green!20.9}10.19 \\
 WizardLM-7B          & \cellcolor{Green!0.0}0.10  & \cellcolor{Green!75.0}1.86  & \cellcolor{Green!9.8}18.32  & \cellcolor{Green!8.2}7.00   & \cellcolor{Green!0.0}13.54 & \cellcolor{Green!0.0}32.44 & \cellcolor{Green!0.0}4.24  & \cellcolor{Green!22.2}4.60  & \cellcolor{Green!29.0}10.99 & \cellcolor{Green!0.0}3.32  & \cellcolor{Green!14.4}9.64  \\
 StableBeluga2        & \cellcolor{Green!0.0}7.70  & \cellcolor{Green!26.0}25.57 & \cellcolor{Green!5.4}44.20  & \cellcolor{Green!1.8}39.00  & \cellcolor{Green!0.0}52.03 & \cellcolor{Green!0.0}65.54 & \cellcolor{Green!0.0}39.07 & \cellcolor{Green!0.2}45.80  & \cellcolor{Green!3.6}65.27  & \cellcolor{Green!0.0}41.64 & \cellcolor{Green!3.7}42.58  \\
 ChatGPT              & \cellcolor{Green!0.0}9.10  & \cellcolor{Green!3.0}32.37  & \cellcolor{Green!1.0}51.73  & \cellcolor{Green!3.4}41.20  & \cellcolor{Green!0.0}53.75 & \cellcolor{Green!0.0}69.55 & \cellcolor{Green!0.0}33.49 & \cellcolor{Green!0.0}36.40  & \cellcolor{Green!4.8}66.48  & \cellcolor{Green!0.0}39.05 & \cellcolor{Green!1.2}43.31  \\
 GPT-4                & \cellcolor{Green!0.0}12.53 & \cellcolor{Green!0.0}41.65  & \cellcolor{Green!0.0}69.79  & \cellcolor{Green!1.6}44.00  & \cellcolor{Green!0.0}56.50 & \cellcolor{Green!0.0}76.60 & \cellcolor{Green!0.0}37.92 & \cellcolor{Green!0.0}61.20  & \cellcolor{Green!2.2}78.82  & \cellcolor{Green!0.0}65.09 & \cellcolor{Green!0.4}54.41  \\
\midrule
\textbf{Chinese Oriented LLMs} \\
 MOSS-Moon            & \cellcolor{Green!0.0}3.00  & \cellcolor{Green!60.2}2.68  & \cellcolor{Green!28.8}12.69 & \cellcolor{Green!87.2}1.80  & \cellcolor{Green!0.0}9.24  & \cellcolor{Green!0.0}6.17  & \cellcolor{Green!0.0}10.72 & \cellcolor{Green!18.4}8.20  & \cellcolor{Green!45.2}12.07 & \cellcolor{Green!0.0}1.42  & \cellcolor{Green!24.0}6.80  \\
 MOSS-Moon-SFT        & \cellcolor{Green!0.0}6.84  & \cellcolor{Green!34.2}5.15  & \cellcolor{Green!1.4}17.94  & \cellcolor{Green!10.0}3.20  & \cellcolor{Green!0.0}2.55  & \cellcolor{Green!0.0}10.84 & \cellcolor{Green!0.0}6.37  & \cellcolor{Green!6.2}3.00   & \cellcolor{Green!4.8}13.96  & \cellcolor{Green!0.0}1.67  & \cellcolor{Green!5.7}7.15   \\
 TigerBot-Base        & \cellcolor{Green!0.0}12.99 & \cellcolor{Green!65.8}5.57  & \cellcolor{Green!41.8}10.72 & \cellcolor{Green!12.2}22.20 & \cellcolor{Green!0.0}24.33 & \cellcolor{Green!0.0}27.48 & \cellcolor{Green!0.0}29.16 & \cellcolor{Green!2.4}13.80  & \cellcolor{Green!10.6}18.59 & \cellcolor{Green!0.0}13.99 & \cellcolor{Green!13.3}17.88 \\
 TigerBot-SFT         & \cellcolor{Green!0.0}10.43 & \cellcolor{Green!5.0}8.04   & \cellcolor{Green!4.0}20.88  & \cellcolor{Green!6.2}18.60  & \cellcolor{Green!0.0}42.10 & \cellcolor{Green!0.0}63.60 & \cellcolor{Green!0.0}37.10 & \cellcolor{Green!12.4}2.80  & \cellcolor{Green!0.6}20.35  & \cellcolor{Green!0.0}10.57 & \cellcolor{Green!2.8}23.45  \\
 GoGPT                & \cellcolor{Green!0.0}2.46  & \cellcolor{Green!89.4}1.44  & \cellcolor{Green!39.4}22.63 & \cellcolor{Green!51.6}9.00  & \cellcolor{Green!0.0}15.55 & \cellcolor{Green!0.0}26.02 & \cellcolor{Green!0.0}10.48 & \cellcolor{Green!100.0}0.00 & \cellcolor{Green!89.4}1.36  & \cellcolor{Green!0.0}0.73  & \cellcolor{Green!37.0}8.97  \\
 ChatGLM2-6B          & \cellcolor{Green!0.0}1.84  & \cellcolor{Green!12.4}1.65  & \cellcolor{Green!47.4}13.20 & \cellcolor{Green!24.8}22.40 & \cellcolor{Green!0.0}30.33 & \cellcolor{Green!0.0}36.81 & \cellcolor{Green!0.0}19.50 & \cellcolor{Green!0.2}4.00   & \cellcolor{Green!45.6}11.60 & \cellcolor{Green!0.0}2.81  & \cellcolor{Green!13.0}14.41 \\
 Ziya-LLaMA-13B       & \cellcolor{Green!0.0}8.13  & \cellcolor{Green!84.2}2.06  & \cellcolor{Green!9.6}44.18  & \cellcolor{Green!34.0}24.60 & \cellcolor{Green!0.0}27.93 & \cellcolor{Green!0.0}43.44 & \cellcolor{Green!0.0}2.00  & \cellcolor{Green!26.4}11.80 & \cellcolor{Green!9.4}31.82  & \cellcolor{Green!0.0}3.06  & \cellcolor{Green!16.4}19.90 \\
 Baichuan-7B          & \cellcolor{Green!0.0}4.65  & \cellcolor{Green!71.0}4.95  & \cellcolor{Green!17.2}21.96 & \cellcolor{Green!87.4}2.80  & \cellcolor{Green!0.0}10.07 & \cellcolor{Green!0.0}14.82 & \cellcolor{Green!0.0}3.69  & \cellcolor{Green!29.2}5.40  & \cellcolor{Green!39.8}11.91 & \cellcolor{Green!0.0}1.04  & \cellcolor{Green!24.5}8.13  \\
 Baichuan-13B         & \cellcolor{Green!0.0}8.56  & \cellcolor{Green!47.4}6.60  & \cellcolor{Green!18.2}35.65 & \cellcolor{Green!83.2}3.60  & \cellcolor{Green!0.0}6.44  & \cellcolor{Green!0.0}26.76 & \cellcolor{Green!0.0}7.92  & \cellcolor{Green!1.4}13.40  & \cellcolor{Green!26.0}12.19 & \cellcolor{Green!0.0}2.44  & \cellcolor{Green!17.6}12.35 \\
 Baichuan-13B-Chat    & \cellcolor{Green!0.0}4.80  & \cellcolor{Green!17.0}12.99 & \cellcolor{Green!22.4}32.07 & \cellcolor{Green!94.0}2.60  & \cellcolor{Green!0.0}21.84 & \cellcolor{Green!0.0}55.11 & \cellcolor{Green!0.0}1.34  & \cellcolor{Green!99.2}0.00  & \cellcolor{Green!95.4}1.41  & \cellcolor{Green!0.0}2.77  & \cellcolor{Green!32.8}13.49 \\
 XVERSE-13B           & \cellcolor{Green!0.0}0.28  & \cellcolor{Green!16.4}0.00  & \cellcolor{Green!1.4}15.76  & \cellcolor{Green!51.0}1.80  & \cellcolor{Green!0.0}3.47  & \cellcolor{Green!0.0}2.20  & \cellcolor{Green!0.0}16.67 & \cellcolor{Green!0.0}0.00   & \cellcolor{Green!0.2}13.64  & \cellcolor{Green!0.0}0.85  & \cellcolor{Green!6.9}5.47   \\
 InternLM-7B          & \cellcolor{Green!0.0}6.88  & \cellcolor{Green!61.8}4.74  & \cellcolor{Green!4.2}18.31  & \cellcolor{Green!87.6}2.00  & \cellcolor{Green!0.0}5.20  & \cellcolor{Green!0.0}25.31 & \cellcolor{Green!0.0}6.71  & \cellcolor{Green!4.4}0.40   & \cellcolor{Green!47.4}12.27 & \cellcolor{Green!0.0}0.60  & \cellcolor{Green!20.5}8.24  \\
 InternLM-Chat-7B     & \cellcolor{Green!0.0}21.07 & \cellcolor{Green!6.6}33.40  & \cellcolor{Green!27.0}27.09 & \cellcolor{Green!7.8}35.60  & \cellcolor{Green!0.0}55.62 & \cellcolor{Green!0.0}12.73 & \cellcolor{Green!0.0}25.87 & \cellcolor{Green!0.0}27.60  & \cellcolor{Green!9.6}60.21  & \cellcolor{Green!0.0}20.43 & \cellcolor{Green!5.1}31.96  \\
 InternLM-Chat-7B-8K  & \cellcolor{Green!0.0}22.64 & \cellcolor{Green!6.8}35.46  & \cellcolor{Green!14.2}28.96 & \cellcolor{Green!7.8}35.60  & \cellcolor{Green!0.0}54.13 & \cellcolor{Green!0.0}17.95 & \cellcolor{Green!0.0}27.11 & \cellcolor{Green!0.0}36.20  & \cellcolor{Green!10.6}62.93 & \cellcolor{Green!0.0}20.94 & \cellcolor{Green!3.9}34.19  \\
 Qwen-7B              & \cellcolor{Green!0.0}5.09  & \cellcolor{Green!75.6}4.95  & \cellcolor{Green!45.2}7.81  & \cellcolor{Green!26.4}24.80 & \cellcolor{Green!0.0}31.66 & \cellcolor{Green!0.0}38.24 & \cellcolor{Green!0.0}1.67  & \cellcolor{Green!48.2}7.40  & \cellcolor{Green!34.0}26.09 & \cellcolor{Green!0.0}8.39  & \cellcolor{Green!22.9}15.61 \\
 Qwen-7B-Chat         & \cellcolor{Green!0.0}22.56 & \cellcolor{Green!28.4}27.42 & \cellcolor{Green!19.4}31.42 & \cellcolor{Green!13.0}35.00 & \cellcolor{Green!0.0}48.48 & \cellcolor{Green!0.0}37.88 & \cellcolor{Green!0.0}36.04 & \cellcolor{Green!0.0}24.00  & \cellcolor{Green!20.2}44.88 & \cellcolor{Green!0.0}18.90 & \cellcolor{Green!8.1}32.66  \\
 Yulan-Chat-2-13B     & \cellcolor{Green!0.0}6.99  & \cellcolor{Green!12.4}20.00 & \cellcolor{Green!13.8}35.20 & \cellcolor{Green!9.4}33.80  & \cellcolor{Green!0.0}44.75 & \cellcolor{Green!0.0}56.18 & \cellcolor{Green!0.0}38.06 & \cellcolor{Green!0.0}26.00  & \cellcolor{Green!24.0}37.82 & \cellcolor{Green!0.0}23.13 & \cellcolor{Green!6.0}32.19  \\
 BELLE-LLaMA-2        & \cellcolor{Green!0.0}5.78  & \cellcolor{Green!25.6}9.07  & \cellcolor{Green!1.6}38.84  & \cellcolor{Green!2.0}38.00  & \cellcolor{Green!0.0}41.56 & \cellcolor{Green!0.0}54.89 & \cellcolor{Green!0.0}36.59 & \cellcolor{Green!2.4}20.80  & \cellcolor{Green!2.6}34.73  & \cellcolor{Green!0.0}16.83 & \cellcolor{Green!3.4}29.71  \\
 Chinese-LLaMA-2-7B   & \cellcolor{Green!0.0}1.58  & \cellcolor{Green!78.4}4.95  & \cellcolor{Green!89.6}1.33  & \cellcolor{Green!93.0}1.60  & \cellcolor{Green!0.0}1.87  & \cellcolor{Green!0.0}20.24 & \cellcolor{Green!0.0}9.86  & \cellcolor{Green!87.0}0.60  & \cellcolor{Green!68.8}9.17  & \cellcolor{Green!0.0}0.36  & \cellcolor{Green!41.7}5.16  \\
 Chinese-Alpaca-2-7B  & \cellcolor{Green!0.0}4.11  & \cellcolor{Green!70.4}3.71  & \cellcolor{Green!9.2}31.12  & \cellcolor{Green!16.4}20.20 & \cellcolor{Green!0.0}0.55  & \cellcolor{Green!0.0}51.91 & \cellcolor{Green!0.0}2.92  & \cellcolor{Green!98.8}0.00  & \cellcolor{Green!11.8}31.62 & \cellcolor{Green!0.0}5.74  & \cellcolor{Green!20.7}15.19 \\
 LLaMA-2-Chinese-7B   & \cellcolor{Green!0.0}0.12  & \cellcolor{Green!91.0}1.24  & \cellcolor{Green!91.4}0.24  & \cellcolor{Green!99.4}0.20  & \cellcolor{Green!0.0}6.91  & \cellcolor{Green!0.0}2.33  & \cellcolor{Green!0.0}2.00  & \cellcolor{Green!98.2}0.00  & \cellcolor{Green!83.4}3.76  & \cellcolor{Green!0.0}1.28  & \cellcolor{Green!46.3}1.81  \\
 LLaMA-2-Chinese-13B  & \cellcolor{Green!0.0}0.24  & \cellcolor{Green!88.8}1.86  & \cellcolor{Green!93.6}0.48  & \cellcolor{Green!99.6}0.00  & \cellcolor{Green!0.0}5.18  & \cellcolor{Green!0.0}2.20  & \cellcolor{Green!0.0}1.05  & \cellcolor{Green!100.0}0.00 & \cellcolor{Green!85.8}1.97  & \cellcolor{Green!0.0}0.78  & \cellcolor{Green!46.8}1.38  \\
\midrule
\textbf{Law Specific LLMs} \\
 LexiLaw              & \cellcolor{Green!0.0}6.24  & \cellcolor{Green!10.4}3.30  & \cellcolor{Green!52.8}15.60 & \cellcolor{Green!23.8}22.80 & \cellcolor{Green!0.0}45.39 & \cellcolor{Green!0.0}48.74 & \cellcolor{Green!0.0}33.12 & \cellcolor{Green!3.0}21.60  & \cellcolor{Green!0.6}15.30  & \cellcolor{Green!0.0}11.17 & \cellcolor{Green!9.1}22.32  \\
 Fuzi-Mingcha         & \cellcolor{Green!0.0}4.93  & \cellcolor{Green!23.4}19.59 & \cellcolor{Green!27.2}28.46 & \cellcolor{Green!29.4}18.60 & \cellcolor{Green!0.0}97.59 & \cellcolor{Green!0.0}44.07 & \cellcolor{Green!0.0}54.32 & \cellcolor{Green!20.6}8.80  & \cellcolor{Green!2.4}16.90  & \cellcolor{Green!0.0}7.78  & \cellcolor{Green!10.3}30.10 \\
 HanFei               & \cellcolor{Green!0.0}19.40 & \cellcolor{Green!2.8}6.39   & \cellcolor{Green!20.2}30.44 & \cellcolor{Green!19.2}30.20 & \cellcolor{Green!0.0}38.50 & \cellcolor{Green!0.0}59.23 & \cellcolor{Green!0.0}40.07 & \cellcolor{Green!0.2}23.40  & \cellcolor{Green!0.4}14.73  & \cellcolor{Green!0.0}6.05  & \cellcolor{Green!4.3}26.84  \\
 LaWGPT-7B-beta1.0    & \cellcolor{Green!0.0}3.41  & \cellcolor{Green!93.0}0.62  & \cellcolor{Green!100.0}0.00 & \cellcolor{Green!99.4}0.00  & \cellcolor{Green!0.0}0.01  & \cellcolor{Green!0.0}2.20  & \cellcolor{Green!0.0}0.19  & \cellcolor{Green!100.0}0.00 & \cellcolor{Green!99.8}0.13  & \cellcolor{Green!0.0}0.14  & \cellcolor{Green!49.2}0.67  \\
 LaWGPT-7B-beta1.1    & \cellcolor{Green!0.0}6.22  & \cellcolor{Green!81.2}4.95  & \cellcolor{Green!61.8}6.85  & \cellcolor{Green!86.6}2.40  & \cellcolor{Green!0.0}2.27  & \cellcolor{Green!0.0}2.00  & \cellcolor{Green!0.0}8.61  & \cellcolor{Green!71.8}1.00  & \cellcolor{Green!41.8}14.94 & \cellcolor{Green!0.0}1.06  & \cellcolor{Green!34.3}5.03  \\
 Wisdom-Interrogatory & \cellcolor{Green!0.0}30.97 & \cellcolor{Green!5.4}7.84   & \cellcolor{Green!20.6}36.72 & \cellcolor{Green!15.4}21.00 & \cellcolor{Green!0.0}35.56 & \cellcolor{Green!0.0}57.06 & \cellcolor{Green!0.0}33.34 & \cellcolor{Green!0.0}10.60  & \cellcolor{Green!1.0}15.98  & \cellcolor{Green!0.0}6.24  & \cellcolor{Green!4.2}25.53  \\
 ChatLaw-13B          & \cellcolor{Green!0.0}12.22 & \cellcolor{Green!11.6}2.68  & \cellcolor{Green!27.0}42.24 & \cellcolor{Green!15.8}27.60 & \cellcolor{Green!0.0}39.11 & \cellcolor{Green!0.0}54.89 & \cellcolor{Green!0.0}38.45 & \cellcolor{Green!1.0}18.60  & \cellcolor{Green!1.0}31.74  & \cellcolor{Green!0.0}14.56 & \cellcolor{Green!5.6}28.21  \\
 Lawyer-LLaMA         & \cellcolor{Green!0.0}4.33  & \cellcolor{Green!6.8}8.25   & \cellcolor{Green!38.6}15.88 & \cellcolor{Green!78.6}4.40  & \cellcolor{Green!0.0}34.61 & \cellcolor{Green!0.0}41.65 & \cellcolor{Green!0.0}38.51 & \cellcolor{Green!8.2}9.60   & \cellcolor{Green!7.8}29.78  & \cellcolor{Green!0.0}2.38  & \cellcolor{Green!14.0}18.94 \\
 ChatLaw-33B          & \cellcolor{Green!0.0}3.67  & \cellcolor{Green!33.0}8.04  & \cellcolor{Green!30.0}32.08 & \cellcolor{Green!44.4}19.80 & \cellcolor{Green!0.0}37.16 & \cellcolor{Green!0.0}30.14 & \cellcolor{Green!0.0}35.47 & \cellcolor{Green!2.2}26.40  & \cellcolor{Green!0.4}22.14  & \cellcolor{Green!0.0}10.56 & \cellcolor{Green!11.0}22.55 \\
\hline
\end{tabular}
}
\begin{tabular}{ccccccc}
	\multicolumn{2}{c}{\%abstention\qquad} & \cellcolor{Green!0.0}\quad0\% \quad & \cellcolor{Green!25}\quad25\% \quad & \cellcolor{Green!50}\quad50\% \quad &
	\cellcolor{Green!75}\quad75\% \quad & \cellcolor{Green!100}\quad100\% \quad \
\end{tabular}
\caption{Zero-shot Results on Legal Knowledge Understanding Tasks}

\end{table*}
\endgroup

\begingroup
\begin{table*}
\centering
\small
\begin{tabular}{llllllllll}
\hline
 model\_name           & 3-1                         & 3-2                        & 3-3                         & 3-4                         & 3-5                         & 3-6                         & 3-7                         & 3-8                        & Application                 \\
\hline
\textbf{Multilingual LLMs} \\
 MPT-7B               & \cellcolor{Green!8.4}0.69   & \cellcolor{Green!0.0}8.20  & \cellcolor{Green!94.6}2.71  & \cellcolor{Green!26.4}58.41 & \cellcolor{Green!29.8}56.09 & \cellcolor{Green!0.6}0.00   & \cellcolor{Green!20.0}23.20 & \cellcolor{Green!0.0}9.67  & \cellcolor{Green!22.5}19.87 \\
 MPT-Instruct-7B      & \cellcolor{Green!18.4}0.66  & \cellcolor{Green!0.0}6.67  & \cellcolor{Green!37.2}12.19 & \cellcolor{Green!35.6}51.40 & \cellcolor{Green!52.6}37.39 & \cellcolor{Green!13.0}2.00  & \cellcolor{Green!12.4}18.60 & \cellcolor{Green!0.0}5.67  & \cellcolor{Green!21.2}16.82 \\
 LLaMA-7B             & \cellcolor{Green!12.4}0.95  & \cellcolor{Green!0.0}7.45  & \cellcolor{Green!55.0}6.98  & \cellcolor{Green!48.4}41.28 & \cellcolor{Green!30.8}54.22 & \cellcolor{Green!3.8}0.20   & \cellcolor{Green!21.2}14.40 & \cellcolor{Green!0.0}7.84  & \cellcolor{Green!21.4}16.67 \\
 LLaMA-13B            & \cellcolor{Green!34.6}0.10  & \cellcolor{Green!0.0}5.13  & \cellcolor{Green!51.6}4.67  & \cellcolor{Green!55.6}33.28 & \cellcolor{Green!51.8}37.19 & \cellcolor{Green!14.4}0.00  & \cellcolor{Green!12.2}21.60 & \cellcolor{Green!0.0}7.48  & \cellcolor{Green!27.5}13.68 \\
 LLaMA-30B            & \cellcolor{Green!20.2}0.80  & \cellcolor{Green!0.0}8.97  & \cellcolor{Green!65.6}4.77  & \cellcolor{Green!40.8}47.19 & \cellcolor{Green!37.4}48.62 & \cellcolor{Green!25.2}0.60  & \cellcolor{Green!7.8}28.20  & \cellcolor{Green!0.0}7.87  & \cellcolor{Green!24.6}18.38 \\
 LLaMA-65B            & \cellcolor{Green!17.2}0.33  & \cellcolor{Green!0.0}10.35 & \cellcolor{Green!62.4}7.55  & \cellcolor{Green!38.4}46.09 & \cellcolor{Green!51.6}37.74 & \cellcolor{Green!16.4}0.00  & \cellcolor{Green!7.4}27.20  & \cellcolor{Green!0.0}1.18  & \cellcolor{Green!24.2}16.31 \\
 LLaMA-2-7B           & \cellcolor{Green!30.0}0.99  & \cellcolor{Green!0.0}7.95  & \cellcolor{Green!37.2}7.59  & \cellcolor{Green!31.2}52.78 & \cellcolor{Green!43.4}43.05 & \cellcolor{Green!2.0}0.00   & \cellcolor{Green!6.0}17.60  & \cellcolor{Green!0.0}9.38  & \cellcolor{Green!18.7}17.42 \\
 LLaMA-2-13B          & \cellcolor{Green!29.8}0.90  & \cellcolor{Green!0.0}12.00 & \cellcolor{Green!71.0}5.43  & \cellcolor{Green!34.4}52.62 & \cellcolor{Green!57.2}32.74 & \cellcolor{Green!4.4}0.60   & \cellcolor{Green!33.4}11.20 & \cellcolor{Green!0.0}1.63  & \cellcolor{Green!28.8}14.64 \\
 LLaMA-2-70B          & \cellcolor{Green!64.2}3.70  & \cellcolor{Green!0.0}11.45 & \cellcolor{Green!78.4}6.36  & \cellcolor{Green!41.8}42.02 & \cellcolor{Green!44.8}42.82 & \cellcolor{Green!0.4}0.00   & \cellcolor{Green!11.8}29.20 & \cellcolor{Green!0.0}2.90  & \cellcolor{Green!30.2}17.31 \\
 LLaMA-2-Chat-7B      & \cellcolor{Green!7.8}1.87   & \cellcolor{Green!0.0}0.64  & \cellcolor{Green!53.2}11.57 & \cellcolor{Green!86.8}10.13 & \cellcolor{Green!90.2}7.21  & \cellcolor{Green!1.4}5.60   & \cellcolor{Green!14.2}36.20 & \cellcolor{Green!0.0}0.36  & \cellcolor{Green!31.7}9.20  \\
 LLaMA-2-Chat-13B     & \cellcolor{Green!8.0}1.40   & \cellcolor{Green!0.0}0.49  & \cellcolor{Green!40.8}15.29 & \cellcolor{Green!37.6}50.57 & \cellcolor{Green!34.8}52.64 & \cellcolor{Green!1.4}4.60   & \cellcolor{Green!1.8}28.60  & \cellcolor{Green!0.0}0.40  & \cellcolor{Green!15.6}19.25 \\
 LLaMA-2-Chat-70B     & \cellcolor{Green!8.4}6.82   & \cellcolor{Green!0.0}0.89  & \cellcolor{Green!52.2}14.91 & \cellcolor{Green!77.0}17.54 & \cellcolor{Green!82.0}12.81 & \cellcolor{Green!0.0}9.60   & \cellcolor{Green!24.6}42.40 & \cellcolor{Green!0.0}4.61  & \cellcolor{Green!30.5}13.70 \\
 Alpaca-v1.0-7B       & \cellcolor{Green!25.8}0.69  & \cellcolor{Green!0.0}13.11 & \cellcolor{Green!80.0}6.09  & \cellcolor{Green!25.4}59.61 & \cellcolor{Green!35.4}50.93 & \cellcolor{Green!50.8}2.00  & \cellcolor{Green!29.8}18.60 & \cellcolor{Green!0.0}1.56  & \cellcolor{Green!30.9}19.07 \\
 Vicuna-v1.3-7B       & \cellcolor{Green!72.2}0.16  & \cellcolor{Green!0.0}19.21 & \cellcolor{Green!92.4}2.20  & \cellcolor{Green!53.2}33.24 & \cellcolor{Green!68.0}23.95 & \cellcolor{Green!0.2}10.60  & \cellcolor{Green!16.6}11.00 & \cellcolor{Green!0.0}13.13 & \cellcolor{Green!37.8}14.19 \\
 Vicuna-v1.3-13B      & \cellcolor{Green!75.6}0.15  & \cellcolor{Green!0.0}19.28 & \cellcolor{Green!91.0}3.57  & \cellcolor{Green!64.0}27.01 & \cellcolor{Green!78.2}16.69 & \cellcolor{Green!1.2}18.80  & \cellcolor{Green!34.6}17.80 & \cellcolor{Green!0.0}14.34 & \cellcolor{Green!43.1}14.70 \\
 Vicuna-v1.3-33B      & \cellcolor{Green!72.6}0.44  & \cellcolor{Green!0.0}23.73 & \cellcolor{Green!81.0}4.56  & \cellcolor{Green!84.0}10.75 & \cellcolor{Green!85.2}10.95 & \cellcolor{Green!4.6}23.80  & \cellcolor{Green!8.6}9.00   & \cellcolor{Green!0.0}11.47 & \cellcolor{Green!42.0}11.84 \\
 WizardLM-7B          & \cellcolor{Green!4.8}0.60   & \cellcolor{Green!0.0}14.31 & \cellcolor{Green!47.0}8.16  & \cellcolor{Green!27.4}58.08 & \cellcolor{Green!37.6}48.80 & \cellcolor{Green!1.8}11.00  & \cellcolor{Green!23.8}18.40 & \cellcolor{Green!0.0}5.86  & \cellcolor{Green!17.8}20.65 \\
 StableBeluga2        & \cellcolor{Green!1.0}16.41  & \cellcolor{Green!0.0}24.52 & \cellcolor{Green!43.6}22.82 & \cellcolor{Green!4.2}76.06  & \cellcolor{Green!14.8}65.35 & \cellcolor{Green!0.0}34.40  & \cellcolor{Green!0.2}56.60  & \cellcolor{Green!0.0}13.39 & \cellcolor{Green!8.0}38.69  \\
 ChatGPT              & \cellcolor{Green!1.2}29.50  & \cellcolor{Green!0.0}31.30 & \cellcolor{Green!29.0}35.52 & \cellcolor{Green!3.8}78.75  & \cellcolor{Green!4.0}76.84  & \cellcolor{Green!0.6}27.40  & \cellcolor{Green!0.4}61.20  & \cellcolor{Green!0.0}17.45 & \cellcolor{Green!4.9}44.74  \\
 GPT-4                & \cellcolor{Green!0.4}52.47  & \cellcolor{Green!0.0}27.54 & \cellcolor{Green!25.0}41.99 & \cellcolor{Green!0.4}82.62  & \cellcolor{Green!0.4}81.91  & \cellcolor{Green!0.0}48.60  & \cellcolor{Green!0.4}77.60  & \cellcolor{Green!0.0}19.65 & \cellcolor{Green!3.3}54.05  \\
\midrule
\textbf{Chinese Oriented LLMs} \\
 MOSS-Moon            & \cellcolor{Green!6.0}23.73  & \cellcolor{Green!0.0}7.68  & \cellcolor{Green!54.6}13.11 & \cellcolor{Green!14.2}69.31 & \cellcolor{Green!30.4}53.45 & \cellcolor{Green!2.0}19.60  & \cellcolor{Green!50.0}9.80  & \cellcolor{Green!0.0}15.57 & \cellcolor{Green!19.7}26.53 \\
 MOSS-Moon-SFT        & \cellcolor{Green!1.0}5.23   & \cellcolor{Green!0.0}5.54  & \cellcolor{Green!15.0}20.80 & \cellcolor{Green!11.4}70.27 & \cellcolor{Green!5.2}76.55  & \cellcolor{Green!5.4}6.80   & \cellcolor{Green!6.2}26.80  & \cellcolor{Green!0.0}10.41 & \cellcolor{Green!5.5}27.80  \\
 TigerBot-Base        & \cellcolor{Green!37.0}16.61 & \cellcolor{Green!0.0}27.98 & \cellcolor{Green!39.6}24.81 & \cellcolor{Green!11.0}73.24 & \cellcolor{Green!18.4}65.64 & \cellcolor{Green!93.2}1.00  & \cellcolor{Green!51.8}12.00 & \cellcolor{Green!0.0}14.58 & \cellcolor{Green!31.4}29.48 \\
 TigerBot-SFT         & \cellcolor{Green!59.4}3.54  & \cellcolor{Green!0.0}25.58 & \cellcolor{Green!34.2}21.97 & \cellcolor{Green!4.8}76.57  & \cellcolor{Green!12.4}71.63 & \cellcolor{Green!2.4}12.60  & \cellcolor{Green!0.6}30.40  & \cellcolor{Green!0.0}16.19 & \cellcolor{Green!14.2}32.31 \\
 GoGPT                & \cellcolor{Green!95.8}0.00  & \cellcolor{Green!0.0}11.61 & \cellcolor{Green!73.2}5.91  & \cellcolor{Green!84.6}10.71 & \cellcolor{Green!96.6}2.10  & \cellcolor{Green!41.6}14.20 & \cellcolor{Green!19.8}10.60 & \cellcolor{Green!0.0}9.47  & \cellcolor{Green!51.4}8.07  \\
 ChatGLM2-6B          & \cellcolor{Green!6.0}23.12  & \cellcolor{Green!0.0}24.96 & \cellcolor{Green!42.6}27.12 & \cellcolor{Green!23.8}59.79 & \cellcolor{Green!9.0}72.84  & \cellcolor{Green!1.4}10.80  & \cellcolor{Green!16.0}16.80 & \cellcolor{Green!0.0}17.56 & \cellcolor{Green!12.3}31.62 \\
 Ziya-LLaMA-13B       & \cellcolor{Green!66.2}10.20 & \cellcolor{Green!0.0}21.23 & \cellcolor{Green!48.2}25.93 & \cellcolor{Green!98.0}0.97  & \cellcolor{Green!98.6}0.57  & \cellcolor{Green!2.0}30.60  & \cellcolor{Green!41.6}12.80 & \cellcolor{Green!0.0}13.36 & \cellcolor{Green!44.3}14.46 \\
 Baichuan-7B          & \cellcolor{Green!5.8}10.05  & \cellcolor{Green!0.0}8.36  & \cellcolor{Green!44.6}22.56 & \cellcolor{Green!5.4}75.71  & \cellcolor{Green!5.0}76.42  & \cellcolor{Green!5.8}5.40   & \cellcolor{Green!16.4}17.80 & \cellcolor{Green!0.0}2.40  & \cellcolor{Green!10.4}27.34 \\
 Baichuan-13B         & \cellcolor{Green!18.6}27.48 & \cellcolor{Green!0.0}9.51  & \cellcolor{Green!44.2}29.61 & \cellcolor{Green!10.4}72.53 & \cellcolor{Green!16.0}67.64 & \cellcolor{Green!17.0}14.40 & \cellcolor{Green!19.4}19.20 & \cellcolor{Green!0.0}4.60  & \cellcolor{Green!15.7}30.62 \\
 Baichuan-13B-Chat    & \cellcolor{Green!20.8}22.79 & \cellcolor{Green!0.0}27.23 & \cellcolor{Green!58.4}24.21 & \cellcolor{Green!54.2}35.61 & \cellcolor{Green!41.4}46.11 & \cellcolor{Green!19.8}27.00 & \cellcolor{Green!8.0}43.00  & \cellcolor{Green!0.0}20.29 & \cellcolor{Green!25.3}30.78 \\
 XVERSE-13B           & \cellcolor{Green!6.8}1.31   & \cellcolor{Green!0.0}8.68  & \cellcolor{Green!4.0}15.76  & \cellcolor{Green!7.0}75.68  & \cellcolor{Green!17.6}66.66 & \cellcolor{Green!0.0}0.00   & \cellcolor{Green!11.2}37.60 & \cellcolor{Green!0.0}8.85  & \cellcolor{Green!5.8}26.82  \\
 InternLM-7B          & \cellcolor{Green!11.6}23.05 & \cellcolor{Green!0.0}6.72  & \cellcolor{Green!20.4}32.43 & \cellcolor{Green!3.6}73.38  & \cellcolor{Green!27.0}56.71 & \cellcolor{Green!0.8}10.40  & \cellcolor{Green!7.6}22.80  & \cellcolor{Green!0.0}11.30 & \cellcolor{Green!8.9}29.60  \\
 InternLM-Chat-7B     & \cellcolor{Green!24.0}24.94 & \cellcolor{Green!0.0}20.80 & \cellcolor{Green!21.6}42.35 & \cellcolor{Green!18.8}67.33 & \cellcolor{Green!20.0}64.97 & \cellcolor{Green!0.0}37.80  & \cellcolor{Green!0.6}44.80  & \cellcolor{Green!0.0}15.36 & \cellcolor{Green!10.6}39.79 \\
 InternLM-Chat-7B-8K  & \cellcolor{Green!7.0}34.86  & \cellcolor{Green!0.0}19.11 & \cellcolor{Green!23.0}41.05 & \cellcolor{Green!21.8}63.21 & \cellcolor{Green!16.6}67.20 & \cellcolor{Green!0.0}34.20  & \cellcolor{Green!0.6}43.80  & \cellcolor{Green!0.0}13.37 & \cellcolor{Green!8.6}39.60  \\
 Qwen-7B              & \cellcolor{Green!59.8}19.88 & \cellcolor{Green!0.0}20.80 & \cellcolor{Green!63.8}20.16 & \cellcolor{Green!7.6}74.59  & \cellcolor{Green!27.4}56.77 & \cellcolor{Green!89.8}0.40  & \cellcolor{Green!3.4}32.80  & \cellcolor{Green!0.0}10.07 & \cellcolor{Green!31.5}29.43 \\
 Qwen-7B-Chat         & \cellcolor{Green!8.4}44.62  & \cellcolor{Green!0.0}33.50 & \cellcolor{Green!24.6}40.67 & \cellcolor{Green!7.6}76.74  & \cellcolor{Green!7.4}77.19  & \cellcolor{Green!0.4}26.80  & \cellcolor{Green!0.4}42.00  & \cellcolor{Green!0.0}19.32 & \cellcolor{Green!6.1}45.10  \\
 Yulan-Chat-2-13B     & \cellcolor{Green!1.4}7.64   & \cellcolor{Green!0.0}30.67 & \cellcolor{Green!24.2}38.30 & \cellcolor{Green!3.6}76.66  & \cellcolor{Green!4.4}75.78  & \cellcolor{Green!0.2}23.20  & \cellcolor{Green!1.0}46.00  & \cellcolor{Green!0.0}17.54 & \cellcolor{Green!4.3}39.47  \\
 BELLE-LLaMA-2        & \cellcolor{Green!6.0}3.27   & \cellcolor{Green!0.0}31.79 & \cellcolor{Green!39.6}19.65 & \cellcolor{Green!13.0}68.21 & \cellcolor{Green!23.0}61.17 & \cellcolor{Green!2.2}22.20  & \cellcolor{Green!7.6}42.80  & \cellcolor{Green!0.0}17.04 & \cellcolor{Green!11.4}33.26 \\
 Chinese-LLaMA-2-7B   & \cellcolor{Green!88.0}0.14  & \cellcolor{Green!0.0}4.43  & \cellcolor{Green!58.8}18.13 & \cellcolor{Green!48.4}38.80 & \cellcolor{Green!56.4}33.49 & \cellcolor{Green!70.4}0.40  & \cellcolor{Green!26.0}16.80 & \cellcolor{Green!0.0}7.63  & \cellcolor{Green!43.5}14.98 \\
 Chinese-Alpaca-2-7B  & \cellcolor{Green!96.2}0.05  & \cellcolor{Green!0.0}35.51 & \cellcolor{Green!38.0}26.35 & \cellcolor{Green!98.6}0.55  & \cellcolor{Green!84.0}12.10 & \cellcolor{Green!11.6}7.00  & \cellcolor{Green!46.0}8.60  & \cellcolor{Green!0.0}14.07 & \cellcolor{Green!46.8}13.03 \\
 LLaMA-2-Chinese-7B   & \cellcolor{Green!79.0}0.00  & \cellcolor{Green!0.0}5.04  & \cellcolor{Green!96.8}1.29  & \cellcolor{Green!89.4}7.09  & \cellcolor{Green!68.0}23.82 & \cellcolor{Green!98.0}0.00  & \cellcolor{Green!81.8}2.60  & \cellcolor{Green!0.0}1.22  & \cellcolor{Green!64.1}5.13  \\
 LLaMA-2-Chinese-13B  & \cellcolor{Green!77.0}0.00  & \cellcolor{Green!0.0}0.15  & \cellcolor{Green!95.8}1.27  & \cellcolor{Green!61.2}29.39 & \cellcolor{Green!63.0}28.23 & \cellcolor{Green!99.8}0.00  & \cellcolor{Green!69.8}11.00 & \cellcolor{Green!0.0}2.37  & \cellcolor{Green!58.3}9.05  \\
\midrule
\textbf{Law Specific LLMs} \\
 LexiLaw              & \cellcolor{Green!9.8}13.15  & \cellcolor{Green!0.0}35.78 & \cellcolor{Green!24.0}39.99 & \cellcolor{Green!4.2}78.08  & \cellcolor{Green!8.2}74.92  & \cellcolor{Green!10.0}20.80 & \cellcolor{Green!6.8}35.80  & \cellcolor{Green!0.0}15.82 & \cellcolor{Green!7.9}39.29  \\
 Fuzi-Mingcha         & \cellcolor{Green!0.0}25.19  & \cellcolor{Green!0.0}22.18 & \cellcolor{Green!5.4}55.93  & \cellcolor{Green!2.4}77.23  & \cellcolor{Green!2.0}75.52  & \cellcolor{Green!29.2}7.00  & \cellcolor{Green!0.8}47.20  & \cellcolor{Green!0.0}16.64 & \cellcolor{Green!5.0}40.86  \\
 HanFei               & \cellcolor{Green!2.6}2.64   & \cellcolor{Green!0.0}33.60 & \cellcolor{Green!21.2}30.96 & \cellcolor{Green!8.6}73.11  & \cellcolor{Green!11.2}69.63 & \cellcolor{Green!3.8}23.60  & \cellcolor{Green!1.2}39.40  & \cellcolor{Green!0.0}16.06 & \cellcolor{Green!6.1}36.13  \\
 LaWGPT-7B-beta1.0    & \cellcolor{Green!83.4}0.00  & \cellcolor{Green!0.0}8.16  & \cellcolor{Green!92.0}3.23  & \cellcolor{Green!98.6}0.34  & \cellcolor{Green!96.8}1.72  & \cellcolor{Green!100.0}0.00 & \cellcolor{Green!99.0}0.00  & \cellcolor{Green!0.0}0.80  & \cellcolor{Green!71.2}1.78  \\
 LaWGPT-7B-beta1.1    & \cellcolor{Green!16.0}0.15  & \cellcolor{Green!0.0}11.01 & \cellcolor{Green!54.8}15.68 & \cellcolor{Green!42.4}42.37 & \cellcolor{Green!45.4}40.80 & \cellcolor{Green!26.6}6.20  & \cellcolor{Green!31.8}15.40 & \cellcolor{Green!0.0}7.62  & \cellcolor{Green!27.1}17.40 \\
 Wisdom-Interrogatory & \cellcolor{Green!6.2}32.84  & \cellcolor{Green!0.0}32.01 & \cellcolor{Green!35.2}35.09 & \cellcolor{Green!1.0}80.36  & \cellcolor{Green!0.8}81.10  & \cellcolor{Green!0.2}15.40  & \cellcolor{Green!0.0}17.40  & \cellcolor{Green!0.0}20.17 & \cellcolor{Green!5.4}39.29  \\
 ChatLaw-13B          & \cellcolor{Green!5.8}33.28  & \cellcolor{Green!0.0}31.55 & \cellcolor{Green!35.4}27.90 & \cellcolor{Green!3.8}76.18  & \cellcolor{Green!5.0}73.57  & \cellcolor{Green!0.6}28.80  & \cellcolor{Green!6.0}41.40  & \cellcolor{Green!0.0}17.17 & \cellcolor{Green!7.1}41.23  \\
 Lawyer-LLaMA         & \cellcolor{Green!3.0}0.60   & \cellcolor{Green!0.0}25.94 & \cellcolor{Green!30.4}31.30 & \cellcolor{Green!6.6}74.19  & \cellcolor{Green!4.0}75.52  & \cellcolor{Green!28.6}17.80 & \cellcolor{Green!2.8}39.20  & \cellcolor{Green!0.0}16.94 & \cellcolor{Green!9.4}35.19  \\
 ChatLaw-33B          & \cellcolor{Green!5.0}5.35   & \cellcolor{Green!0.0}26.02 & \cellcolor{Green!46.8}12.73 & \cellcolor{Green!16.8}67.00 & \cellcolor{Green!34.4}53.63 & \cellcolor{Green!0.2}34.20  & \cellcolor{Green!9.6}41.60  & \cellcolor{Green!0.0}16.55 & \cellcolor{Green!14.1}32.14 \\
\hline
\end{tabular}

\begin{tabular}{ccccccc}
	\multicolumn{2}{c}{\%abstention\qquad} & \cellcolor{Green!0.0}\quad0\% \quad & \cellcolor{Green!25}\quad25\% \quad & \cellcolor{Green!50}\quad50\% \quad &
	\cellcolor{Green!75}\quad75\% \quad & \cellcolor{Green!100}\quad100\% \quad \
\end{tabular}
\caption{Zero-shot Results on Legal Knowledge Application Tasks}

\end{table*}
\endgroup

\begingroup
\begin{table*}
\centering
\small
\begin{tabular}{lllll}
\hline
 model\_name           & Memorization                & Understanding               & Application                 & Overall                     \\
\hline
\textbf{Multilingual LLMs} \\
 MPT-7B               & \cellcolor{Green!0.2}1.78   & \cellcolor{Green!24.6}6.16  & \cellcolor{Green!22.5}19.87 & \cellcolor{Green!21.3}11.21 \\
 MPT-Instruct-7B      & \cellcolor{Green!2.4}1.65   & \cellcolor{Green!25.2}6.87  & \cellcolor{Green!21.2}16.82 & \cellcolor{Green!21.3}10.33 \\
 LLaMA-7B             & \cellcolor{Green!5.0}1.51   & \cellcolor{Green!26.6}5.80  & \cellcolor{Green!21.4}16.67 & \cellcolor{Green!22.4}9.72  \\
 LLaMA-13B            & \cellcolor{Green!8.5}1.72   & \cellcolor{Green!24.1}6.20  & \cellcolor{Green!27.5}13.68 & \cellcolor{Green!23.9}8.74  \\
 LLaMA-30B            & \cellcolor{Green!3.9}1.80   & \cellcolor{Green!24.8}3.46  & \cellcolor{Green!24.6}18.38 & \cellcolor{Green!22.7}9.26  \\
 LLaMA-65B            & \cellcolor{Green!4.1}1.20   & \cellcolor{Green!28.8}3.60  & \cellcolor{Green!24.2}16.31 & \cellcolor{Green!24.5}8.44  \\
 LLaMA-2-7B           & \cellcolor{Green!1.7}1.52   & \cellcolor{Green!23.6}9.15  & \cellcolor{Green!18.7}17.42 & \cellcolor{Green!19.5}11.69 \\
 LLaMA-2-13B          & \cellcolor{Green!0.9}1.85   & \cellcolor{Green!26.0}7.48  & \cellcolor{Green!28.8}14.64 & \cellcolor{Green!24.6}9.78  \\
 LLaMA-2-70B          & \cellcolor{Green!0.1}1.77   & \cellcolor{Green!13.8}12.91 & \cellcolor{Green!30.2}17.31 & \cellcolor{Green!19.0}13.56 \\
 LLaMA-2-Chat-7B      & \cellcolor{Green!0.9}3.60   & \cellcolor{Green!12.2}6.24  & \cellcolor{Green!31.7}9.20  & \cellcolor{Green!18.9}7.16  \\
 LLaMA-2-Chat-13B     & \cellcolor{Green!0.7}2.12   & \cellcolor{Green!5.6}13.46  & \cellcolor{Green!15.6}19.25 & \cellcolor{Green!9.1}14.64  \\
 LLaMA-2-Chat-70B     & \cellcolor{Green!0.0}5.43   & \cellcolor{Green!6.7}17.37  & \cellcolor{Green!30.5}13.70 & \cellcolor{Green!15.5}14.71 \\
 Alpaca-v1.0-7B       & \cellcolor{Green!24.5}0.80  & \cellcolor{Green!35.9}4.63  & \cellcolor{Green!30.9}19.07 & \cellcolor{Green!32.8}10.02 \\
 Vicuna-v1.3-7B       & \cellcolor{Green!0.8}7.60   & \cellcolor{Green!17.4}10.16 & \cellcolor{Green!37.8}14.19 & \cellcolor{Green!23.9}11.52 \\
 Vicuna-v1.3-13B      & \cellcolor{Green!0.1}12.91  & \cellcolor{Green!10.6}10.44 & \cellcolor{Green!43.1}14.70 & \cellcolor{Green!22.5}12.39 \\
 Vicuna-v1.3-33B      & \cellcolor{Green!1.5}17.43  & \cellcolor{Green!20.9}10.19 & \cellcolor{Green!42.0}11.84 & \cellcolor{Green!27.4}11.57 \\
 WizardLM-7B          & \cellcolor{Green!0.4}7.14   & \cellcolor{Green!14.4}9.64  & \cellcolor{Green!17.8}20.65 & \cellcolor{Green!14.4}13.79 \\
 StableBeluga2        & \cellcolor{Green!0.0}24.59  & \cellcolor{Green!3.7}42.58  & \cellcolor{Green!8.0}38.69  & \cellcolor{Green!5.0}39.23  \\
 ChatGPT              & \cellcolor{Green!0.0}25.93  & \cellcolor{Green!1.2}43.31  & \cellcolor{Green!4.9}44.74  & \cellcolor{Green!2.6}42.15  \\
 GPT-4                & \cellcolor{Green!0.1}35.29  & \cellcolor{Green!0.4}54.41  & \cellcolor{Green!3.3}54.05  & \cellcolor{Green!1.5}52.35  \\
\midrule
\textbf{Chinese Oriented LLMs} \\
 MOSS-Moon            & \cellcolor{Green!3.7}8.32   & \cellcolor{Green!24.0}6.80  & \cellcolor{Green!19.7}26.53 & \cellcolor{Green!20.2}14.84 \\
 MOSS-Moon-SFT        & \cellcolor{Green!4.4}5.25   & \cellcolor{Green!5.7}7.15   & \cellcolor{Green!5.5}27.80  & \cellcolor{Green!5.5}15.22  \\
 TigerBot-Base        & \cellcolor{Green!40.8}9.71  & \cellcolor{Green!13.3}17.88 & \cellcolor{Green!31.4}29.48 & \cellcolor{Green!23.3}21.71 \\
 TigerBot-SFT         & \cellcolor{Green!2.7}12.39  & \cellcolor{Green!2.8}23.45  & \cellcolor{Green!14.2}32.31 & \cellcolor{Green!7.4}25.89  \\
 GoGPT                & \cellcolor{Green!15.5}12.05 & \cellcolor{Green!37.0}8.97  & \cellcolor{Green!51.4}8.07  & \cellcolor{Green!40.6}8.92  \\
 ChatGLM2-6B          & \cellcolor{Green!0.9}12.98  & \cellcolor{Green!13.0}14.41 & \cellcolor{Green!12.3}31.62 & \cellcolor{Green!11.6}21.15 \\
 Ziya-LLaMA-13B       & \cellcolor{Green!1.0}22.52  & \cellcolor{Green!16.4}19.90 & \cellcolor{Green!44.3}14.46 & \cellcolor{Green!26.0}17.99 \\
 Baichuan-7B          & \cellcolor{Green!17.9}0.92  & \cellcolor{Green!24.5}8.13  & \cellcolor{Green!10.4}27.34 & \cellcolor{Green!18.2}15.09 \\
 Baichuan-13B         & \cellcolor{Green!22.3}6.35  & \cellcolor{Green!17.6}12.35 & \cellcolor{Green!15.7}30.62 & \cellcolor{Green!17.3}19.06 \\
 Baichuan-13B-Chat    & \cellcolor{Green!7.6}23.56  & \cellcolor{Green!32.8}13.49 & \cellcolor{Green!25.3}30.78 & \cellcolor{Green!27.3}21.41 \\
 XVERSE-13B           & \cellcolor{Green!0.6}0.85   & \cellcolor{Green!6.9}5.47   & \cellcolor{Green!5.8}26.82  & \cellcolor{Green!5.8}13.55  \\
 InternLM-7B          & \cellcolor{Green!0.1}3.08   & \cellcolor{Green!20.5}8.24  & \cellcolor{Green!8.9}29.60  & \cellcolor{Green!13.8}16.27 \\
 InternLM-Chat-7B     & \cellcolor{Green!0.0}27.20  & \cellcolor{Green!5.1}31.96  & \cellcolor{Green!10.6}39.79 & \cellcolor{Green!6.8}34.62  \\
 InternLM-Chat-7B-8K  & \cellcolor{Green!0.0}27.93  & \cellcolor{Green!3.9}34.19  & \cellcolor{Green!8.6}39.60  & \cellcolor{Green!5.4}35.73  \\
 Qwen-7B              & \cellcolor{Green!46.7}0.00  & \cellcolor{Green!22.9}15.61 & \cellcolor{Green!31.5}29.43 & \cellcolor{Green!28.7}19.58 \\
 Qwen-7B-Chat         & \cellcolor{Green!0.1}26.27  & \cellcolor{Green!8.1}32.66  & \cellcolor{Green!6.1}45.10  & \cellcolor{Green!6.5}37.00  \\
 Yulan-Chat-2-13B     & \cellcolor{Green!0.0}18.72  & \cellcolor{Green!6.0}32.19  & \cellcolor{Green!4.3}39.47  & \cellcolor{Green!4.7}33.76  \\
 BELLE-LLaMA-2        & \cellcolor{Green!0.1}22.50  & \cellcolor{Green!3.4}29.71  & \cellcolor{Green!11.4}33.26 & \cellcolor{Green!6.3}30.41  \\
 Chinese-LLaMA-2-7B   & \cellcolor{Green!37.0}2.22  & \cellcolor{Green!41.7}5.16  & \cellcolor{Green!43.5}14.98 & \cellcolor{Green!41.9}8.79  \\
 Chinese-Alpaca-2-7B  & \cellcolor{Green!3.4}9.67   & \cellcolor{Green!20.7}15.19 & \cellcolor{Green!46.8}13.03 & \cellcolor{Green!29.4}13.77 \\
 LLaMA-2-Chinese-7B   & \cellcolor{Green!49.6}0.00  & \cellcolor{Green!46.3}1.81  & \cellcolor{Green!64.1}5.13  & \cellcolor{Green!53.8}2.96  \\
 LLaMA-2-Chinese-13B  & \cellcolor{Green!50.0}0.00  & \cellcolor{Green!46.8}1.38  & \cellcolor{Green!58.3}9.05  & \cellcolor{Green!51.7}4.31  \\
\midrule
\textbf{Law Specific LLMs} \\
 LexiLaw              & \cellcolor{Green!1.7}18.98  & \cellcolor{Green!9.1}22.32  & \cellcolor{Green!7.9}39.29  & \cellcolor{Green!7.9}28.78  \\
 Fuzi-Mingcha         & \cellcolor{Green!5.7}16.51  & \cellcolor{Green!10.3}30.10 & \cellcolor{Green!5.0}40.86  & \cellcolor{Green!7.7}33.05  \\
 HanFei               & \cellcolor{Green!0.6}18.42  & \cellcolor{Green!4.3}26.84  & \cellcolor{Green!6.1}36.13  & \cellcolor{Green!4.6}29.71  \\
 LaWGPT-7B-beta1.0    & \cellcolor{Green!50.0}4.67  & \cellcolor{Green!49.2}0.67  & \cellcolor{Green!71.2}1.78  & \cellcolor{Green!58.1}1.51  \\
 LaWGPT-7B-beta1.1    & \cellcolor{Green!6.8}4.37   & \cellcolor{Green!34.3}5.03  & \cellcolor{Green!27.1}17.40 & \cellcolor{Green!28.7}9.91  \\
 Wisdom-Interrogatory & \cellcolor{Green!0.6}29.23  & \cellcolor{Green!4.2}25.53  & \cellcolor{Green!5.4}39.29  & \cellcolor{Green!4.4}31.41  \\
 ChatLaw-13B          & \cellcolor{Green!0.2}21.63  & \cellcolor{Green!5.6}28.21  & \cellcolor{Green!7.1}41.23  & \cellcolor{Green!5.7}32.76  \\
 Lawyer-LLaMA         & \cellcolor{Green!4.1}17.77  & \cellcolor{Green!14.0}18.94 & \cellcolor{Green!9.4}35.19  & \cellcolor{Green!11.2}25.32 \\
 ChatLaw-33B          & \cellcolor{Green!0.0}20.17  & \cellcolor{Green!11.0}22.55 & \cellcolor{Green!14.1}32.14 & \cellcolor{Green!11.1}26.14 \\
\hline
\end{tabular}

\begin{tabular}{ccccccc}
	\multicolumn{2}{c}{\%abstention\qquad} & \cellcolor{Green!0.0}\quad0\% \quad & \cellcolor{Green!25}\quad25\% \quad & \cellcolor{Green!50}\quad50\% \quad &
	\cellcolor{Green!75}\quad75\% \quad & \cellcolor{Green!100}\quad100\% \quad \
\end{tabular}
\caption{Zero-shot Results on Overall}
\label{tab:zero-shotResults on Overall}

\end{table*}
\endgroup

\begingroup
\begin{table*}
\centering
\small
\begin{tabular}{llll}
\hline
 model\_name           & 1-1                        & 1-2                         & Memorization                \\
\hline
\textbf{Multilingual LLMs} \\
 MPT-7B               & \cellcolor{Green!0.0}3.26  & \cellcolor{Green!0.2}0.40   & \cellcolor{Green!0.1}1.83   \\
 MPT-Instruct-7B      & \cellcolor{Green!0.0}2.06  & \cellcolor{Green!16.4}2.00  & \cellcolor{Green!8.2}2.03   \\
 LLaMA-7B             & \cellcolor{Green!0.0}3.14  & \cellcolor{Green!0.2}1.20   & \cellcolor{Green!0.1}2.17   \\
 LLaMA-13B            & \cellcolor{Green!0.0}3.00  & \cellcolor{Green!0.0}2.60   & \cellcolor{Green!0.0}2.80   \\
 LLaMA-30B            & \cellcolor{Green!0.0}3.15  & \cellcolor{Green!0.4}1.00   & \cellcolor{Green!0.2}2.07   \\
 LLaMA-65B            & \cellcolor{Green!0.0}2.00  & \cellcolor{Green!0.0}1.20   & \cellcolor{Green!0.0}1.60   \\
 LLaMA-2-7B           & \cellcolor{Green!0.0}3.42  & \cellcolor{Green!0.2}1.80   & \cellcolor{Green!0.1}2.61   \\
 LLaMA-2-13B          & \cellcolor{Green!0.0}3.58  & \cellcolor{Green!0.0}1.20   & \cellcolor{Green!0.0}2.39   \\
 LLaMA-2-70B          & \cellcolor{Green!0.0}3.33  & \cellcolor{Green!0.0}0.40   & \cellcolor{Green!0.0}1.86   \\
 LLaMA-2-Chat-7B      & \cellcolor{Green!0.0}0.82  & \cellcolor{Green!8.8}7.40   & \cellcolor{Green!4.4}4.11   \\
 LLaMA-2-Chat-13B     & \cellcolor{Green!0.0}1.62  & \cellcolor{Green!1.6}3.80   & \cellcolor{Green!0.8}2.71   \\
 LLaMA-2-Chat-70B     & \cellcolor{Green!0.0}0.73  & \cellcolor{Green!0.0}11.00  & \cellcolor{Green!0.0}5.86   \\
 Alpaca-v1.0-7B       & \cellcolor{Green!0.0}9.18  & \cellcolor{Green!21.4}4.60  & \cellcolor{Green!10.7}6.89  \\
 Vicuna-v1.3-7B       & \cellcolor{Green!0.0}11.99 & \cellcolor{Green!0.0}1.60   & \cellcolor{Green!0.0}6.80   \\
 Vicuna-v1.3-13B      & \cellcolor{Green!0.0}6.01  & \cellcolor{Green!0.0}14.40  & \cellcolor{Green!0.0}10.21  \\
 Vicuna-v1.3-33B      & \cellcolor{Green!0.0}7.67  & \cellcolor{Green!1.2}22.80  & \cellcolor{Green!0.6}15.24  \\
 WizardLM-7B          & \cellcolor{Green!0.0}10.81 & \cellcolor{Green!4.0}8.60   & \cellcolor{Green!2.0}9.71   \\
 StableBeluga2        & \cellcolor{Green!0.0}15.03 & \cellcolor{Green!0.0}36.00  & \cellcolor{Green!0.0}25.51  \\
 ChatGPT              & \cellcolor{Green!0.0}16.15 & \cellcolor{Green!0.0}37.20  & \cellcolor{Green!0.0}26.67  \\
 GPT-4                & \cellcolor{Green!0.0}17.21 & \cellcolor{Green!0.2}54.80  & \cellcolor{Green!0.1}36.00  \\
\midrule
\textbf{Chinese Oriented LLMs} \\
 MOSS-Moon            & \cellcolor{Green!0.0}3.09  & \cellcolor{Green!0.0}0.40   & \cellcolor{Green!0.0}1.75   \\
 MOSS-Moon-SFT        & \cellcolor{Green!0.0}2.07  & \cellcolor{Green!17.8}1.40  & \cellcolor{Green!8.9}1.74   \\
 TigerBot-Base        & \cellcolor{Green!0.0}14.09 & \cellcolor{Green!69.6}0.20  & \cellcolor{Green!34.8}7.14  \\
 TigerBot-SFT         & \cellcolor{Green!0.0}11.09 & \cellcolor{Green!2.8}6.20   & \cellcolor{Green!1.4}8.64   \\
 GoGPT                & \cellcolor{Green!0.0}8.21  & \cellcolor{Green!25.2}20.40 & \cellcolor{Green!12.6}14.30 \\
 ChatGLM2-6B          & \cellcolor{Green!0.0}14.67 & \cellcolor{Green!8.2}24.80  & \cellcolor{Green!4.1}19.74  \\
 Ziya-LLaMA-13B       & \cellcolor{Green!0.0}0.00  & \cellcolor{Green!1.0}28.00  & \cellcolor{Green!0.5}14.00  \\
 Baichuan-7B          & \cellcolor{Green!0.0}3.88  & \cellcolor{Green!0.0}0.40   & \cellcolor{Green!0.0}2.14   \\
 Baichuan-13B         & \cellcolor{Green!0.0}4.05  & \cellcolor{Green!0.0}1.00   & \cellcolor{Green!0.0}2.53   \\
 Baichuan-13B-Chat    & \cellcolor{Green!0.0}9.78  & \cellcolor{Green!4.8}40.40  & \cellcolor{Green!2.4}25.09  \\
 XVERSE-13B           & \cellcolor{Green!0.0}2.43  & \cellcolor{Green!2.2}0.00   & \cellcolor{Green!1.1}1.22   \\
 InternLM-7B          & \cellcolor{Green!0.0}2.79  & \cellcolor{Green!0.2}1.00   & \cellcolor{Green!0.1}1.89   \\
 InternLM-Chat-7B     & \cellcolor{Green!0.0}13.85 & \cellcolor{Green!0.2}39.60  & \cellcolor{Green!0.1}26.73  \\
 InternLM-Chat-7B-8K  & \cellcolor{Green!0.0}15.16 & \cellcolor{Green!0.2}40.60  & \cellcolor{Green!0.1}27.88  \\
 Qwen-7B              & \cellcolor{Green!0.0}0.00  & \cellcolor{Green!81.0}0.00  & \cellcolor{Green!40.5}0.00  \\
 Qwen-7B-Chat         & \cellcolor{Green!0.0}17.73 & \cellcolor{Green!0.0}28.60  & \cellcolor{Green!0.0}23.16  \\
 Yulan-Chat-2-13B     & \cellcolor{Green!0.0}15.41 & \cellcolor{Green!0.0}21.40  & \cellcolor{Green!0.0}18.40  \\
 BELLE-LLaMA-2        & \cellcolor{Green!0.0}16.58 & \cellcolor{Green!0.2}21.40  & \cellcolor{Green!0.1}18.99  \\
 Chinese-LLaMA-2-7B   & \cellcolor{Green!0.0}2.77  & \cellcolor{Green!1.4}1.20   & \cellcolor{Green!0.7}1.98   \\
 Chinese-Alpaca-2-7B  & \cellcolor{Green!0.0}11.84 & \cellcolor{Green!1.2}14.00  & \cellcolor{Green!0.6}12.92  \\
 LLaMA-2-Chinese-7B   & \cellcolor{Green!0.0}0.00  & \cellcolor{Green!97.8}0.40  & \cellcolor{Green!48.9}0.20  \\
 LLaMA-2-Chinese-13B  & \cellcolor{Green!0.0}0.00  & \cellcolor{Green!89.8}3.20  & \cellcolor{Green!44.9}1.60  \\
\midrule
\textbf{Law Specific LLMs} \\
 LexiLaw              & \cellcolor{Green!0.0}15.47 & \cellcolor{Green!4.2}14.40  & \cellcolor{Green!2.1}14.94  \\
 Fuzi-Mingcha         & \cellcolor{Green!0.0}20.21 & \cellcolor{Green!5.8}12.80  & \cellcolor{Green!2.9}16.50  \\
 HanFei               & \cellcolor{Green!0.0}16.23 & \cellcolor{Green!0.8}8.60   & \cellcolor{Green!0.4}12.41  \\
 LaWGPT-7B-beta1.0    & \cellcolor{Green!0.0}1.24  & \cellcolor{Green!99.4}0.00  & \cellcolor{Green!49.7}0.62  \\
 LaWGPT-7B-beta1.1    & \cellcolor{Green!0.0}3.03  & \cellcolor{Green!4.2}6.60   & \cellcolor{Green!2.1}4.82   \\
 Wisdom-Interrogatory & \cellcolor{Green!0.0}37.41 & \cellcolor{Green!0.0}15.20  & \cellcolor{Green!0.0}26.30  \\
 ChatLaw-13B          & \cellcolor{Green!0.0}15.98 & \cellcolor{Green!0.0}29.40  & \cellcolor{Green!0.0}22.69  \\
 Lawyer-LLaMA         & \cellcolor{Green!0.0}13.04 & \cellcolor{Green!23.8}10.60 & \cellcolor{Green!11.9}11.82 \\
 ChatLaw-33B          & \cellcolor{Green!0.0}14.36 & \cellcolor{Green!0.0}27.80  & \cellcolor{Green!0.0}21.08  \\
\hline
\end{tabular}

\begin{tabular}{ccccccc}
	\multicolumn{2}{c}{\%abstention\qquad} & \cellcolor{Green!0.0}\quad0\% \quad & \cellcolor{Green!25}\quad25\% \quad & \cellcolor{Green!50}\quad50\% \quad &
	\cellcolor{Green!75}\quad75\% \quad & \cellcolor{Green!100}\quad100\% \quad \
\end{tabular}
\caption{One-shot Results on Legal Knowledge Memorization Tasks}
\label{tab:one-shotResults on legal knowledge memorisation tasks}
\end{table*}
\endgroup

\begingroup
\begin{table*}
\centering
\small
\resizebox{\columnwidth}{!}{
\begin{tabular}{llllllllllll}
\hline
 model\_name           & 2-1                        & 2-2                         & 2-3                         & 2-4                         & 2-5                        & 2-6                        & 2-7                        & 2-8                         & 2-9                         & 2-10                       & Understanding               \\
\hline
\textbf{Multilingual LLMs} \\
 MPT-7B               & \cellcolor{Green!0.0}2.35  & \cellcolor{Green!42.6}4.00  & \cellcolor{Green!18.0}22.37 & \cellcolor{Green!35.4}6.80  & \cellcolor{Green!0.0}4.39  & \cellcolor{Green!0.0}34.91 & \cellcolor{Green!0.0}16.09 & \cellcolor{Green!7.0}1.40   & \cellcolor{Green!60.4}12.67 & \cellcolor{Green!0.0}1.53  & \cellcolor{Green!16.3}10.65 \\
 MPT-Instruct-7B      & \cellcolor{Green!0.0}2.36  & \cellcolor{Green!49.0}4.20  & \cellcolor{Green!15.2}16.38 & \cellcolor{Green!50.0}8.00  & \cellcolor{Green!0.0}10.63 & \cellcolor{Green!0.0}36.96 & \cellcolor{Green!0.0}13.11 & \cellcolor{Green!28.6}9.00  & \cellcolor{Green!34.4}21.69 & \cellcolor{Green!0.0}3.21  & \cellcolor{Green!17.7}12.55 \\
 LLaMA-7B             & \cellcolor{Green!0.0}1.16  & \cellcolor{Green!48.0}5.00  & \cellcolor{Green!12.4}25.45 & \cellcolor{Green!10.4}7.40  & \cellcolor{Green!0.0}21.93 & \cellcolor{Green!0.0}16.74 & \cellcolor{Green!0.0}21.70 & \cellcolor{Green!5.8}1.60   & \cellcolor{Green!50.6}14.43 & \cellcolor{Green!0.0}0.57  & \cellcolor{Green!12.7}11.60 \\
 LLaMA-13B            & \cellcolor{Green!0.0}0.68  & \cellcolor{Green!46.4}6.00  & \cellcolor{Green!25.8}22.49 & \cellcolor{Green!81.2}2.40  & \cellcolor{Green!0.0}20.51 & \cellcolor{Green!0.0}30.85 & \cellcolor{Green!0.0}19.69 & \cellcolor{Green!4.0}5.00   & \cellcolor{Green!44.0}18.06 & \cellcolor{Green!0.0}1.39  & \cellcolor{Green!20.1}12.71 \\
 LLaMA-30B            & \cellcolor{Green!0.0}0.57  & \cellcolor{Green!29.2}9.60  & \cellcolor{Green!3.4}26.74  & \cellcolor{Green!45.2}3.20  & \cellcolor{Green!0.0}23.41 & \cellcolor{Green!0.0}8.83  & \cellcolor{Green!0.0}10.31 & \cellcolor{Green!0.8}1.20   & \cellcolor{Green!39.0}27.57 & \cellcolor{Green!0.0}2.85  & \cellcolor{Green!11.8}11.43 \\
 LLaMA-65B            & \cellcolor{Green!0.0}3.94  & \cellcolor{Green!31.2}9.20  & \cellcolor{Green!21.6}23.88 & \cellcolor{Green!91.0}2.20  & \cellcolor{Green!0.0}25.26 & \cellcolor{Green!0.0}20.32 & \cellcolor{Green!0.0}13.66 & \cellcolor{Green!2.0}4.60   & \cellcolor{Green!26.0}33.72 & \cellcolor{Green!0.0}2.17  & \cellcolor{Green!17.2}13.89 \\
 LLaMA-2-7B           & \cellcolor{Green!0.0}1.42  & \cellcolor{Green!52.0}6.80  & \cellcolor{Green!21.2}19.23 & \cellcolor{Green!32.8}6.20  & \cellcolor{Green!0.0}23.53 & \cellcolor{Green!0.0}34.55 & \cellcolor{Green!0.0}30.96 & \cellcolor{Green!5.6}4.60   & \cellcolor{Green!42.4}25.29 & \cellcolor{Green!0.0}1.03  & \cellcolor{Green!15.4}15.36 \\
 LLaMA-2-13B          & \cellcolor{Green!0.0}3.73  & \cellcolor{Green!34.2}8.60  & \cellcolor{Green!7.8}31.54  & \cellcolor{Green!82.2}2.80  & \cellcolor{Green!0.0}24.88 & \cellcolor{Green!0.0}25.83 & \cellcolor{Green!0.0}20.55 & \cellcolor{Green!4.8}1.40   & \cellcolor{Green!25.2}31.02 & \cellcolor{Green!0.0}3.08  & \cellcolor{Green!15.4}15.34 \\
 LLaMA-2-70B          & \cellcolor{Green!0.0}9.93  & \cellcolor{Green!21.6}11.00 & \cellcolor{Green!11.2}34.15 & \cellcolor{Green!52.4}3.80  & \cellcolor{Green!0.0}27.27 & \cellcolor{Green!0.0}31.71 & \cellcolor{Green!0.0}4.66  & \cellcolor{Green!1.8}6.20   & \cellcolor{Green!38.4}34.30 & \cellcolor{Green!0.0}3.84  & \cellcolor{Green!12.5}16.69 \\
 LLaMA-2-Chat-7B      & \cellcolor{Green!0.0}0.13  & \cellcolor{Green!50.4}3.40  & \cellcolor{Green!22.4}16.78 & \cellcolor{Green!48.4}5.40  & \cellcolor{Green!0.0}0.64  & \cellcolor{Green!0.0}7.87  & \cellcolor{Green!0.0}1.60  & \cellcolor{Green!43.8}4.20  & \cellcolor{Green!0.6}16.56  & \cellcolor{Green!0.0}2.29  & \cellcolor{Green!16.6}5.89  \\
 LLaMA-2-Chat-13B     & \cellcolor{Green!0.0}0.23  & \cellcolor{Green!22.4}6.20  & \cellcolor{Green!3.2}31.00  & \cellcolor{Green!26.2}17.80 & \cellcolor{Green!0.0}0.97  & \cellcolor{Green!0.0}3.16  & \cellcolor{Green!0.0}1.03  & \cellcolor{Green!0.2}5.20   & \cellcolor{Green!11.2}26.85 & \cellcolor{Green!0.0}2.83  & \cellcolor{Green!6.3}9.53   \\
 LLaMA-2-Chat-70B     & \cellcolor{Green!0.0}0.00  & \cellcolor{Green!33.0}17.40 & \cellcolor{Green!6.8}38.05  & \cellcolor{Green!1.2}28.40  & \cellcolor{Green!0.0}1.16  & \cellcolor{Green!0.0}3.13  & \cellcolor{Green!0.0}1.53  & \cellcolor{Green!0.6}6.40   & \cellcolor{Green!18.0}34.88 & \cellcolor{Green!0.0}8.39  & \cellcolor{Green!6.0}13.93  \\
 Alpaca-v1.0-7B       & \cellcolor{Green!0.0}1.11  & \cellcolor{Green!77.4}3.60  & \cellcolor{Green!12.0}19.74 & \cellcolor{Green!8.8}5.60   & \cellcolor{Green!0.0}10.31 & \cellcolor{Green!0.0}9.23  & \cellcolor{Green!0.0}10.53 & \cellcolor{Green!21.4}13.60 & \cellcolor{Green!45.4}17.09 & \cellcolor{Green!0.0}3.16  & \cellcolor{Green!16.5}9.40  \\
 Vicuna-v1.3-7B       & \cellcolor{Green!0.0}1.45  & \cellcolor{Green!47.2}6.60  & \cellcolor{Green!9.0}23.36  & \cellcolor{Green!2.8}10.80  & \cellcolor{Green!0.0}9.92  & \cellcolor{Green!0.0}34.49 & \cellcolor{Green!0.0}17.25 & \cellcolor{Green!4.6}13.20  & \cellcolor{Green!18.6}24.98 & \cellcolor{Green!0.0}4.10  & \cellcolor{Green!8.2}14.61  \\
 Vicuna-v1.3-13B      & \cellcolor{Green!0.0}2.03  & \cellcolor{Green!58.2}5.00  & \cellcolor{Green!12.6}26.17 & \cellcolor{Green!7.8}8.40   & \cellcolor{Green!0.0}20.33 & \cellcolor{Green!0.0}34.89 & \cellcolor{Green!0.0}10.40 & \cellcolor{Green!5.0}16.60  & \cellcolor{Green!18.6}19.97 & \cellcolor{Green!0.0}7.20  & \cellcolor{Green!10.2}15.10 \\
 Vicuna-v1.3-33B      & \cellcolor{Green!0.0}2.02  & \cellcolor{Green!63.4}5.40  & \cellcolor{Green!11.6}23.44 & \cellcolor{Green!7.8}16.80  & \cellcolor{Green!0.0}24.64 & \cellcolor{Green!0.0}36.40 & \cellcolor{Green!0.0}24.86 & \cellcolor{Green!6.4}16.20  & \cellcolor{Green!10.0}21.97 & \cellcolor{Green!0.0}9.98  & \cellcolor{Green!9.9}18.17  \\
 WizardLM-7B          & \cellcolor{Green!0.0}0.82  & \cellcolor{Green!75.6}2.40  & \cellcolor{Green!6.8}17.24  & \cellcolor{Green!1.8}1.00   & \cellcolor{Green!0.0}13.29 & \cellcolor{Green!0.0}27.84 & \cellcolor{Green!0.0}7.14  & \cellcolor{Green!15.0}12.80 & \cellcolor{Green!21.6}14.49 & \cellcolor{Green!0.0}2.63  & \cellcolor{Green!12.1}9.97  \\
 StableBeluga2        & \cellcolor{Green!0.0}8.93  & \cellcolor{Green!35.0}15.00 & \cellcolor{Green!1.6}41.76  & \cellcolor{Green!1.0}38.00  & \cellcolor{Green!0.0}53.55 & \cellcolor{Green!0.0}64.99 & \cellcolor{Green!0.0}45.06 & \cellcolor{Green!0.2}37.60  & \cellcolor{Green!3.4}65.89  & \cellcolor{Green!0.0}40.54 & \cellcolor{Green!4.1}41.13  \\
 ChatGPT              & \cellcolor{Green!0.0}13.50 & \cellcolor{Green!2.2}40.60  & \cellcolor{Green!0.4}54.01  & \cellcolor{Green!2.8}41.40  & \cellcolor{Green!0.0}61.98 & \cellcolor{Green!0.0}74.04 & \cellcolor{Green!0.0}40.68 & \cellcolor{Green!0.0}37.40  & \cellcolor{Green!5.4}67.59  & \cellcolor{Green!0.0}40.04 & \cellcolor{Green!1.1}47.12  \\
 GPT-4                & \cellcolor{Green!0.0}18.31 & \cellcolor{Green!0.0}46.00  & \cellcolor{Green!0.0}69.99  & \cellcolor{Green!3.0}44.40  & \cellcolor{Green!0.0}64.80 & \cellcolor{Green!0.0}79.96 & \cellcolor{Green!0.0}40.52 & \cellcolor{Green!0.0}59.00  & \cellcolor{Green!2.6}76.55  & \cellcolor{Green!0.0}65.26 & \cellcolor{Green!0.6}56.48  \\
\midrule
\textbf{Chinese Oriented LLMs} \\
 MOSS-Moon            & \cellcolor{Green!0.0}2.83  & \cellcolor{Green!14.0}3.00  & \cellcolor{Green!17.4}17.44 & \cellcolor{Green!3.4}9.40   & \cellcolor{Green!0.0}2.62  & \cellcolor{Green!0.0}26.34 & \cellcolor{Green!0.0}5.87  & \cellcolor{Green!39.0}1.20  & \cellcolor{Green!50.8}25.87 & \cellcolor{Green!0.0}1.27  & \cellcolor{Green!12.5}9.58  \\
 MOSS-Moon-SFT        & \cellcolor{Green!0.0}5.10  & \cellcolor{Green!34.0}3.80  & \cellcolor{Green!10.0}21.16 & \cellcolor{Green!5.6}3.40   & \cellcolor{Green!0.0}2.28  & \cellcolor{Green!0.0}17.56 & \cellcolor{Green!0.0}6.80  & \cellcolor{Green!10.6}1.00  & \cellcolor{Green!18.0}24.58 & \cellcolor{Green!0.0}2.47  & \cellcolor{Green!7.8}8.82   \\
 TigerBot-Base        & \cellcolor{Green!0.0}12.32 & \cellcolor{Green!48.8}4.40  & \cellcolor{Green!38.8}16.94 & \cellcolor{Green!9.0}20.40  & \cellcolor{Green!0.0}36.47 & \cellcolor{Green!0.0}10.06 & \cellcolor{Green!0.0}22.05 & \cellcolor{Green!13.6}1.60  & \cellcolor{Green!14.6}41.35 & \cellcolor{Green!0.0}9.94  & \cellcolor{Green!12.5}17.55 \\
 TigerBot-SFT         & \cellcolor{Green!0.0}15.31 & \cellcolor{Green!4.4}2.60   & \cellcolor{Green!12.0}17.42 & \cellcolor{Green!9.6}22.40  & \cellcolor{Green!0.0}47.29 & \cellcolor{Green!0.0}62.78 & \cellcolor{Green!0.0}26.89 & \cellcolor{Green!0.2}0.40   & \cellcolor{Green!2.2}33.91  & \cellcolor{Green!0.0}10.42 & \cellcolor{Green!2.8}23.94  \\
 GoGPT                & \cellcolor{Green!0.0}1.98  & \cellcolor{Green!97.8}1.40  & \cellcolor{Green!38.0}15.17 & \cellcolor{Green!29.8}6.80  & \cellcolor{Green!0.0}35.32 & \cellcolor{Green!0.0}38.20 & \cellcolor{Green!0.0}19.06 & \cellcolor{Green!37.8}16.40 & \cellcolor{Green!25.2}34.53 & \cellcolor{Green!0.0}7.76  & \cellcolor{Green!22.9}17.66 \\
 ChatGLM2-6B          & \cellcolor{Green!0.0}3.17  & \cellcolor{Green!17.6}10.80 & \cellcolor{Green!58.2}10.31 & \cellcolor{Green!73.4}4.80  & \cellcolor{Green!0.0}38.38 & \cellcolor{Green!0.0}36.61 & \cellcolor{Green!0.0}15.59 & \cellcolor{Green!0.6}12.80  & \cellcolor{Green!18.2}26.72 & \cellcolor{Green!0.0}5.64  & \cellcolor{Green!16.8}16.48 \\
 Ziya-LLaMA-13B       & \cellcolor{Green!0.0}10.27 & \cellcolor{Green!48.0}15.40 & \cellcolor{Green!9.0}42.75  & \cellcolor{Green!2.2}25.20  & \cellcolor{Green!0.0}7.64  & \cellcolor{Green!0.0}56.84 & \cellcolor{Green!0.0}2.75  & \cellcolor{Green!19.0}17.00 & \cellcolor{Green!9.0}31.13  & \cellcolor{Green!0.0}2.65  & \cellcolor{Green!8.7}21.16  \\
 Baichuan-7B          & \cellcolor{Green!0.0}13.21 & \cellcolor{Green!45.8}9.60  & \cellcolor{Green!10.2}32.67 & \cellcolor{Green!18.0}13.80 & \cellcolor{Green!0.0}3.46  & \cellcolor{Green!0.0}17.79 & \cellcolor{Green!0.0}2.96  & \cellcolor{Green!7.8}2.20   & \cellcolor{Green!44.2}24.73 & \cellcolor{Green!0.0}1.29  & \cellcolor{Green!12.6}12.17 \\
 Baichuan-13B         & \cellcolor{Green!0.0}14.76 & \cellcolor{Green!26.4}11.80 & \cellcolor{Green!4.0}43.01  & \cellcolor{Green!19.8}7.40  & \cellcolor{Green!0.0}3.38  & \cellcolor{Green!0.0}12.36 & \cellcolor{Green!0.0}9.02  & \cellcolor{Green!1.8}1.60   & \cellcolor{Green!20.2}40.09 & \cellcolor{Green!0.0}2.06  & \cellcolor{Green!7.2}14.55  \\
 Baichuan-13B-Chat    & \cellcolor{Green!0.0}2.24  & \cellcolor{Green!15.6}10.80 & \cellcolor{Green!7.6}35.26  & \cellcolor{Green!99.8}0.00  & \cellcolor{Green!0.0}61.71 & \cellcolor{Green!0.0}52.76 & \cellcolor{Green!0.0}2.47  & \cellcolor{Green!97.4}1.20  & \cellcolor{Green!65.4}16.76 & \cellcolor{Green!0.0}1.57  & \cellcolor{Green!28.6}18.48 \\
 XVERSE-13B           & \cellcolor{Green!0.0}0.09  & \cellcolor{Green!18.4}0.80  & \cellcolor{Green!1.6}15.79  & \cellcolor{Green!33.6}2.00  & \cellcolor{Green!0.0}3.69  & \cellcolor{Green!0.0}1.71  & \cellcolor{Green!0.0}16.39 & \cellcolor{Green!0.4}0.00   & \cellcolor{Green!0.6}20.09  & \cellcolor{Green!0.0}0.41  & \cellcolor{Green!5.5}6.10   \\
 InternLM-7B          & \cellcolor{Green!0.0}10.73 & \cellcolor{Green!28.4}6.40  & \cellcolor{Green!1.6}26.72  & \cellcolor{Green!5.2}8.40   & \cellcolor{Green!0.0}6.04  & \cellcolor{Green!0.0}35.56 & \cellcolor{Green!0.0}14.56 & \cellcolor{Green!6.0}2.80   & \cellcolor{Green!23.2}35.04 & \cellcolor{Green!0.0}1.58  & \cellcolor{Green!6.4}14.78  \\
 InternLM-Chat-7B     & \cellcolor{Green!0.0}20.35 & \cellcolor{Green!14.0}34.20 & \cellcolor{Green!34.0}29.72 & \cellcolor{Green!6.6}32.80  & \cellcolor{Green!0.0}54.44 & \cellcolor{Green!0.0}21.93 & \cellcolor{Green!0.0}28.76 & \cellcolor{Green!0.0}26.20  & \cellcolor{Green!4.0}62.68  & \cellcolor{Green!0.0}21.14 & \cellcolor{Green!5.9}33.22  \\
 InternLM-Chat-7B-8K  & \cellcolor{Green!0.0}21.64 & \cellcolor{Green!10.0}36.60 & \cellcolor{Green!16.6}30.91 & \cellcolor{Green!6.0}33.20  & \cellcolor{Green!0.0}54.35 & \cellcolor{Green!0.0}26.86 & \cellcolor{Green!0.0}30.56 & \cellcolor{Green!0.0}30.60  & \cellcolor{Green!4.4}63.42  & \cellcolor{Green!0.0}20.69 & \cellcolor{Green!3.7}34.88  \\
 Qwen-7B              & \cellcolor{Green!0.0}8.12  & \cellcolor{Green!62.6}8.80  & \cellcolor{Green!21.4}23.17 & \cellcolor{Green!23.0}21.60 & \cellcolor{Green!0.0}6.69  & \cellcolor{Green!0.0}42.51 & \cellcolor{Green!0.0}1.72  & \cellcolor{Green!6.4}1.40   & \cellcolor{Green!20.8}34.87 & \cellcolor{Green!0.0}5.66  & \cellcolor{Green!13.4}15.45 \\
 Qwen-7B-Chat         & \cellcolor{Green!0.0}25.16 & \cellcolor{Green!38.4}27.40 & \cellcolor{Green!21.8}32.96 & \cellcolor{Green!4.0}31.20  & \cellcolor{Green!0.0}46.71 & \cellcolor{Green!0.0}57.34 & \cellcolor{Green!0.0}42.58 & \cellcolor{Green!0.0}26.80  & \cellcolor{Green!6.4}50.63  & \cellcolor{Green!0.0}21.27 & \cellcolor{Green!7.1}36.21  \\
 Yulan-Chat-2-13B     & \cellcolor{Green!0.0}6.91  & \cellcolor{Green!14.2}30.20 & \cellcolor{Green!10.6}34.00 & \cellcolor{Green!6.6}36.00  & \cellcolor{Green!0.0}34.40 & \cellcolor{Green!0.0}58.65 & \cellcolor{Green!0.0}40.80 & \cellcolor{Green!0.2}28.60  & \cellcolor{Green!16.4}49.37 & \cellcolor{Green!0.0}22.20 & \cellcolor{Green!4.8}34.11  \\
 BELLE-LLaMA-2        & \cellcolor{Green!0.0}5.99  & \cellcolor{Green!56.8}13.00 & \cellcolor{Green!1.2}41.76  & \cellcolor{Green!3.6}30.40  & \cellcolor{Green!0.0}43.30 & \cellcolor{Green!0.0}65.60 & \cellcolor{Green!0.0}41.03 & \cellcolor{Green!34.4}9.40  & \cellcolor{Green!1.2}46.64  & \cellcolor{Green!0.0}20.16 & \cellcolor{Green!9.7}31.73  \\
 Chinese-LLaMA-2-7B   & \cellcolor{Green!0.0}7.79  & \cellcolor{Green!67.6}3.60  & \cellcolor{Green!29.8}7.71  & \cellcolor{Green!64.4}6.20  & \cellcolor{Green!0.0}4.45  & \cellcolor{Green!0.0}18.48 & \cellcolor{Green!0.0}14.04 & \cellcolor{Green!1.8}8.20   & \cellcolor{Green!48.6}21.84 & \cellcolor{Green!0.0}0.64  & \cellcolor{Green!21.2}9.30  \\
 Chinese-Alpaca-2-7B  & \cellcolor{Green!0.0}7.39  & \cellcolor{Green!59.0}12.60 & \cellcolor{Green!16.2}29.65 & \cellcolor{Green!18.8}20.60 & \cellcolor{Green!0.0}37.70 & \cellcolor{Green!0.0}32.20 & \cellcolor{Green!0.0}5.47  & \cellcolor{Green!4.2}16.60  & \cellcolor{Green!10.4}37.14 & \cellcolor{Green!0.0}4.32  & \cellcolor{Green!10.9}20.37 \\
 LLaMA-2-Chinese-7B   & \cellcolor{Green!0.0}0.12  & \cellcolor{Green!81.4}5.20  & \cellcolor{Green!48.8}11.27 & \cellcolor{Green!87.4}1.40  & \cellcolor{Green!0.0}11.70 & \cellcolor{Green!0.0}31.12 & \cellcolor{Green!0.0}2.22  & \cellcolor{Green!62.0}7.00  & \cellcolor{Green!20.6}22.95 & \cellcolor{Green!0.0}4.79  & \cellcolor{Green!30.0}9.78  \\
 LLaMA-2-Chinese-13B  & \cellcolor{Green!0.0}0.24  & \cellcolor{Green!65.6}10.00 & \cellcolor{Green!2.2}24.18  & \cellcolor{Green!100.0}0.00 & \cellcolor{Green!0.0}19.79 & \cellcolor{Green!0.0}43.47 & \cellcolor{Green!0.0}1.48  & \cellcolor{Green!48.4}10.80 & \cellcolor{Green!12.6}30.91 & \cellcolor{Green!0.0}12.17 & \cellcolor{Green!22.9}15.30 \\
\midrule
\textbf{Law Specific LLMs} \\
 LexiLaw              & \cellcolor{Green!0.0}4.18  & \cellcolor{Green!25.0}15.40 & \cellcolor{Green!31.2}21.49 & \cellcolor{Green!20.2}27.20 & \cellcolor{Green!0.0}41.64 & \cellcolor{Green!0.0}31.54 & \cellcolor{Green!0.0}34.57 & \cellcolor{Green!1.4}6.00   & \cellcolor{Green!1.4}19.84  & \cellcolor{Green!0.0}8.36  & \cellcolor{Green!7.9}21.02  \\
 Fuzi-Mingcha         & \cellcolor{Green!0.0}2.86  & \cellcolor{Green!21.6}2.40  & \cellcolor{Green!42.4}17.44 & \cellcolor{Green!38.6}8.80  & \cellcolor{Green!0.0}93.35 & \cellcolor{Green!0.0}42.28 & \cellcolor{Green!0.0}31.43 & \cellcolor{Green!1.4}11.40  & \cellcolor{Green!23.2}21.26 & \cellcolor{Green!0.0}7.04  & \cellcolor{Green!12.7}23.83 \\
 HanFei               & \cellcolor{Green!0.0}14.98 & \cellcolor{Green!4.2}9.20   & \cellcolor{Green!11.2}32.71 & \cellcolor{Green!11.2}12.60 & \cellcolor{Green!0.0}36.64 & \cellcolor{Green!0.0}52.33 & \cellcolor{Green!0.0}31.15 & \cellcolor{Green!0.2}17.00  & \cellcolor{Green!6.2}25.38  & \cellcolor{Green!0.0}4.09  & \cellcolor{Green!3.3}23.61  \\
 LaWGPT-7B-beta1.0    & \cellcolor{Green!0.0}2.08  & \cellcolor{Green!83.6}4.00  & \cellcolor{Green!100.0}0.00 & \cellcolor{Green!97.0}0.40  & \cellcolor{Green!0.0}0.79  & \cellcolor{Green!0.0}3.19  & \cellcolor{Green!0.0}0.00  & \cellcolor{Green!99.8}0.00  & \cellcolor{Green!99.0}0.70  & \cellcolor{Green!0.0}0.43  & \cellcolor{Green!47.9}1.16  \\
 LaWGPT-7B-beta1.1    & \cellcolor{Green!0.0}4.70  & \cellcolor{Green!57.6}3.40  & \cellcolor{Green!53.6}4.61  & \cellcolor{Green!14.8}10.20 & \cellcolor{Green!0.0}2.26  & \cellcolor{Green!0.0}8.11  & \cellcolor{Green!0.0}4.71  & \cellcolor{Green!51.8}0.60  & \cellcolor{Green!40.0}16.66 & \cellcolor{Green!0.0}0.25  & \cellcolor{Green!21.8}5.55  \\
 Wisdom-Interrogatory & \cellcolor{Green!0.0}22.16 & \cellcolor{Green!6.8}10.00  & \cellcolor{Green!0.0}24.29  & \cellcolor{Green!0.6}13.40  & \cellcolor{Green!0.0}13.23 & \cellcolor{Green!0.0}40.10 & \cellcolor{Green!0.0}39.71 & \cellcolor{Green!0.0}0.20   & \cellcolor{Green!0.0}15.81  & \cellcolor{Green!0.0}4.02  & \cellcolor{Green!0.7}18.29  \\
 ChatLaw-13B          & \cellcolor{Green!0.0}13.01 & \cellcolor{Green!17.0}9.00  & \cellcolor{Green!47.6}30.91 & \cellcolor{Green!12.6}26.60 & \cellcolor{Green!0.0}41.41 & \cellcolor{Green!0.0}60.68 & \cellcolor{Green!0.0}42.71 & \cellcolor{Green!1.8}20.20  & \cellcolor{Green!3.2}40.27  & \cellcolor{Green!0.0}17.37 & \cellcolor{Green!8.2}30.22  \\
 Lawyer-LLaMA         & \cellcolor{Green!0.0}4.90  & \cellcolor{Green!16.0}19.20 & \cellcolor{Green!54.4}9.03  & \cellcolor{Green!75.8}3.00  & \cellcolor{Green!0.0}39.65 & \cellcolor{Green!0.0}36.33 & \cellcolor{Green!0.0}37.10 & \cellcolor{Green!6.6}0.40   & \cellcolor{Green!7.8}33.19  & \cellcolor{Green!0.0}6.12  & \cellcolor{Green!16.1}18.89 \\
 ChatLaw-33B          & \cellcolor{Green!0.0}4.30  & \cellcolor{Green!59.4}12.20 & \cellcolor{Green!23.0}33.49 & \cellcolor{Green!85.2}4.20  & \cellcolor{Green!0.0}38.87 & \cellcolor{Green!0.0}28.83 & \cellcolor{Green!0.0}34.20 & \cellcolor{Green!2.8}15.40  & \cellcolor{Green!1.4}26.18  & \cellcolor{Green!0.0}15.95 & \cellcolor{Green!17.2}21.36 \\
\hline
\end{tabular}}

\begin{tabular}{ccccccc}
	\multicolumn{2}{c}{\%abstention\qquad} & \cellcolor{Green!0.0}\quad0\% \quad & \cellcolor{Green!25}\quad25\% \quad & \cellcolor{Green!50}\quad50\% \quad &
	\cellcolor{Green!75}\quad75\% \quad & \cellcolor{Green!100}\quad100\% \quad \
\end{tabular}
\caption{One-shot Results on Legal Knowledge Understanding Tasks}
\label{tab:one-shotResults on Legal Knowledge Understanding Tasks}

\end{table*}
\endgroup

\begingroup
\begin{table*}
\centering
\small
\begin{tabular}{llllllllll}
\hline
 model\_name           & 3-1                         & 3-2                        & 3-3                         & 3-4                         & 3-5                         & 3-6                         & 3-7                         & 3-8                        & Application                 \\
\hline
\textbf{Multilingual LLMs} \\
 MPT-7B               & \cellcolor{Green!3.4}0.40   & \cellcolor{Green!0.0}9.82  & \cellcolor{Green!88.8}3.86  & \cellcolor{Green!18.6}62.23 & \cellcolor{Green!29.2}55.84 & \cellcolor{Green!0.0}2.00   & \cellcolor{Green!11.4}32.80 & \cellcolor{Green!0.0}9.41  & \cellcolor{Green!18.9}22.05 \\
 MPT-Instruct-7B      & \cellcolor{Green!13.4}0.18  & \cellcolor{Green!0.0}7.88  & \cellcolor{Green!38.0}5.75  & \cellcolor{Green!80.2}13.89 & \cellcolor{Green!51.6}37.38 & \cellcolor{Green!9.6}20.60  & \cellcolor{Green!9.0}29.40  & \cellcolor{Green!0.0}7.88  & \cellcolor{Green!25.2}15.37 \\
 LLaMA-7B             & \cellcolor{Green!9.0}0.38   & \cellcolor{Green!0.0}10.16 & \cellcolor{Green!75.4}10.45 & \cellcolor{Green!25.2}58.73 & \cellcolor{Green!34.8}50.47 & \cellcolor{Green!0.4}6.80   & \cellcolor{Green!9.2}27.40  & \cellcolor{Green!0.0}10.51 & \cellcolor{Green!19.2}21.86 \\
 LLaMA-13B            & \cellcolor{Green!10.2}1.01  & \cellcolor{Green!0.0}10.00 & \cellcolor{Green!48.6}12.42 & \cellcolor{Green!10.4}72.42 & \cellcolor{Green!33.2}53.56 & \cellcolor{Green!0.8}6.40   & \cellcolor{Green!9.6}33.60  & \cellcolor{Green!0.0}10.11 & \cellcolor{Green!14.1}24.94 \\
 LLaMA-30B            & \cellcolor{Green!8.8}4.06   & \cellcolor{Green!0.0}10.01 & \cellcolor{Green!61.0}12.80 & \cellcolor{Green!16.4}65.83 & \cellcolor{Green!34.8}50.72 & \cellcolor{Green!0.4}7.40   & \cellcolor{Green!1.8}49.60  & \cellcolor{Green!0.0}11.52 & \cellcolor{Green!15.4}26.49 \\
 LLaMA-65B            & \cellcolor{Green!9.2}6.00   & \cellcolor{Green!0.0}9.94  & \cellcolor{Green!44.4}12.51 & \cellcolor{Green!10.2}71.80 & \cellcolor{Green!24.2}60.94 & \cellcolor{Green!0.6}6.80   & \cellcolor{Green!3.2}32.80  & \cellcolor{Green!0.0}10.55 & \cellcolor{Green!11.5}26.42 \\
 LLaMA-2-7B           & \cellcolor{Green!7.8}1.29   & \cellcolor{Green!0.0}10.42 & \cellcolor{Green!44.0}12.54 & \cellcolor{Green!10.2}73.09 & \cellcolor{Green!26.8}57.36 & \cellcolor{Green!0.0}8.80   & \cellcolor{Green!3.2}32.00  & \cellcolor{Green!0.0}10.98 & \cellcolor{Green!11.5}25.81 \\
 LLaMA-2-13B          & \cellcolor{Green!7.8}3.60   & \cellcolor{Green!0.0}11.01 & \cellcolor{Green!36.4}15.07 & \cellcolor{Green!14.6}68.34 & \cellcolor{Green!30.2}55.05 & \cellcolor{Green!0.0}6.60   & \cellcolor{Green!4.6}37.40  & \cellcolor{Green!0.0}12.19 & \cellcolor{Green!11.7}26.16 \\
 LLaMA-2-70B          & \cellcolor{Green!8.0}19.50  & \cellcolor{Green!0.0}11.86 & \cellcolor{Green!40.8}18.50 & \cellcolor{Green!10.8}66.05 & \cellcolor{Green!21.8}60.36 & \cellcolor{Green!0.0}2.80   & \cellcolor{Green!5.0}53.00  & \cellcolor{Green!0.0}13.38 & \cellcolor{Green!10.8}30.68 \\
 LLaMA-2-Chat-7B      & \cellcolor{Green!11.2}1.00  & \cellcolor{Green!0.0}1.22  & \cellcolor{Green!45.6}6.55  & \cellcolor{Green!78.2}17.00 & \cellcolor{Green!88.8}8.16  & \cellcolor{Green!4.0}5.40   & \cellcolor{Green!13.4}29.80 & \cellcolor{Green!0.0}0.32  & \cellcolor{Green!30.1}8.68  \\
 LLaMA-2-Chat-13B     & \cellcolor{Green!10.6}0.53  & \cellcolor{Green!0.0}13.32 & \cellcolor{Green!52.4}16.53 & \cellcolor{Green!58.0}33.50 & \cellcolor{Green!50.8}38.01 & \cellcolor{Green!0.8}2.40   & \cellcolor{Green!2.6}35.80  & \cellcolor{Green!0.0}1.31  & \cellcolor{Green!21.9}17.68 \\
 LLaMA-2-Chat-70B     & \cellcolor{Green!18.4}6.39  & \cellcolor{Green!0.0}1.41  & \cellcolor{Green!54.6}15.87 & \cellcolor{Green!82.8}13.17 & \cellcolor{Green!91.4}6.22  & \cellcolor{Green!0.0}13.00  & \cellcolor{Green!27.4}41.60 & \cellcolor{Green!0.0}4.57  & \cellcolor{Green!34.3}12.78 \\
 Alpaca-v1.0-7B       & \cellcolor{Green!5.8}0.88   & \cellcolor{Green!0.0}27.21 & \cellcolor{Green!73.8}8.05  & \cellcolor{Green!19.2}64.85 & \cellcolor{Green!37.2}49.44 & \cellcolor{Green!18.6}17.40 & \cellcolor{Green!12.2}26.60 & \cellcolor{Green!0.0}10.25 & \cellcolor{Green!20.8}25.58 \\
 Vicuna-v1.3-7B       & \cellcolor{Green!74.4}0.24  & \cellcolor{Green!0.0}16.95 & \cellcolor{Green!74.2}8.78  & \cellcolor{Green!80.8}13.21 & \cellcolor{Green!74.0}19.31 & \cellcolor{Green!0.4}2.60   & \cellcolor{Green!8.0}38.40  & \cellcolor{Green!0.0}13.71 & \cellcolor{Green!39.0}14.15 \\
 Vicuna-v1.3-13B      & \cellcolor{Green!21.4}1.04  & \cellcolor{Green!0.0}14.39 & \cellcolor{Green!72.0}10.83 & \cellcolor{Green!46.4}40.93 & \cellcolor{Green!72.4}21.78 & \cellcolor{Green!0.6}12.40  & \cellcolor{Green!24.6}29.20 & \cellcolor{Green!0.0}14.25 & \cellcolor{Green!29.7}18.10 \\
 Vicuna-v1.3-33B      & \cellcolor{Green!15.4}2.97  & \cellcolor{Green!0.0}20.16 & \cellcolor{Green!59.4}12.24 & \cellcolor{Green!38.0}47.77 & \cellcolor{Green!76.8}17.31 & \cellcolor{Green!0.4}28.40  & \cellcolor{Green!7.2}30.00  & \cellcolor{Green!0.0}11.17 & \cellcolor{Green!24.7}21.25 \\
 WizardLM-7B          & \cellcolor{Green!20.0}0.67  & \cellcolor{Green!0.0}28.10 & \cellcolor{Green!55.6}5.30  & \cellcolor{Green!52.6}37.53 & \cellcolor{Green!51.6}37.61 & \cellcolor{Green!18.0}16.20 & \cellcolor{Green!15.8}30.80 & \cellcolor{Green!0.0}14.13 & \cellcolor{Green!26.7}21.29 \\
 StableBeluga2        & \cellcolor{Green!2.4}16.87  & \cellcolor{Green!0.0}32.44 & \cellcolor{Green!49.4}23.07 & \cellcolor{Green!4.6}75.80  & \cellcolor{Green!15.2}63.59 & \cellcolor{Green!0.0}33.00  & \cellcolor{Green!0.8}56.00  & \cellcolor{Green!0.0}16.24 & \cellcolor{Green!9.1}39.63  \\
 ChatGPT              & \cellcolor{Green!1.2}30.81  & \cellcolor{Green!0.0}34.49 & \cellcolor{Green!32.4}34.55 & \cellcolor{Green!5.6}77.12  & \cellcolor{Green!5.4}73.72  & \cellcolor{Green!0.0}31.60  & \cellcolor{Green!0.0}66.40  & \cellcolor{Green!0.0}17.17 & \cellcolor{Green!5.6}45.73  \\
 GPT-4                & \cellcolor{Green!0.4}53.20  & \cellcolor{Green!0.0}33.15 & \cellcolor{Green!23.8}41.30 & \cellcolor{Green!0.4}83.21  & \cellcolor{Green!0.4}82.74  & \cellcolor{Green!0.0}49.60  & \cellcolor{Green!0.2}77.00  & \cellcolor{Green!0.0}19.90 & \cellcolor{Green!3.1}55.01  \\
\midrule
\textbf{Chinese Oriented LLMs} \\
 MOSS-Moon            & \cellcolor{Green!20.2}18.24 & \cellcolor{Green!0.0}5.14  & \cellcolor{Green!74.6}7.59  & \cellcolor{Green!5.2}77.55  & \cellcolor{Green!20.6}63.36 & \cellcolor{Green!0.0}2.80   & \cellcolor{Green!25.2}29.20 & \cellcolor{Green!0.0}10.32 & \cellcolor{Green!18.2}26.77 \\
 MOSS-Moon-SFT        & \cellcolor{Green!7.0}18.03  & \cellcolor{Green!0.0}4.81  & \cellcolor{Green!45.0}11.31 & \cellcolor{Green!10.8}70.15 & \cellcolor{Green!6.4}75.66  & \cellcolor{Green!3.2}2.00   & \cellcolor{Green!4.2}29.20  & \cellcolor{Green!0.0}9.31  & \cellcolor{Green!9.6}27.56  \\
 TigerBot-Base        & \cellcolor{Green!38.4}14.83 & \cellcolor{Green!0.0}29.46 & \cellcolor{Green!47.0}21.81 & \cellcolor{Green!8.6}72.47  & \cellcolor{Green!53.8}36.49 & \cellcolor{Green!80.0}1.00  & \cellcolor{Green!31.4}23.20 & \cellcolor{Green!0.0}17.65 & \cellcolor{Green!32.4}27.11 \\
 TigerBot-SFT         & \cellcolor{Green!55.2}4.93  & \cellcolor{Green!0.0}28.49 & \cellcolor{Green!23.0}15.77 & \cellcolor{Green!5.0}78.04  & \cellcolor{Green!11.4}72.36 & \cellcolor{Green!1.2}2.60   & \cellcolor{Green!4.2}37.80  & \cellcolor{Green!0.0}15.77 & \cellcolor{Green!12.5}31.97 \\
 GoGPT                & \cellcolor{Green!88.6}0.00  & \cellcolor{Green!0.0}22.63 & \cellcolor{Green!43.6}16.39 & \cellcolor{Green!80.4}15.12 & \cellcolor{Green!93.6}4.13  & \cellcolor{Green!25.4}20.20 & \cellcolor{Green!13.8}30.60 & \cellcolor{Green!0.0}10.37 & \cellcolor{Green!43.2}14.93 \\
 ChatGLM2-6B          & \cellcolor{Green!41.4}14.34 & \cellcolor{Green!0.0}34.23 & \cellcolor{Green!36.0}32.09 & \cellcolor{Green!51.4}38.25 & \cellcolor{Green!23.0}61.35 & \cellcolor{Green!5.4}16.60  & \cellcolor{Green!2.0}42.80  & \cellcolor{Green!0.0}14.89 & \cellcolor{Green!19.9}31.82 \\
 Ziya-LLaMA-13B       & \cellcolor{Green!79.4}5.83  & \cellcolor{Green!0.0}33.63 & \cellcolor{Green!37.2}25.98 & \cellcolor{Green!93.4}4.51  & \cellcolor{Green!98.6}0.55  & \cellcolor{Green!2.0}26.00  & \cellcolor{Green!7.8}24.20  & \cellcolor{Green!0.0}15.53 & \cellcolor{Green!39.8}17.03 \\
 Baichuan-7B          & \cellcolor{Green!0.2}40.12  & \cellcolor{Green!0.0}6.03  & \cellcolor{Green!24.8}39.94 & \cellcolor{Green!3.6}79.02  & \cellcolor{Green!3.2}78.66  & \cellcolor{Green!0.0}7.80   & \cellcolor{Green!9.0}34.80  & \cellcolor{Green!0.0}11.27 & \cellcolor{Green!5.1}37.21  \\
 Baichuan-13B         & \cellcolor{Green!0.2}55.94  & \cellcolor{Green!0.0}6.24  & \cellcolor{Green!27.2}43.76 & \cellcolor{Green!1.2}80.07  & \cellcolor{Green!3.6}77.24  & \cellcolor{Green!0.0}14.40  & \cellcolor{Green!1.6}42.20  & \cellcolor{Green!0.0}10.36 & \cellcolor{Green!4.2}41.28  \\
 Baichuan-13B-Chat    & \cellcolor{Green!10.2}46.49 & \cellcolor{Green!0.0}33.76 & \cellcolor{Green!20.2}50.20 & \cellcolor{Green!18.8}64.70 & \cellcolor{Green!33.4}53.17 & \cellcolor{Green!12.0}30.40 & \cellcolor{Green!1.0}59.40  & \cellcolor{Green!0.0}18.71 & \cellcolor{Green!12.0}44.60 \\
 XVERSE-13B           & \cellcolor{Green!6.6}1.24   & \cellcolor{Green!0.0}6.93  & \cellcolor{Green!4.6}15.52  & \cellcolor{Green!7.2}75.53  & \cellcolor{Green!19.0}65.00 & \cellcolor{Green!0.0}0.00   & \cellcolor{Green!4.4}36.60  & \cellcolor{Green!0.0}9.88  & \cellcolor{Green!5.2}26.34  \\
 InternLM-7B          & \cellcolor{Green!0.4}42.50  & \cellcolor{Green!0.0}5.92  & \cellcolor{Green!19.8}41.43 & \cellcolor{Green!2.6}76.70  & \cellcolor{Green!4.4}77.10  & \cellcolor{Green!0.0}5.60   & \cellcolor{Green!2.6}38.00  & \cellcolor{Green!0.0}10.44 & \cellcolor{Green!3.7}37.21  \\
 InternLM-Chat-7B     & \cellcolor{Green!17.2}30.88 & \cellcolor{Green!0.0}29.10 & \cellcolor{Green!23.2}40.91 & \cellcolor{Green!8.2}75.57  & \cellcolor{Green!14.2}70.40 & \cellcolor{Green!0.0}36.00  & \cellcolor{Green!1.0}39.00  & \cellcolor{Green!0.0}14.73 & \cellcolor{Green!8.0}42.07  \\
 InternLM-Chat-7B-8K  & \cellcolor{Green!7.4}38.88  & \cellcolor{Green!0.0}28.70 & \cellcolor{Green!22.6}42.25 & \cellcolor{Green!18.8}67.74 & \cellcolor{Green!13.0}71.10 & \cellcolor{Green!0.0}36.20  & \cellcolor{Green!0.8}44.00  & \cellcolor{Green!0.0}12.11 & \cellcolor{Green!7.8}42.62  \\
 Qwen-7B              & \cellcolor{Green!89.0}3.35  & \cellcolor{Green!0.0}5.15  & \cellcolor{Green!42.8}30.63 & \cellcolor{Green!5.0}77.89  & \cellcolor{Green!33.6}51.60 & \cellcolor{Green!77.8}0.60  & \cellcolor{Green!1.0}44.60  & \cellcolor{Green!0.0}4.14  & \cellcolor{Green!31.1}27.24 \\
 Qwen-7B-Chat         & \cellcolor{Green!3.4}52.86  & \cellcolor{Green!0.0}34.49 & \cellcolor{Green!26.2}39.91 & \cellcolor{Green!6.0}78.47  & \cellcolor{Green!11.6}73.92 & \cellcolor{Green!0.0}26.80  & \cellcolor{Green!0.0}44.60  & \cellcolor{Green!0.0}20.39 & \cellcolor{Green!5.9}46.43  \\
 Yulan-Chat-2-13B     & \cellcolor{Green!2.0}5.15   & \cellcolor{Green!0.0}33.02 & \cellcolor{Green!23.4}38.94 & \cellcolor{Green!4.2}75.86  & \cellcolor{Green!5.2}74.05  & \cellcolor{Green!0.2}18.40  & \cellcolor{Green!1.0}49.40  & \cellcolor{Green!0.0}17.35 & \cellcolor{Green!4.5}39.02  \\
 BELLE-LLaMA-2        & \cellcolor{Green!5.8}2.96   & \cellcolor{Green!0.0}33.37 & \cellcolor{Green!44.0}19.67 & \cellcolor{Green!16.6}63.34 & \cellcolor{Green!25.0}57.15 & \cellcolor{Green!0.2}26.00  & \cellcolor{Green!7.6}42.20  & \cellcolor{Green!0.0}15.77 & \cellcolor{Green!12.4}32.56 \\
 Chinese-LLaMA-2-7B   & \cellcolor{Green!62.0}0.19  & \cellcolor{Green!0.0}5.97  & \cellcolor{Green!47.4}27.53 & \cellcolor{Green!66.6}25.47 & \cellcolor{Green!42.4}43.48 & \cellcolor{Green!3.0}7.80   & \cellcolor{Green!15.0}32.40 & \cellcolor{Green!0.0}11.23 & \cellcolor{Green!29.5}19.26 \\
 Chinese-Alpaca-2-7B  & \cellcolor{Green!72.0}2.37  & \cellcolor{Green!0.0}27.06 & \cellcolor{Green!35.8}32.46 & \cellcolor{Green!92.0}5.66  & \cellcolor{Green!79.2}15.53 & \cellcolor{Green!0.8}22.40  & \cellcolor{Green!7.0}35.20  & \cellcolor{Green!0.0}14.22 & \cellcolor{Green!35.9}19.36 \\
 LLaMA-2-Chinese-7B   & \cellcolor{Green!56.8}0.00  & \cellcolor{Green!0.0}30.99 & \cellcolor{Green!66.0}8.19  & \cellcolor{Green!78.2}16.42 & \cellcolor{Green!67.8}24.68 & \cellcolor{Green!93.4}1.80  & \cellcolor{Green!38.8}23.20 & \cellcolor{Green!0.0}9.87  & \cellcolor{Green!50.1}14.39 \\
 LLaMA-2-Chinese-13B  & \cellcolor{Green!47.0}0.40  & \cellcolor{Green!0.0}19.47 & \cellcolor{Green!72.0}9.70  & \cellcolor{Green!22.2}62.90 & \cellcolor{Green!66.6}25.54 & \cellcolor{Green!34.4}22.60 & \cellcolor{Green!30.0}33.60 & \cellcolor{Green!0.0}11.14 & \cellcolor{Green!34.0}23.17 \\
\midrule
\textbf{Law Specific LLMs} \\
 LexiLaw              & \cellcolor{Green!18.8}15.41 & \cellcolor{Green!0.0}33.94 & \cellcolor{Green!23.0}34.03 & \cellcolor{Green!7.0}73.66  & \cellcolor{Green!10.4}70.93 & \cellcolor{Green!28.4}12.20 & \cellcolor{Green!4.0}33.20  & \cellcolor{Green!0.0}14.68 & \cellcolor{Green!11.5}36.01 \\
 Fuzi-Mingcha         & \cellcolor{Green!0.2}3.86   & \cellcolor{Green!0.0}32.96 & \cellcolor{Green!11.6}43.60 & \cellcolor{Green!1.4}78.95  & \cellcolor{Green!1.2}79.00  & \cellcolor{Green!20.0}13.80 & \cellcolor{Green!0.0}38.20  & \cellcolor{Green!0.0}13.95 & \cellcolor{Green!4.3}38.04  \\
 HanFei               & \cellcolor{Green!18.4}7.98  & \cellcolor{Green!0.0}34.27 & \cellcolor{Green!4.8}9.80   & \cellcolor{Green!8.6}74.34  & \cellcolor{Green!10.6}72.19 & \cellcolor{Green!3.6}10.40  & \cellcolor{Green!0.2}13.40  & \cellcolor{Green!0.0}14.99 & \cellcolor{Green!5.8}29.67  \\
 LaWGPT-7B-beta1.0    & \cellcolor{Green!93.4}0.00  & \cellcolor{Green!0.0}23.75 & \cellcolor{Green!71.4}0.80  & \cellcolor{Green!99.0}0.01  & \cellcolor{Green!97.8}0.80  & \cellcolor{Green!97.8}0.00  & \cellcolor{Green!36.2}8.00  & \cellcolor{Green!0.0}0.46  & \cellcolor{Green!61.9}4.23  \\
 LaWGPT-7B-beta1.1    & \cellcolor{Green!17.4}0.49  & \cellcolor{Green!0.0}12.54 & \cellcolor{Green!31.2}13.80 & \cellcolor{Green!44.2}40.18 & \cellcolor{Green!37.0}47.30 & \cellcolor{Green!1.6}17.60  & \cellcolor{Green!7.6}26.00  & \cellcolor{Green!0.0}8.77  & \cellcolor{Green!17.4}20.84 \\
 Wisdom-Interrogatory & \cellcolor{Green!0.0}20.02  & \cellcolor{Green!0.0}23.33 & \cellcolor{Green!24.0}39.22 & \cellcolor{Green!1.0}81.16  & \cellcolor{Green!0.8}81.57  & \cellcolor{Green!0.0}13.00  & \cellcolor{Green!1.8}40.40  & \cellcolor{Green!0.0}20.67 & \cellcolor{Green!3.5}39.92  \\
 ChatLaw-13B          & \cellcolor{Green!0.4}25.99  & \cellcolor{Green!0.0}33.96 & \cellcolor{Green!26.0}12.24 & \cellcolor{Green!5.6}74.31  & \cellcolor{Green!5.4}73.01  & \cellcolor{Green!0.0}26.80  & \cellcolor{Green!3.6}42.00  & \cellcolor{Green!0.0}16.72 & \cellcolor{Green!5.1}38.13  \\
 Lawyer-LLaMA         & \cellcolor{Green!2.4}0.33   & \cellcolor{Green!0.0}27.23 & \cellcolor{Green!22.0}19.36 & \cellcolor{Green!6.0}70.99  & \cellcolor{Green!3.2}73.56  & \cellcolor{Green!59.2}6.60  & \cellcolor{Green!2.6}33.80  & \cellcolor{Green!0.0}16.02 & \cellcolor{Green!11.9}30.99 \\
 ChatLaw-33B          & \cellcolor{Green!6.0}4.89   & \cellcolor{Green!0.0}27.97 & \cellcolor{Green!58.0}17.54 & \cellcolor{Green!20.0}63.30 & \cellcolor{Green!33.0}53.03 & \cellcolor{Green!0.2}26.20  & \cellcolor{Green!10.2}43.20 & \cellcolor{Green!0.0}16.29 & \cellcolor{Green!15.9}31.55 \\
\hline
\end{tabular}

\begin{tabular}{ccccccc}
	\multicolumn{2}{c}{\%abstention\qquad} & \cellcolor{Green!0.0}\quad0\% \quad & \cellcolor{Green!25}\quad25\% \quad & \cellcolor{Green!50}\quad50\% \quad &
	\cellcolor{Green!75}\quad75\% \quad & \cellcolor{Green!100}\quad100\% \quad \
\end{tabular}
\caption{One-shot Results on Legal Knowledge Application Tasks}
\label{tab:one-shotResults on Legal Knowledge Application Tasks}

\end{table*}
\endgroup

\begingroup
\begin{table*}
\centering
\small
\begin{tabular}{lllll}
\hline
 model\_name           & Memorization                & Understanding               & Application                 & Overall                     \\
\hline
\textbf{Multilingual LLMs} \\
 MPT-7B               & \cellcolor{Green!0.1}1.83   & \cellcolor{Green!16.3}10.65 & \cellcolor{Green!18.9}22.05 & \cellcolor{Green!15.8}14.33 \\
 MPT-Instruct-7B      & \cellcolor{Green!8.2}2.03   & \cellcolor{Green!17.7}12.55 & \cellcolor{Green!25.2}15.37 & \cellcolor{Green!19.8}12.63 \\
 LLaMA-7B             & \cellcolor{Green!0.1}2.17   & \cellcolor{Green!12.7}11.60 & \cellcolor{Green!19.2}21.86 & \cellcolor{Green!14.1}14.76 \\
 LLaMA-13B            & \cellcolor{Green!0.0}2.80   & \cellcolor{Green!20.1}12.71 & \cellcolor{Green!14.1}24.94 & \cellcolor{Green!15.7}16.61 \\
 LLaMA-30B            & \cellcolor{Green!0.2}2.07   & \cellcolor{Green!11.8}11.43 & \cellcolor{Green!15.4}26.49 & \cellcolor{Green!12.1}16.52 \\
 LLaMA-65B            & \cellcolor{Green!0.0}1.60   & \cellcolor{Green!17.2}13.89 & \cellcolor{Green!11.5}26.42 & \cellcolor{Green!13.2}17.67 \\
 LLaMA-2-7B           & \cellcolor{Green!0.1}2.61   & \cellcolor{Green!15.4}15.36 & \cellcolor{Green!11.5}25.81 & \cellcolor{Green!12.3}18.26 \\
 LLaMA-2-13B          & \cellcolor{Green!0.0}2.39   & \cellcolor{Green!15.4}15.34 & \cellcolor{Green!11.7}26.16 & \cellcolor{Green!12.4}18.37 \\
 LLaMA-2-70B          & \cellcolor{Green!0.0}1.86   & \cellcolor{Green!12.5}16.69 & \cellcolor{Green!10.8}30.68 & \cellcolor{Green!10.6}20.80 \\
 LLaMA-2-Chat-7B      & \cellcolor{Green!4.4}4.11   & \cellcolor{Green!16.6}5.89  & \cellcolor{Green!30.1}8.68  & \cellcolor{Green!20.8}6.83  \\
 LLaMA-2-Chat-13B     & \cellcolor{Green!0.8}2.71   & \cellcolor{Green!6.3}9.53   & \cellcolor{Green!21.9}17.68 & \cellcolor{Green!12.0}12.11 \\
 LLaMA-2-Chat-70B     & \cellcolor{Green!0.0}5.86   & \cellcolor{Green!6.0}13.93  & \cellcolor{Green!34.3}12.78 & \cellcolor{Green!16.7}12.67 \\
 Alpaca-v1.0-7B       & \cellcolor{Green!10.7}6.89  & \cellcolor{Green!16.5}9.40  & \cellcolor{Green!20.8}25.58 & \cellcolor{Green!17.7}15.62 \\
 Vicuna-v1.3-7B       & \cellcolor{Green!0.0}6.80   & \cellcolor{Green!8.2}14.61  & \cellcolor{Green!39.0}14.15 & \cellcolor{Green!19.7}13.65 \\
 Vicuna-v1.3-13B      & \cellcolor{Green!0.0}10.21  & \cellcolor{Green!10.2}15.10 & \cellcolor{Green!29.7}18.10 & \cellcolor{Green!17.0}15.81 \\
 Vicuna-v1.3-33B      & \cellcolor{Green!0.6}15.24  & \cellcolor{Green!9.9}18.17  & \cellcolor{Green!24.7}21.25 & \cellcolor{Green!14.9}19.11 \\
 WizardLM-7B          & \cellcolor{Green!2.0}9.71   & \cellcolor{Green!12.1}9.97  & \cellcolor{Green!26.7}21.29 & \cellcolor{Green!16.9}14.47 \\
 StableBeluga2        & \cellcolor{Green!0.0}25.51  & \cellcolor{Green!4.1}41.13  & \cellcolor{Green!9.1}39.63  & \cellcolor{Green!5.7}38.97  \\
 ChatGPT              & \cellcolor{Green!0.0}26.67  & \cellcolor{Green!1.1}47.12  & \cellcolor{Green!5.6}45.73  & \cellcolor{Green!2.8}44.52  \\
 GPT-4                & \cellcolor{Green!0.1}36.00  & \cellcolor{Green!0.6}56.48  & \cellcolor{Green!3.1}55.01  & \cellcolor{Green!1.6}53.85  \\
\midrule
\textbf{Chinese Oriented LLMs} \\
 MOSS-Moon            & \cellcolor{Green!0.0}1.75   & \cellcolor{Green!12.5}9.58  & \cellcolor{Green!18.2}26.77 & \cellcolor{Green!13.5}15.68 \\
 MOSS-Moon-SFT        & \cellcolor{Green!8.9}1.74   & \cellcolor{Green!7.8}8.82   & \cellcolor{Green!9.6}27.56  & \cellcolor{Green!8.6}15.60  \\
 TigerBot-Base        & \cellcolor{Green!34.8}7.14  & \cellcolor{Green!12.5}17.55 & \cellcolor{Green!32.4}27.11 & \cellcolor{Green!22.7}20.34 \\
 TigerBot-SFT         & \cellcolor{Green!1.4}8.64   & \cellcolor{Green!2.8}23.94  & \cellcolor{Green!12.5}31.97 & \cellcolor{Green!6.6}25.62  \\
 GoGPT                & \cellcolor{Green!12.6}14.30 & \cellcolor{Green!22.9}17.66 & \cellcolor{Green!43.2}14.93 & \cellcolor{Green!30.0}16.23 \\
 ChatGLM2-6B          & \cellcolor{Green!4.1}19.74  & \cellcolor{Green!16.8}16.48 & \cellcolor{Green!19.9}31.82 & \cellcolor{Green!16.8}22.94 \\
 Ziya-LLaMA-13B       & \cellcolor{Green!0.5}14.00  & \cellcolor{Green!8.7}21.16  & \cellcolor{Green!39.8}17.03 & \cellcolor{Green!20.3}18.79 \\
 Baichuan-7B          & \cellcolor{Green!0.0}2.14   & \cellcolor{Green!12.6}12.17 & \cellcolor{Green!5.1}37.21  & \cellcolor{Green!8.3}21.18  \\
 Baichuan-13B         & \cellcolor{Green!0.0}2.53   & \cellcolor{Green!7.2}14.55  & \cellcolor{Green!4.2}41.28  & \cellcolor{Green!5.3}24.04  \\
 Baichuan-13B-Chat    & \cellcolor{Green!2.4}25.09  & \cellcolor{Green!28.6}18.48 & \cellcolor{Green!12.0}44.60 & \cellcolor{Green!19.3}29.59 \\
 XVERSE-13B           & \cellcolor{Green!1.1}1.22   & \cellcolor{Green!5.5}6.10   & \cellcolor{Green!5.2}26.34  & \cellcolor{Green!4.9}13.71  \\
 InternLM-7B          & \cellcolor{Green!0.1}1.89   & \cellcolor{Green!6.4}14.78  & \cellcolor{Green!3.7}37.21  & \cellcolor{Green!4.7}22.47  \\
 InternLM-Chat-7B     & \cellcolor{Green!0.1}26.73  & \cellcolor{Green!5.9}33.22  & \cellcolor{Green!8.0}42.07  & \cellcolor{Green!6.1}36.11  \\
 InternLM-Chat-7B-8K  & \cellcolor{Green!0.1}27.88  & \cellcolor{Green!3.7}34.88  & \cellcolor{Green!7.8}42.62  & \cellcolor{Green!5.0}37.28  \\
 Qwen-7B              & \cellcolor{Green!40.5}0.00  & \cellcolor{Green!13.4}15.45 & \cellcolor{Green!31.1}27.24 & \cellcolor{Green!23.2}18.62 \\
 Qwen-7B-Chat         & \cellcolor{Green!0.0}23.16  & \cellcolor{Green!7.1}36.21  & \cellcolor{Green!5.9}46.43  & \cellcolor{Green!5.9}38.99  \\
 Yulan-Chat-2-13B     & \cellcolor{Green!0.0}18.40  & \cellcolor{Green!4.8}34.11  & \cellcolor{Green!4.5}39.02  & \cellcolor{Green!4.2}34.51  \\
 BELLE-LLaMA-2        & \cellcolor{Green!0.1}18.99  & \cellcolor{Green!9.7}31.73  & \cellcolor{Green!12.4}32.56 & \cellcolor{Green!9.8}30.79  \\
 Chinese-LLaMA-2-7B   & \cellcolor{Green!0.7}1.98   & \cellcolor{Green!21.2}9.30  & \cellcolor{Green!29.5}19.26 & \cellcolor{Green!22.5}12.55 \\
 Chinese-Alpaca-2-7B  & \cellcolor{Green!0.6}12.92  & \cellcolor{Green!10.9}20.37 & \cellcolor{Green!35.9}19.36 & \cellcolor{Green!19.8}19.22 \\
 LLaMA-2-Chinese-7B   & \cellcolor{Green!48.9}0.20  & \cellcolor{Green!30.0}9.78  & \cellcolor{Green!50.1}14.39 & \cellcolor{Green!40.0}10.67 \\
 LLaMA-2-Chinese-13B  & \cellcolor{Green!44.9}1.60  & \cellcolor{Green!22.9}15.30 & \cellcolor{Green!34.0}23.17 & \cellcolor{Green!29.5}17.08 \\
\midrule
\textbf{Law Specific LLMs} \\
 LexiLaw              & \cellcolor{Green!2.1}14.94  & \cellcolor{Green!7.9}21.02  & \cellcolor{Green!11.5}36.01 & \cellcolor{Green!8.8}26.41  \\
 Fuzi-Mingcha         & \cellcolor{Green!2.9}16.50  & \cellcolor{Green!12.7}23.83 & \cellcolor{Green!4.3}38.04  & \cellcolor{Green!8.4}28.78  \\
 HanFei               & \cellcolor{Green!0.4}12.41  & \cellcolor{Green!3.3}23.61  & \cellcolor{Green!5.8}29.67  & \cellcolor{Green!4.0}24.91  \\
 LaWGPT-7B-beta1.0    & \cellcolor{Green!49.7}0.62  & \cellcolor{Green!47.9}1.16  & \cellcolor{Green!61.9}4.23  & \cellcolor{Green!53.7}2.33  \\
 LaWGPT-7B-beta1.1    & \cellcolor{Green!2.1}4.82   & \cellcolor{Green!21.8}5.55  & \cellcolor{Green!17.4}20.84 & \cellcolor{Green!18.1}11.59 \\
 Wisdom-Interrogatory & \cellcolor{Green!0.0}26.30  & \cellcolor{Green!0.7}18.29  & \cellcolor{Green!3.5}39.92  & \cellcolor{Green!1.8}27.74  \\
 ChatLaw-13B          & \cellcolor{Green!0.0}22.69  & \cellcolor{Green!8.2}30.22  & \cellcolor{Green!5.1}38.13  & \cellcolor{Green!6.2}32.63  \\
 Lawyer-LLaMA         & \cellcolor{Green!11.9}11.82 & \cellcolor{Green!16.1}18.89 & \cellcolor{Green!11.9}30.99 & \cellcolor{Green!14.0}23.02 \\
 ChatLaw-33B          & \cellcolor{Green!0.0}21.08  & \cellcolor{Green!17.2}21.36 & \cellcolor{Green!15.9}31.55 & \cellcolor{Green!15.0}25.41 \\
\hline
\end{tabular}

\begin{tabular}{ccccccc}
	\multicolumn{2}{c}{\%abstention\qquad} & \cellcolor{Green!0.0}\quad0\% \quad & \cellcolor{Green!25}\quad25\% \quad & \cellcolor{Green!50}\quad50\% \quad &
	\cellcolor{Green!75}\quad75\% \quad & \cellcolor{Green!100}\quad100\% \quad \
\end{tabular}
\caption{One-shot Results on Overall}
\label{tab:overall_results}

\end{table*}
\endgroup

\end{document}